\documentclass[runningheads]{llncs}

\usepackage{eccv}

\usepackage{eccvabbrv}

\usepackage{graphicx}
\usepackage{booktabs}

\usepackage{booktabs}
\usepackage{amssymb}
\usepackage{pifont}
\usepackage{multirow}
\usepackage{wrapfig}
\usepackage{placeins}
\usepackage{float}

\usepackage{tikz}
\usetikzlibrary{calc}
\usetikzlibrary{decorations.pathmorphing}
\usetikzlibrary{patterns,arrows,automata,positioning,decorations,shapes,calc}

\tikzstyle{normalvertex}=[circle,fill=white,draw=black]
\tikzstyle{emptyvertex}=[draw,circle,minimum size=7pt,inner sep=0pt]
\tikzstyle{tinyvertex}=[draw,circle,minimum size=3pt,inner sep=0pt]
\tikzstyle{thickedge}=[draw,gray!60,line width=1.6pt,-]

\usepackage{pgfplots}
\usepgfplotslibrary{colorbrewer}
\pgfplotsset{compat=newest}%

\pgfdeclarelayer{background}
\pgfsetlayers{background,main}

\usetikzlibrary{shapes}
\usetikzlibrary{arrows.meta,backgrounds,automata, chains}
\usetikzlibrary{positioning,shapes,matrix,calc}
\tikzstyle{vertex}=[circle, draw, fill=gray!80!white,thick,scale=1.2]
\tikzstyle{edge}=[draw=black, thick,-]

\def\pathOurs{figs/ours/}

\usepackage{hyperref}

\usepackage{orcidlink}

\begin{document}

\title{SGMatch: Semantic-Guided Non-Rigid \\ Shape Matching with Flow Regularization}
\titlerunning{SGMatch: Semantic-Guided Non-Rigid Shape Matching}

\author{Tianwei Ye \orcidlink{0009-0000-5781-8661} \and
Xiaoguang Mei \orcidlink{0000-0002-0239-8580} \and
Yifan Xia \orcidlink{0009-0000-5837-9496} \and \\
Fan Fan \orcidlink{0000-0002-7507-1810} \and
Jun Huang \orcidlink{0000-0001-5893-4090} \and
Jiayi Ma \orcidlink{0000-0003-3264-3265} \thanks{Corresponding author.}}

\authorrunning{Tianwei Ye et al.}

\institute{Wuhan University, Wuhan 430072, China\\
\email{twye2001@gmail.com, jyma2010@gmail.com}}

\maketitle

\begin{abstract}
Establishing accurate point-to-point correspondences between non-rigid 3D shapes remains a critical challenge, particularly under non-isometric deformations and topological noise. Existing functional map pipelines suffer from ambiguities that geometric descriptors alone cannot resolve, and spatial inconsistencies inherent in the projection of truncated spectral bases to dense pointwise correspondences. In this paper, we introduce SGMatch, a learning-based framework that couples 3D-lifted semantic cues with trajectory-level feature transport regularization. Specifically, we design a Semantic-Guided Local Cross-Attention module that integrates semantic features from vision foundation models into geometric descriptors while preserving local structural continuity. Furthermore, we adapt conditional flow matching as a time-conditioned feature transport regularizer that promotes spatially coherent point-wise recovery. Experimental results on multiple benchmarks demonstrate that SGMatch achieves competitive performance across near-isometric settings and consistent improvements under non-isometric deformations and topological noise.
\end{abstract}

\section{Introduction}
\label{sec:intro}
Establishing point-to-point correspondences between non-rigid 3D shapes is a long-standing challenge in computer vision and graphics, serving as a prerequisite for a wide range of downstream applications such as texture transfer~\cite{dinh2005texture}, pose transfer~\cite{song2023unsupervised, song20213d}, and statistical shape analysis~\cite{egger20203d,li2017learning,loper2023smpl}. While recent deep functional map frameworks~\cite{eisenberger2020deep, cao2023unsupervised, bastian2024hybrid, ye2025dcmatch} have shown significant progress, they still struggle with strong non-isometric deformations and topological noise. Moreover, relying solely on geometric descriptors can lead to ambiguities in symmetric regions. More critically, pointwise correspondences reconstructed from truncated spectral bases are prone to local inconsistencies~\cite{xia2024locality, cao2024synchronous}, even when the alignment of the spectra appears globally plausible.

Since correct mappings should align meaningful parts across different shapes, correspondence is inherently semantic. Recent studies~\cite{dutt2024diffusion, abdelreheem2023zero, morreale2024neural, xie2025echomatch} suggest that semantic cues from foundation models can be lifted onto 3D surfaces to produce robust zero-shot descriptors. However, integrating these signals into functional map pipelines is non-trivial; naive use of global semantic information tends to override local geometric structure, potentially disrupting rather than stabilizing the recovery of pointwise correspondences. We instead treat semantic cues as structure-aware anchors and constrain them to respect the manifold locality. This allows global semantics to disambiguate symmetries while preserving geometric continuity. Despite these semantic enhancements, the spatial inconsistency inherent in spectral truncation calls for a complementary treatment: we introduce continuous feature transport across manifolds as a regularization process to encourage smoothness in the recovered correspondences.

\begin{figure}[t]
\centering
\includegraphics[width=1\linewidth]{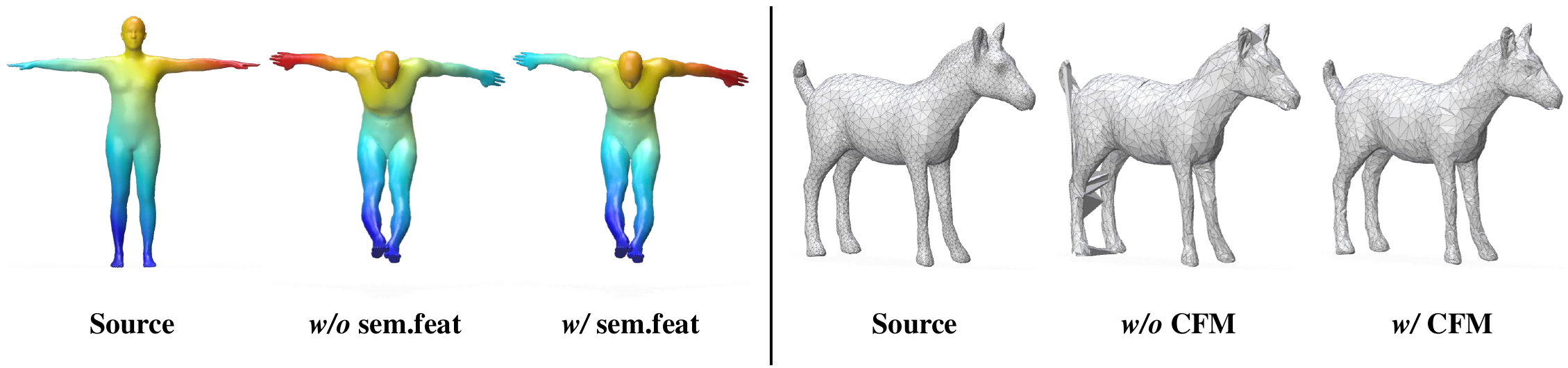}
\caption{
    \textbf{(Left):} Colormap transfer on the SHREC'19 dataset demonstrates that incorporating semantic features resolves ambiguity and yields globally consistent correspondences.
    \textbf{(Right):} Vertex transfer on the SMAL dataset shows that the proposed conditional flow matching regularization promotes spatially smooth correspondences.}
\label{fig:teaser}
\end{figure}

To this end, we propose SGMatch, a unified functional map framework that addresses the two sources of failure above. For descriptor ambiguity, the proposed Semantic-Guided Local Cross-Attention (SGLCA) module uses lifted semantic features to modulate geometric descriptors and restricts semantic interaction to mesh neighborhoods, thereby preserving local structure while improving part-level discrimination. For local inconsistency in dense recovery, we adapt conditional flow matching (CFM)~\cite{lipman2022flow, lipman2024flow} to regularize feature transport between shapes, encouraging recovered correspondences to follow coherent trajectories in feature space. As shown in \cref{fig:teaser}, SGMatch improves semantic consistency in ambiguous regions and spatial coherence under challenging deformations. Extensive experiments on diverse benchmarks demonstrate consistent gains over state-of-the-art approaches under non-isometric deformations and topological noise, while maintaining competitive performance in near-isometric cases.

Our contributions are summarized as follows:
\begin{itemize}
    \item We propose a semantic-guided functional map framework that couples 3D-lifted semantic cues with trajectory-level feature transport regularization.
    \item We design the Semantic-Guided Local Cross-Attention module to inject semantic context into geometric descriptors via channel-wise modulation and mesh-local attention, reducing ambiguity while preserving local structure.
    \item We adapt conditional flow matching as an amortized feature-transport regularizer that promotes spatially coherent point-wise recovery.
    \item SGMatch achieves consistent improvements over state-of-the-art methods, particularly under non-isometric deformations and topological noise.
\end{itemize}

\section{Related Work}
In this section, we focus on the research directions most relevant to the current study, rather than providing an exhaustive survey of the field~\cite{van2011survey, sahilliouglu2020recent}.

\subsection{Functional Maps}
The functional maps framework~\cite{ovsjanikov2012functional} is a foundational paradigm in non-rigid shape matching by shifting the problem from a search for discrete pointwise assignments to the estimation of a linear operator between functional spaces. By projecting functions onto the Laplace--Beltrami eigenbasis~\cite{pinkall1993computing}, the pipeline facilitates the representation of dense correspondences as compact, low-dimensional spectral matrices~\cite{ovsjanikov2012functional}. This formulation inherently enforces a low-frequency structural prior, which ensures stability under near-isometric deformations while significantly reducing the search space for optimization.

Subsequent work has improved functional maps via spectral refinement~\cite{eynard2016coupled, melzi2019zoomout}, structured regularization~\cite{ren2019structured}, and extensions to non-isometric~\cite{nogneng2017informative, ren2018continuous, ren2021discrete}, partial~\cite{litany2017fully, rodola2017partial}, and multi-shape matching~\cite{gao2021isometric, huang2014functional, huang2020consistent, xia2025multi}. Learning-based pipelines further replace hand-crafted descriptors~\cite{aubry2011wave, bronstein2010scale} with task-adaptive features, progressing from supervised~\cite{litany2017deep, donati2020deep, trappolini2021shape, groueix20183d} to unsupervised settings~\cite{cao2022unsupervised, cao2023unsupervised, eisenberger2021neuromorph, bastian2024hybrid, sun2023spatially}, with recent generative priors also explored in the spectral domain~\cite{zhuravlev2025denoising, pierson2025diffumatch}. Nevertheless, functional maps remain sensitive to descriptor discriminability, and point-wise correspondences recovered from truncated spectral bases can still exhibit local inconsistencies under strong non-isometric deformation and topological noise. Recent regularization methods therefore revisit smoother point-wise recovery through diffusion-based consistency constraints~\cite{cao2024synchronous} or coupled spectral-spatial interpolation~\cite{cao2024spectral}.

\subsection{Semantic Features in Shape Matching}
The emergence of large-scale vision foundation models has significantly enhanced the semantic richness of learned representations. Models from the DINO family~\cite{caron2021emerging, oquab2023dinov2, simeoni2025dinov3} leverage self-supervised distillation to capture fine-grained semantic structures, while CLIP~\cite{radford2021learning} aligns visual and textual modalities to endow image features with high-level conceptual understanding. The strong cross-instance consistency exhibited by these features across 2D tasks---ranging from segmentation~\cite{yuan2025infoclip, liang2023open} and retrieval~\cite{baldrati2023composed, oquab2023dinov2} to dense matching~\cite{zhang2024mesa, yang2025distillmatch, jiang2024omniglue}---has motivated the adoption of these features as semantic priors for the geometric analysis of 3D shapes~\cite{abdelreheem2023zero}.

Traditionally, 3D shape matching has relied on intrinsic descriptors such as HKS~\cite{bronstein2010scale} and WKS~\cite{aubry2011wave}, which can be ambiguous across symmetric or cross-category parts. Recent works lift 2D foundation-model features onto 3D surfaces: Diff3F~\cite{dutt2024diffusion} extracts multi-view semantic descriptors, while Surface-Aware Distillation~\cite{uzolas2025surface} improves the feature extraction stage. EchoMatch~\cite{xie2025echomatch} uses semantic cues for partial-to-partial matching, while DV-Matcher~\cite{chen2025dv} operates on point clouds with deformation-aware extrinsic alignment.

\section{Preliminaries}
\subsection{Deep Functional Maps Pipeline}
\label{pre:deep fm}
Given a pair of 3D shapes $\mathcal{X}$ and $\mathcal{Y}$, represented as triangular meshes with $n_{\mathcal{X}}$ and $n_{\mathcal{Y}}$ vertices, the deep functional maps pipeline proceeds as follows:

\begin{itemize}
\item \textbf{Spectral decomposition.}
The discrete Laplace--Beltrami operators $\mathbf{L}_{\mathcal{X}}$ and $\mathbf{L}_{\mathcal{Y}}$ are computed for each shape~\cite{pinkall1993computing}. From these operators, the first $k$ eigenfunctions $\mathbf{\Phi}_{\mathcal{X}} \in \mathbb{R}^{n_{\mathcal{X}} \times k}$ and $\mathbf{\Phi}_{\mathcal{Y}} \in \mathbb{R}^{n_{\mathcal{Y}} \times k}$ are obtained, along with their corresponding eigenvalue matrices $\mathbf{\Lambda}_{\mathcal{X}}, \mathbf{\Lambda}_{\mathcal{Y}} \in \mathbb{R}^{k \times k}$.

\item \textbf{Descriptor extraction.}
Vertex-wise descriptors $\mathbf{F}_{\mathcal{X}}$ and $\mathbf{F}_{\mathcal{Y}}$ are computed through the use of a learnable network $\mathcal{F}_{\theta}$~\cite{sharp2022diffusionnet}.

\item \textbf{Functional map estimation.}
The functional map $\mathbf{C}_{\mathcal{XY}} \in \mathbb{R}^{k \times k}$ is estimated by solving the following optimization problem:
\begin{equation}
\mathbf{C}_{\mathcal{XY}} = \arg\min_{\mathbf{C}} E_{\mathrm{data}}(\mathbf{C}) + \lambda E_{\mathrm{reg}}(\mathbf{C}),
\label{eq:regularied functional map}
\end{equation}
where the data term $E_{\mathrm{data}}(\mathbf{C}) = \left\| \mathbf{C} \mathbf{\Phi}_{\mathcal{X}}^{\dagger} \mathbf{F}_{\mathcal{X}} - \mathbf{\Phi}_{\mathcal{Y}}^{\dagger} \mathbf{F}_{\mathcal{Y}} \right\|_F^2$ enforces the preservation of descriptors within the spectral domain, and $E_{\mathrm{reg}}$ imposes structural constraints such as the commutativity of the Laplacian~\cite{ovsjanikov2012functional}.

\item \textbf{Point-wise map recovery.}
The point-wise correspondence matrix $\mathbf{\Pi}_{\mathcal{YX}}$ is recovered from $\mathbf{C}_{\mathcal{XY}}$ using the relationship $\mathbf{C}_{\mathcal{XY}} = \mathbf{\Phi}_{\mathcal{Y}}^{\dagger} \mathbf{\Pi}_{\mathcal{YX}} \mathbf{\Phi}_{\mathcal{X}}$. This recovery is typically achieved via a nearest-neighbor search or other post-processing strategies~\cite{melzi2019zoomout, xia2024locality}.
\end{itemize}

\subsection{Flow Matching}
\label{pre:fm}
Flow Matching~\cite{lipman2024flow, holderrieth2025introduction} is a framework for learning continuous-time probabilistic paths $(p_t)_{t \in [0,1]}$ that transform a source distribution $p_0$ into a target distribution $p_1$. The central idea is to learn a time-dependent velocity field $u_t(x)$ parameterized by a neural network, which governs the dynamics of the probability flow. For a given spatial point $x$, the velocity field induces a flow $\psi_t(x)$ defined by the ordinary differential equation (ODE) $\frac{d}{dt} \psi_t(x) = u_t\big( \psi_t(x) \big)$, $\psi_0(x) = x$, which describes the trajectory of samples evolving from the initial distribution.

Since directly modeling the marginal velocity field $u_t$ is in general intractable, Lipman et al.~\cite{lipman2022flow} introduced Conditional Flow Matching (CFM), which replaces $u_t(x)$ with a conditional velocity field $u_t(x_t \mid x_1)$ while preserving the same direction of the training gradient. The model is trained by minimizing
\begin{equation}
    \mathcal{L}_{\mathrm{CFM}}(\theta) = \mathbb{E} \left[ \left\| v_t^\theta(x_t) - u_t(x_t \mid x_1) \right\|^2 \right],
\end{equation}
where $v_t^\theta$ denotes the velocity predicted by the neural network. In practice, a simple and effective choice is to adopt a linear interpolation path $x_t = (1 - t)x_0 + t x_1$, under which the corresponding conditional velocity reduces to the constant displacement $u_t = x_1 - x_0$.
This formulation eliminates the need for numerical ODE integration during training, yielding improved optimization efficiency and numerical stability in comparison to simulation-based generative models.

\begin{figure}[!t]
\centering
\includegraphics[width=1\linewidth]{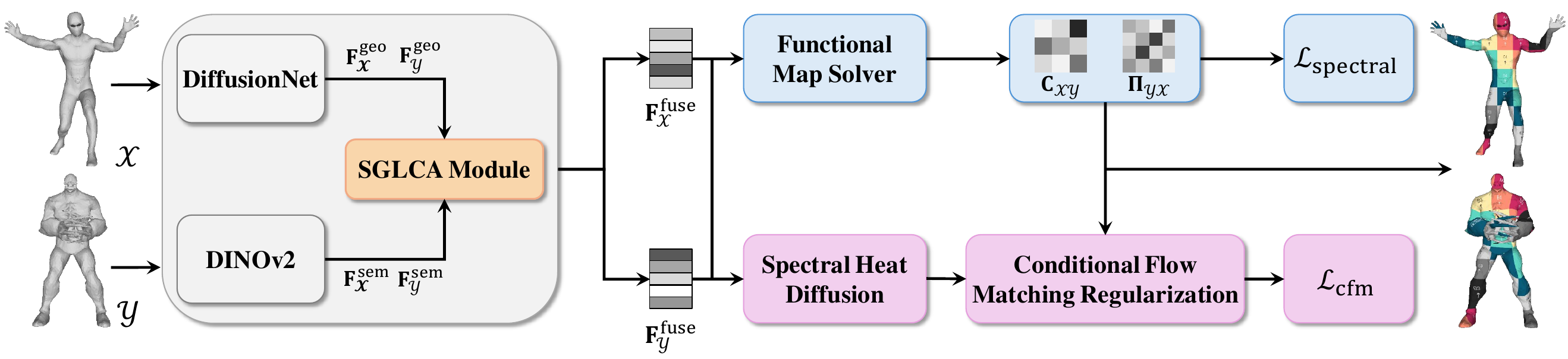}
\caption{\textbf{Overview of SGMatch.} Given a pair of shapes $\mathcal{X}$ and $\mathcal{Y}$, we extract geometric features $\mathbf{F}^{\mathrm{geo}}_{\mathcal{X}}, \mathbf{F}^{\mathrm{geo}}_{\mathcal{Y}}$ and semantic features $\mathbf{F}^{\mathrm{sem}}_{\mathcal{X}}, \mathbf{F}^{\mathrm{sem}}_{\mathcal{Y}}$, which are subsequently fused via the proposed SGLCA module. The resulting fused representations $\mathbf{F}^{\mathrm{fuse}}_{\mathcal{X}}$ and $\mathbf{F}^{\mathrm{fuse}}_{\mathcal{Y}}$ are then used to estimate functional maps $\mathbf{C}_{\mathcal{XY}}$ and to recover dense point-wise correspondences $\mathbf{\Pi}_{\mathcal{YX}}$. In parallel, spectral heat diffusion followed by conditional flow matching regularization constrains feature transport, thereby suppressing local mismatches and promoting locally smooth correspondences.}
\label{fig:framework}
\end{figure}

\section{Our SGMatch Method}
An overview of the proposed framework is illustrated in \cref{fig:framework}. We begin by detailing the feature extraction process and the subsequent feature fusion module, followed by a description of the functional map solver for estimating correspondences in the spectral domain. Finally, we introduce the conditional flow matching regularization and the overall training objective.

\subsection{Feature Extractor}
\subsubsection{Geometric Features}
Following~\cite{cao2023unsupervised,bastian2024hybrid}, we adopt DiffusionNet~\cite{sharp2022diffusionnet} to compute per-vertex descriptors that are robust to variations in mesh resolution and sampling density.
Given shapes $\mathcal{X}$ and $\mathcal{Y}$, the extracted geometric features are denoted as $\mathbf{F}^{\mathrm{geo}}_{\mathcal{X}} \in \mathbb{R}^{n_{\mathcal{X}} \times D^{\mathrm{g}}}$ and $\mathbf{F}^{\mathrm{geo}}_{\mathcal{Y}} \in \mathbb{R}^{n_{\mathcal{Y}} \times D^{\mathrm{g}}}$, respectively, where $D^{\mathrm{g}}$ is the geometric feature dimensionality.

\subsubsection{Semantic Features}
Following Diff3F~\cite{dutt2024diffusion}, we employ a multi-view feature distillation strategy to transfer 2D visual semantics onto 3D surfaces, thereby constructing vertex-level descriptors enriched with high-level semantic awareness.

For a given shape $\mathcal{X}$, we render it from multiple viewpoints $\{\xi_j\}_{j=1}^{n}$ to obtain depth- and normal-guided images $P(\cdot \mid \xi_j): \mathcal{X} \rightarrow I^{\mathcal{X}}_j \in \mathbb{R}^{H \times W}$, where $P(\cdot)$ denotes the rendering operator. A pre-trained DINOv2 encoder~\cite{oquab2023dinov2} with frozen parameters then extracts dense semantic features $\mathbf{S}_j = \psi(I^{\mathcal{X}}_j)$, where $\psi(\cdot)$ denotes the visual backbone.
The 2D semantic features are subsequently back-projected onto the 3D surface using known camera parameters:
\begin{equation}
\mathbf{S}^{3D}_j = P^{-1}(\mathbf{S}_j),
\end{equation}
thereby assigning viewpoint-specific semantic descriptors to surface vertices. Features from all viewpoints are then aggregated via averaging to obtain the vertex-level semantic representation:
\begin{equation}
\mathbf{F}^{\mathrm{sem}}_{\mathcal{X}, i} = \frac{1}{n} \sum_{j=1}^{n} \mathbf{S}^{3D}_j(i),
\end{equation}
where $\mathbf{F}^{\mathrm{sem}}_{\mathcal{X}} \in \mathbb{R}^{n_{\mathcal{X}} \times D^{\mathrm{s}}}$ denotes the semantic feature matrix defined on $\mathcal{X}$, and $D^{\mathrm{s}}$ denotes the semantic feature dimensionality. The semantic features for $\mathcal{Y}$, denoted as $\mathbf{F}^{\mathrm{sem}}_{\mathcal{Y}}$, are obtained in the same manner.

\subsection{Semantic-Guided Local Cross-Attention Module}
\label{sec:fusion}
Given geometric features $\mathbf{F}^{\mathrm{geo}}$ and semantic features $\mathbf{F}^{\mathrm{sem}}$, we aim to inject semantic context into geometric representations while preserving intrinsic structural properties and local spatial coherence. To this end, we design the Semantic-Guided Local Cross-Attention (SGLCA) module, comprising a semantic-guided gating mechanism and a neighborhood-restricted cross-attention.

\subsubsection{Semantic-Guided Gating}
We first project semantic features to the same dimensionality as geometric features via $\widetilde{\mathbf{F}}^{\mathrm{sem}} = \phi(\mathbf{F}^{\mathrm{sem}})$, where $\phi(\cdot)$ is a linear projection.
A lightweight MLP then generates channel-wise gating weights:
\begin{equation}
\mathbf{G} = \sigma\big( \mathrm{MLP}_{\mathrm{gate}}(\widetilde{\mathbf{F}}^{\mathrm{sem}}) \big),
\end{equation}
where $\sigma(\cdot)$ denotes the sigmoid function. The geometric features are then modulated as
\begin{equation}
\widetilde{\mathbf{F}}^{\mathrm{geo}} = \mathbf{F}^{\mathrm{geo}} \odot \big( 1 + \alpha \mathbf{G} \big),
\end{equation}
where $\odot$ denotes element-wise multiplication and $\alpha$ is a learnable scaling parameter.
This gating mechanism enables semantic information to adaptively amplify or attenuate geometric feature channels.

\subsubsection{Local Cross-Attention}
To avoid introducing irrelevant global interactions, attention is restricted to local neighborhoods in the mesh. For each vertex $i$ with neighborhood $\mathcal{N}(i)$, we compute
$\mathbf{Q}_i = \mathbf{W_Q} \widetilde{\mathbf{F}}^{\mathrm{geo}}_i$,
$\mathbf{K}_j = \mathbf{W_K} \widetilde{\mathbf{F}}^{\mathrm{sem}}_j$, and
$\mathbf{V}_j = \mathbf{W_V} \widetilde{\mathbf{F}}^{\mathrm{sem}}_j$,
where $\mathbf{W_Q}$, $\mathbf{W_K}$, and $\mathbf{W_V}$ are learnable projection matrices. The attention weights are defined as
\begin{equation}
\omega_{ij} = \mathrm{Softmax}_{j \in \mathcal{N}(i)} \left( (\mathbf{Q}_i \mathbf{K}_j^\top) / \sqrt{d} \right),
\end{equation}
where $d$ denotes the attention embedding dimension. The fused feature is then obtained via neighborhood aggregation:
\begin{equation}
\mathbf{F}^{\mathrm{fuse}}_i = \widetilde{\mathbf{F}}^{\mathrm{geo}}_i + \mathrm{LN} \left( \sum_{j \in \mathcal{N}(i)} \omega_{ij} \mathbf{V}_j \right),
\end{equation}
where $\mathrm{LN}(\cdot)$ denotes layer normalization.
This design allows geometric features to selectively incorporate semantically relevant information from local spatial neighborhoods, achieving complementary integration of geometry and semantics while preserving structural coherence and avoiding spurious global interactions.

\subsection{Functional Maps Module}
\label{sec:fm module}
\subsubsection{Functional Maps Computation}
We adopt the regularized functional map formulation~\cite{ren2019structured} to compute the bidirectional functional maps $\mathbf{C}_{\mathcal{XY}}$ and $\mathbf{C}_{\mathcal{YX}}$, as described in Sec.~\ref{pre:deep fm}.
Unlike classical approaches that rely solely on geometric descriptors, the functional maps are constructed using the fused vertex-level features $\mathbf{F}^{\mathrm{fuse}}_{\mathcal{X}}$ and
$\mathbf{F}^{\mathrm{fuse}}_{\mathcal{Y}}$ obtained from Sec.~\ref{sec:fusion}.

During training, we minimize a structural regularization loss defined as
\begin{equation}
\mathcal{L}_{\mathrm{struct}} = \lambda_{\mathrm{bij}} \mathcal{L}_{\mathrm{bij}} + \lambda_{\mathrm{orth}} \mathcal{L}_{\mathrm{orth}},
\label{eq:l_struct}
\end{equation}
where $\mathcal{L}_{\mathrm{bij}} = \left\| \mathbf{C}_{\mathcal{XY}} \mathbf{C}_{\mathcal{YX}} - \mathbf{I} \right\|_F^2 + \left\| \mathbf{C}_{\mathcal{YX}} \mathbf{C}_{\mathcal{XY}} - \mathbf{I} \right\|_F^2$ encourages cycle-consistency (bijectivity), and $\mathcal{L}_{\mathrm{orth}} = \left\| \mathbf{C}_{\mathcal{XY}}^{\top} \mathbf{C}_{\mathcal{XY}} - \mathbf{I} \right\|_F^2 + \left\| \mathbf{C}_{\mathcal{YX}}^{\top} \mathbf{C}_{\mathcal{YX}} - \mathbf{I} \right\|_F^2$ promotes local area preservation~\cite{ren2019structured}.

\subsubsection{Point-wise Map Computation}
In theory, a point-wise map $\mathbf{\Pi}_{\mathcal{XY}}$ is a (partial) permutation matrix satisfying
\begin{equation}
\left\{ \mathbf{\Pi} \in \{0,1\}^{n_{\mathcal{X}} \times n_{\mathcal{Y}}} : \mathbf{\Pi}\mathbf{1}_{n_{\mathcal{Y}}} = \mathbf{1}_{n_{\mathcal{X}}}, \; \mathbf{1}_{n_{\mathcal{X}}}^{\top} \mathbf{\Pi} \le \mathbf{1}_{n_{\mathcal{Y}}}^{\top} \right\},
\end{equation}
where $\mathbf{\Pi}_{\mathcal{XY}}(i,j)$ indicates that the $i$-th vertex of $\mathcal{X}$ corresponds to the $j$-th vertex of $\mathcal{Y}$.
Following prior methods~\cite{cao2023unsupervised,eisenberger2021neuromorph}, we compute a soft correspondence matrix using the fused features:
\begin{equation}
\mathbf{\Pi}_{\mathcal{XY}} = \mathrm{Softmax} \left( \left(\mathbf{F}^{\mathrm{fuse}}_{\mathcal{X}} \mathbf{F}^{\mathrm{fuse}\top}_{\mathcal{Y}}\right) / \tau_{\mathrm{T}} \right),
\label{eq:compute Pi}
\end{equation}
where $\tau_{\mathrm{T}}$ is a temperature parameter controlling the sharpness of the correspondence distribution.
The reverse map $\mathbf{\Pi}_{\mathcal{YX}}$ is computed analogously.

\subsection{Conditional Flow Matching Regularization}
\label{sec: flow matching}
Building on the CFM formulation in Sec.~\ref{pre:fm}, we design a task-specific regularization to encourage smooth feature evolution under soft correspondences. Unlike generative settings operating on distributions, the objective here regularizes vertex-level feature transport between corresponding shapes.

\subsubsection{Spectral Heat Diffusion}
To improve robustness to local noise and stabilize feature transport, we first apply spectral heat diffusion to the fused features. Given the Laplace--Beltrami eigenbasis $\mathbf{\Phi} \in \mathbb{R}^{n \times k}$ with eigenvalues
$\mathbf{\Lambda} \in \mathbb{R}^{k \times k}$ and mass matrix $\mathbf{M} \in \mathbb{R}^{n \times n}$, the diffused features are computed as~\cite{sharp2022diffusionnet, behmanesh2023tide}
\begin{equation}
\mathbf{Z} = \mathbf{\Phi} \exp(-\tau \mathbf{\Lambda}) \mathbf{\Phi}^{\top} \mathbf{M} \mathbf{F}^{\mathrm{fuse}},
\end{equation}
where $\tau > 0$ controls the diffusion scale and $\exp(-\tau \mathbf{\Lambda}) = \mathrm{diag}(e^{-\tau \lambda_1}, \dots, e^{-\tau \lambda_k})$. This spectral smoothing preserves intrinsic geometry while promoting spatial smoothness in the feature domain.

\subsubsection{Flow Path and Velocity Modeling}
Let the diffused source feature be $\mathbf{z}_0 = \mathbf{Z}_{\mathcal{X}}$ and the transported target feature be $\mathbf{z}_1 = \mathbf{\Pi}_{\mathcal{XY}} \mathbf{Z}_{\mathcal{Y}}$.
Following the commonly adopted linear interpolation strategy, we define
\begin{equation}
\mathbf{z}_t = (1 - t)\mathbf{z}_0 + t \mathbf{z}_1, \quad t \sim \mathcal{U}(0,1),
\end{equation}
with the target velocity
\begin{equation}
\mathbf{v}_{\mathrm{target}} = \mathbf{z}_1 - \mathbf{z}_0.
\end{equation}

Intuitively, the interpolation path $\mathbf{z}_t$ defines a continuous feature-space trajectory from each source vertex to its transported target feature. Although the CFM loss is evaluated on sampled vertices, it is not an independent post-processing step or a formal substitute for explicit pairwise smoothness. Its spatial bias comes from two sources: the endpoints are heat-smoothed on their respective manifolds, and all vertices share the same time-conditioned velocity network. As a result, locally inconsistent correspondences create high-frequency and mutually incompatible displacement targets that are harder to fit with the shared finite-capacity model. CFM therefore acts as an implicit trajectory-level transport prior that promotes spatially coherent point-wise recovery, rather than as a Laplacian smoothing loss.

We parameterize a learnable velocity field $\mathbf{v}_{\theta}(\mathbf{z}_t, t)$ using an MLP. To effectively encode temporal information, the scalar time variable $t$ is first mapped into a high-dimensional representation via sinusoidal embeddings, enabling the network to model non-linear temporal dependencies across the interpolation trajectory. The time-conditioned features are then injected into the MLP through Feature-wise Linear Modulation (FiLM)~\cite{perez2018film}.

\subsubsection{Importance-Weighted Objective}
Since the soft correspondence matrix $\mathbf{\Pi}_{\mathcal{XY}}$ may contain uncertain matches, we introduce similarity-based importance sampling.
For each vertex $i$, we compute a confidence weight
\begin{equation}
w_i = \exp \big( \alpha \, \cos(\mathbf{z}_{0,i}, \mathbf{z}_{1,i}) \big),
\label{eq:confidence_weight}
\end{equation}
where $\alpha$ is a scaling factor controlling the concentration of confidence weights.
A subset $\mathcal{S}$ of vertices is then sampled according to $\{w_i\}$, and the conditional flow matching objective is optimized using the Charbonnier loss in place of the standard MSE loss:
\begin{equation}
\mathcal{L}_{\mathrm{cfm}} = \mathbb{E}_{t,\, i \in \mathcal{S}} \left[ \sqrt{ \left\| \mathbf{v}_{\theta}(\mathbf{z}_{t,i}, t) - \mathbf{v}_{\mathrm{target},i} \right\|^2 + \varepsilon^2 } \right],
\label{eq:l_cfm}
\end{equation}
where $\varepsilon$ is a small constant. Compared to the MSE loss, the Charbonnier loss is less sensitive to outliers from inaccurate soft correspondences in early training, thereby stabilizing optimization and improving robustness under non-isometric deformations.

\subsection{Loss Functions}
\label{sec:loss}
The overall training objective consists of a spectral loss and a conditional flow matching (CFM) regularization term (see Eq.~\eqref{eq:l_cfm}).
The spectral loss comprises structural regularization (Eq.~\eqref{eq:l_struct}) and a coupling term that enforces consistency between functional and point-wise maps:
\begin{equation}
\mathcal{L}_{\mathrm{couple}} = \left\| \mathbf{C}_{\mathcal{XY}} - \mathbf{\Phi}_{\mathcal{Y}}^{\dagger} \mathbf{\Pi}_{\mathcal{YX}} \mathbf{\Phi}_{\mathcal{X}} \right\|_F^2 + \left\| \mathbf{C}_{\mathcal{YX}} - \mathbf{\Phi}_{\mathcal{X}}^{\dagger} \mathbf{\Pi}_{\mathcal{XY}} \mathbf{\Phi}_{\mathcal{Y}} \right\|_F^2.
\end{equation}
The full spectral loss is then given by
\begin{equation}
\mathcal{L}_{\mathrm{spectral}} = \mathcal{L}_{\mathrm{struct}} + \lambda_{\mathrm{couple}} \mathcal{L}_{\mathrm{couple}}.
\end{equation}
The final training objective is therefore
\begin{equation}
\mathcal{L}_{\mathrm{total}} = \mathcal{L}_{\mathrm{spectral}} + \lambda_{\mathrm{cfm}} \mathcal{L}_{\mathrm{cfm}}.
\end{equation}

\section{Experiments}
In this section, we conduct a comprehensive evaluation of our method against state-of-the-art approaches across multiple established benchmarks.

\subsection{Near-isometric Shape Matching}
\subsubsection{Datasets}
We evaluate on three near-isometric datasets: FAUST~\cite{bogo2014faust}, SCAPE~\cite{anguelov2005scape}, and SHREC'19~\cite{melzi2019shrec}, using their remeshed versions~\cite{ren2018continuous, donati2020deep}. FAUST contains 100 human meshes (10 subjects, 10 poses) with an 80/20 train-test split. SCAPE includes 71 meshes of one subject in various poses, split into 51 training and 20 testing samples. SHREC'19 contains 44 human shapes with diverse body types and articulations, used solely for evaluation, excluding shape 40 due to its non-closed geometry.

\subsubsection{Results}
\begin{figure}[t]
\centering
\begin{tabular}{cc}
    \resizebox{0.55\linewidth}{!}{
\setlength{\tabcolsep}{2.5pt}
\small
\begin{tabular}{@{}lccc@{}}
\toprule
Train  & \textbf{FAUST}   & \textbf{SCAPE}  & \textbf{FAUST + SCAPE} \\
\cmidrule(lr){2-2} \cmidrule(lr){3-3} \cmidrule(lr){4-4}
Test & \textbf{FAUST} & \textbf{SCAPE} & \textbf{SHREC'19}
\\ \midrule
\multicolumn{4}{c}{\textit{Axiomatic Methods}} \\
ZoomOut~\cite{melzi2019zoomout} & 6.1 & 7.5 & -\\
Smooth Shells~\cite{eisenberger2020smooth} & 2.5 & 4.2 & -\\
DiscreteOp~\cite{ren2021discrete} & 5.6 & 13.1 & -\\
\midrule
\multicolumn{4}{c}{\textit{Supervised Methods}} \\
FMNet~\cite{litany2017deep} & 11.0 & 33.0 & - \\
GeomFMaps~\cite{donati2020deep} & 2.6 & 3.0 & 7.9\\
\midrule
\multicolumn{4}{c}{\textit{Unsupervised Methods}} \\
Deep Shell~\cite{eisenberger2020deep}    & 1.7 & 2.5 & 21.1 \\
DUO-FMNet~\cite{donati2022deep}          & 2.5 & 4.2 & 6.4 \\
AttnFMaps~\cite{li2022learning}          & 1.9 & 2.2 & 5.8 \\
ULRSSM~\cite{cao2023unsupervised}        & 1.6 & 1.9 & 4.6 \\
HybridFMap~\cite{bastian2024hybrid}      & 1.5 & \textbf{1.8} & 3.6 \\
DenoisFMap~\cite{zhuravlev2025denoising} & 1.7 & 2.1 & 3.6 \\
DiffuMatch~\cite{pierson2025diffumatch}  & 1.9 & 4.4 & 3.9 \\
DeepFAFM~\cite{luo2025deep}              & 1.5 & 1.9 & 3.6 \\
\textbf{Ours} & \textbf{1.4} & \textbf{1.8} & \textbf{3.3} \\
\bottomrule
\end{tabular}
}
&
\hspace{-0.7cm}
\resizebox{0.45\linewidth}{!}{
\def\rowOnecolumnOne{44}
\def\rowOnecolumnTwo{44-15}
\def\rowTwocolumnOne{44-5}
\def\rowTwocolumnTwo{44-34}
\def\pathShrecNT{figs/ours/shrec19/}
\def\hspaceCols{-0.3cm}
\def\hspaceRows{-1.0cm}
\def\wspaceRows{1.0cm}
\def\height{3.3cm}
\def\heightT{\height}
\begin{tabular}{cc}%
    \setlength{\tabcolsep}{0pt} 
    {\small Source} & \\
    \vspace{\wspaceRows}
    \hspace{\hspaceCols}
    \includegraphics[height=\heightT]{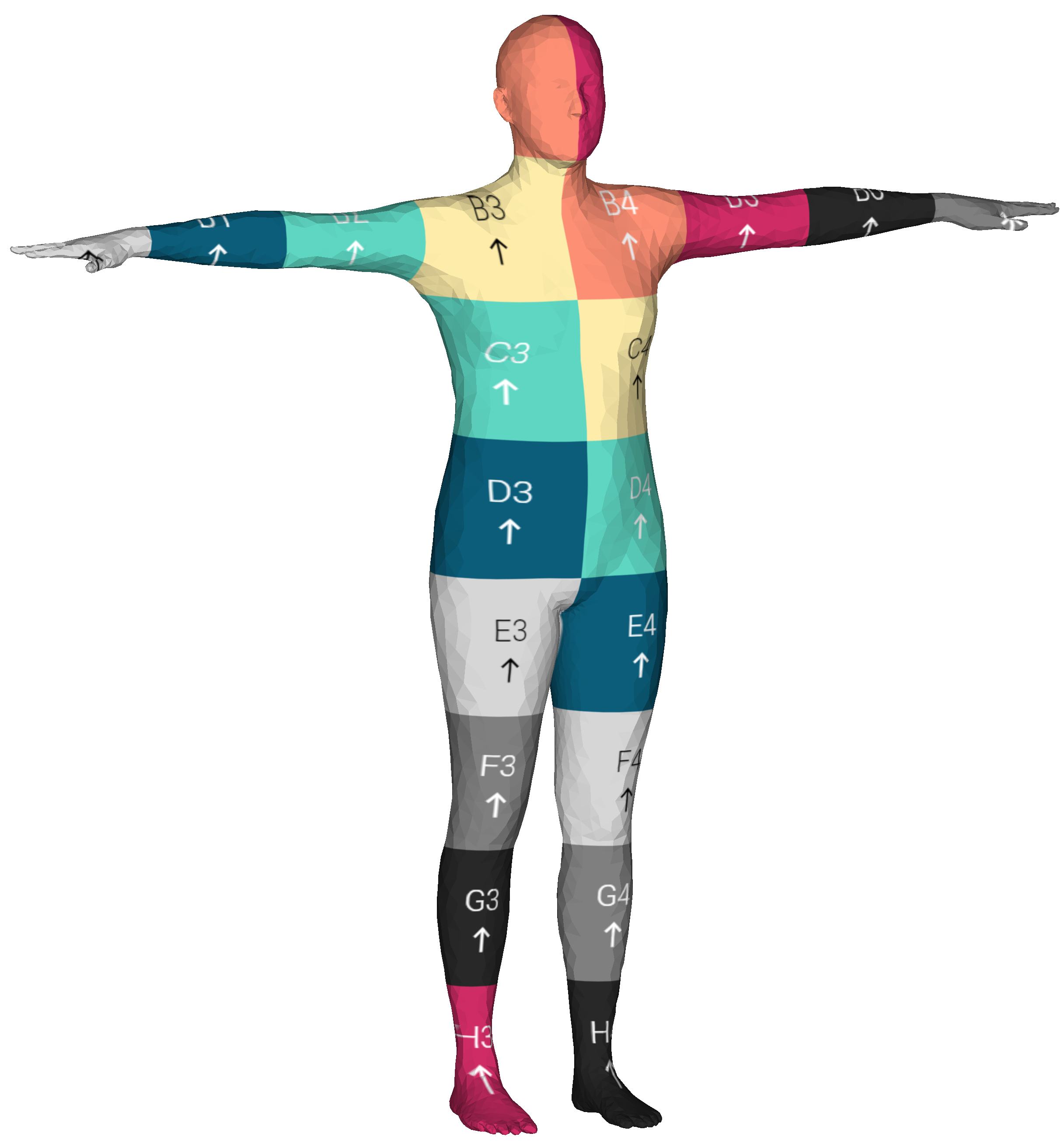}&
    \hspace{\hspaceCols}
    \includegraphics[height=\heightT]{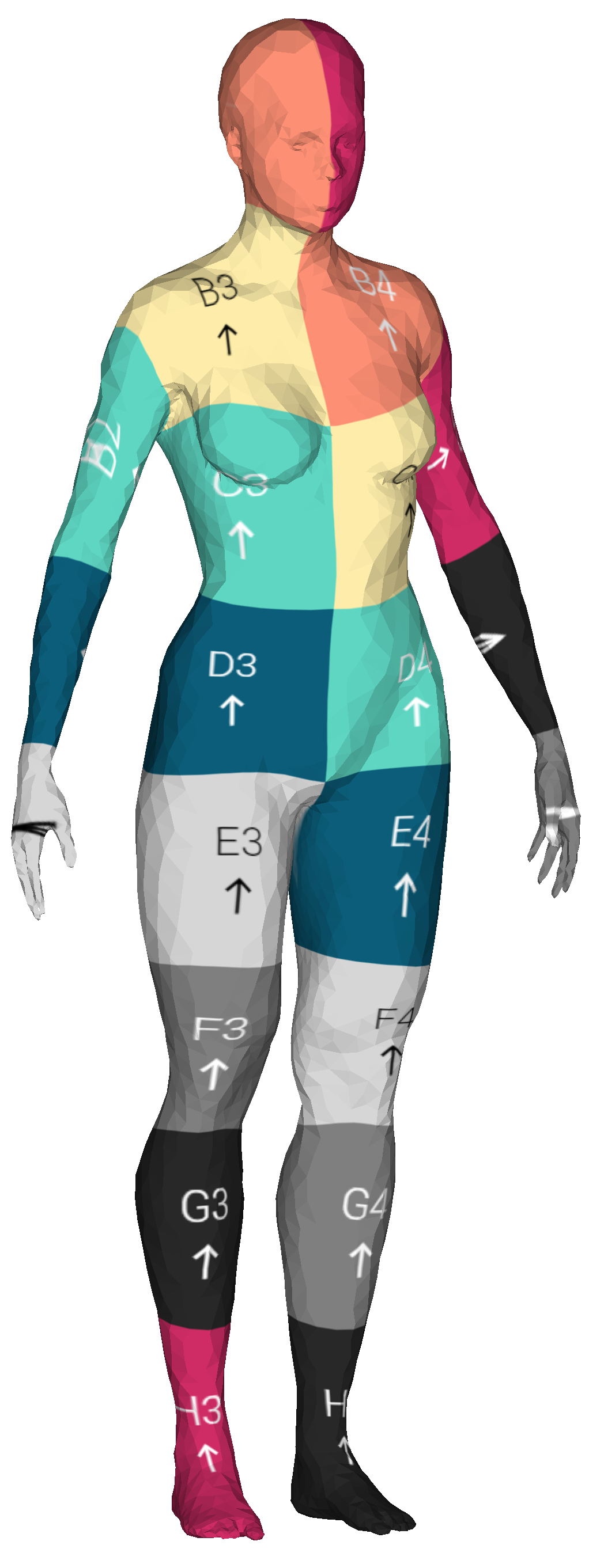}
    \\[\hspaceRows]
    \hspace{\hspaceCols}
    \includegraphics[height=\heightT]{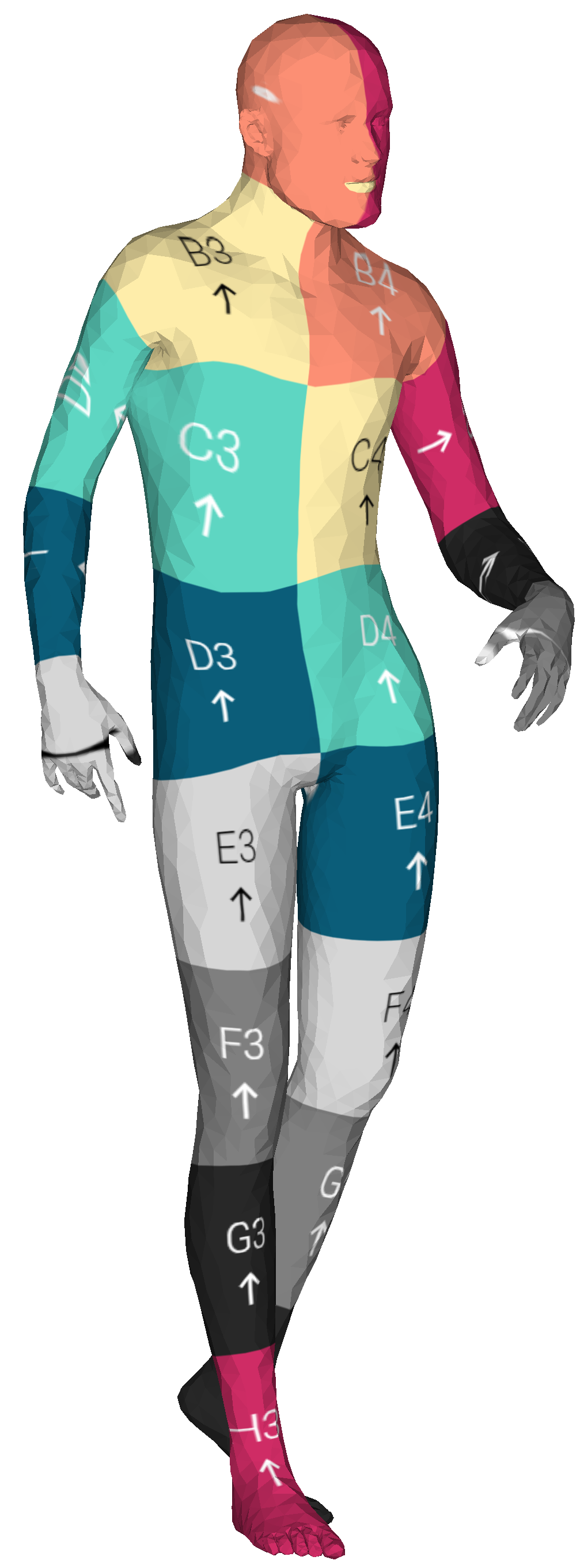}&
    \hspace{\hspaceCols}
    \includegraphics[height=\heightT]{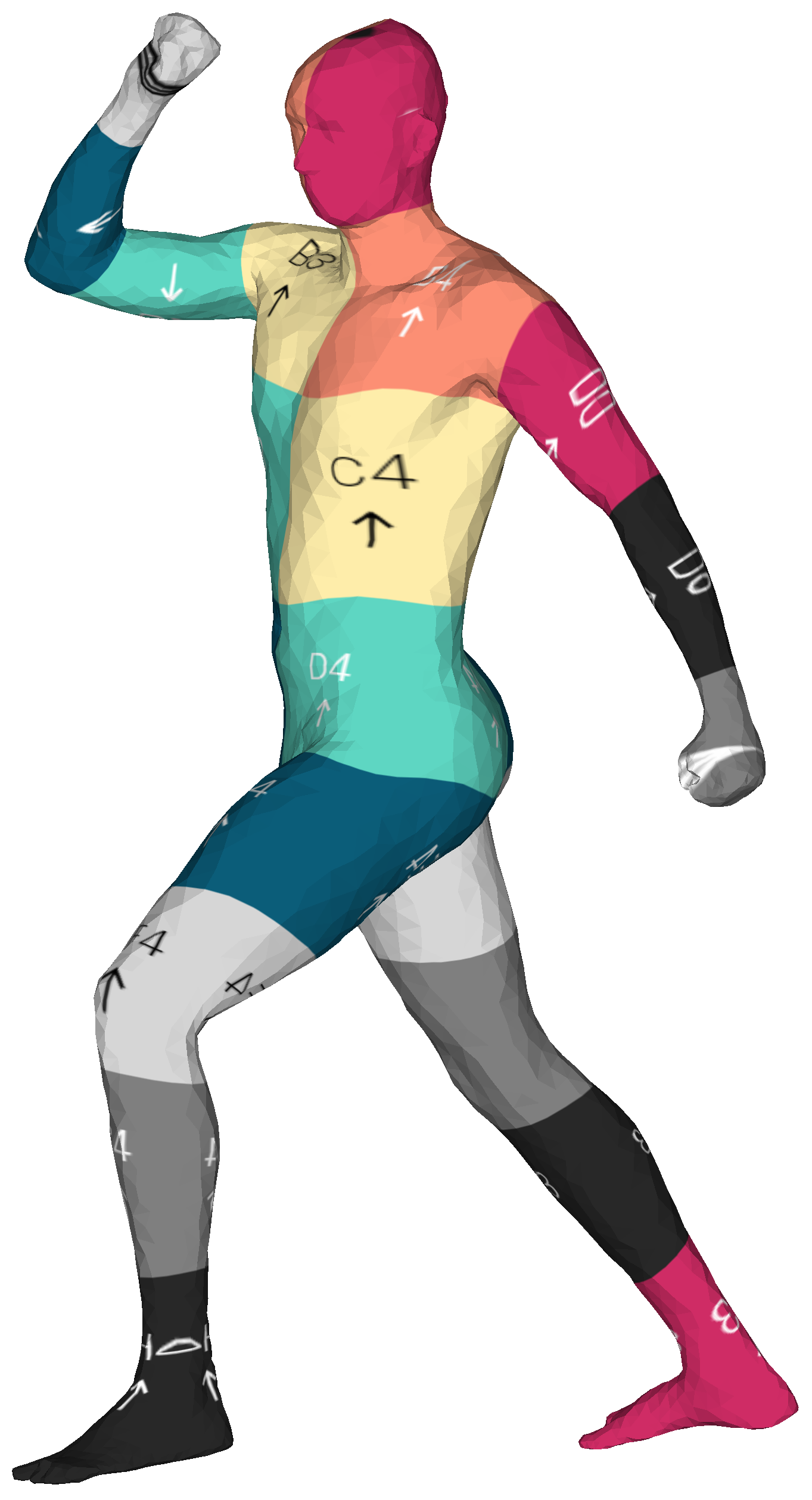}
\end{tabular}} \\
\end{tabular}
\caption{\textbf{Left: } Near-isometric matching and cross-dataset generalisation on FAUST, SCAPE, and SHREC'19. \textbf{Best} results are highlighted. \textbf{Right: } Qualitative results on the challenging SHREC'19 dataset.}
\label{tab:near-iso}
\end{figure}

We adopt the mean geodesic error as the evaluation metric~\cite{kim2011blended}. Quantitative and qualitative results in \cref{tab:near-iso} demonstrate that our method achieves competitive performance under the near-isometric setting, with marginal differences from leading baselines on FAUST and SCAPE where geometric invariance already provides strong discriminative power. Notably, on the SHREC'19 dataset, which evaluates cross-dataset generalization to shapes with diverse body types and articulations, our approach attains the best results. This suggests that the SGLCA module contributes most prominently when generalizing beyond the training distribution, where semantic priors provide complementary cues that purely geometric descriptors tend to underutilize.

\subsection{Non-isometric Shape Matching}
\subsubsection{Datasets}
We conduct non-isometric shape matching experiments on two widely adopted benchmarks: SMAL~\cite{zuffi20173d} and DT4D-H~\cite{magnet2022smooth}. The SMAL dataset comprises 49 animal shapes spanning 8 distinct species, of which 5 species are used for training and the remaining 3 are reserved for testing, resulting in a 29/20 train-test split. The DT4D-H dataset is designed for non-isometric human shape matching across multiple categories. We use 9 categories for evaluation, with 198 shapes allocated for training and 95 for testing.

\begin{table}[t]
\caption{Quantitative comparison on non-isometric shape matching.}
\label{tab:non-iso}
\renewcommand\arraystretch{0.9}
\centering
\small
\setlength{\tabcolsep}{20pt}
\begin{tabular}{lccc}
\toprule
\multirow{2}{*}{\textbf{Geo.Err ($\times 100$)}} & \multirow{2}{*}{\textbf{SMAL}} & \multicolumn{2}{c}{\textbf{DT4D-H}} \\
\cmidrule(lr){3-4}
&  & \textbf{intra} & \textbf{inter} \\
\midrule
\multicolumn{4}{c}{\textit{Axiomatic Methods}} \\
ZoomOut~\cite{melzi2019zoomout}            & 38.4 & 4.0 & 29.0 \\
Smooth Shells~\cite{eisenberger2020smooth} & 30.0 & 1.2 & 6.4  \\
DiscreteOp~\cite{ren2021discrete}          & 38.1 & 3.6 & 27.6 \\
\midrule
\multicolumn{4}{c}{\textit{Supervised Methods}} \\
FMNet~\cite{litany2017deep}     & 42.0 & 9.6 & 38.0 \\
GeomFMaps~\cite{donati2020deep} & 8.4  & 1.9 & 4.2  \\
\midrule
\multicolumn{4}{c}{\textit{Unsupervised Methods}} \\
Deep Shell~\cite{eisenberger2020deep}    & 29.3 & 3.4 & 31.1 \\
DUO-FMNet~\cite{donati2022deep}          & 6.7  & 2.6 & 15.8 \\
AttnFMaps~\cite{li2022learning}          & 5.4  & 1.7 & 11.6 \\
ULRSSM~\cite{cao2023unsupervised}        & 3.9  & \textbf{0.9} & 4.1  \\
HybridFMap~\cite{bastian2024hybrid}      & 3.3  & 1.0 & 3.5  \\
DenoisFMap~\cite{zhuravlev2025denoising} & 4.3  & 5.8 & 16.9   \\
DiffuMatch~\cite{pierson2025diffumatch}  & 10.1 & 1.8 & 8.6  \\
DeepFAFM~\cite{luo2025deep}              & 3.8  & \textbf{0.9} & 3.9  \\
\textbf{Ours}                                    & \textbf{2.5}  & 1.0 & \textbf{3.4}  \\
\bottomrule
\end{tabular}
\end{table}

\begin{figure}[!ht]
\centering
\begin{tabular}{cc|c}
    \hspace{-1cm}
    \newcommand{\pckLineWidth}{1pt}
\newcommand{\plotWidth}{0.6\columnwidth}
\newcommand{\plotHeight}{0.4\columnwidth}
\newcommand{\pckTitle}{\textbf{SMAL}}
\definecolor{cPLOT0}{RGB}{28,213,227}
\definecolor{cPLOT1}{RGB}{80,150,80}
\definecolor{cPLOT2}{RGB}{90,130,213}
\definecolor{cPLOT3}{RGB}{247,179,43}
\definecolor{cPLOT4}{RGB}{124,42,43}
\definecolor{cPLOT5}{RGB}{242,64,0}

\pgfplotsset{%
    label style = {font=\normalsize},
    tick label style = {font=\normalsize},
    title style =  {font=\normalsize},
    legend style={  fill= gray!10,
                    fill opacity=0.6, 
                    font=\normalsize,
                    draw=gray!20, %
                    text opacity=1}
}
\begin{tikzpicture}[scale=0.55, transform shape]
	\begin{axis}[
		width=\plotWidth,
		height=\plotHeight,
		grid=major,
		title=\pckTitle,
		legend style={
			at={(0.97,0.03)},
			anchor=south east,
			legend columns=1},
		legend cell align={left},
        xlabel={\large Mean Geodesic Error},
		xmin=0,
        xmax=0.2,
        ylabel near ticks,
        xtick={0, 0.05, 0.1, 0.15, 0.2},
	ymin=0,
        ymax=1,
        ytick={0.20, 0.40, 0.60, 0.80, 1.0}
	]

\addplot [color=cPLOT3, smooth, line width=\pckLineWidth]
table[row sep=crcr]{%
0.0 0.11676459244021599 \\
0.005128205128205128 0.1284462586783235 \\
0.010256410256410256 0.19742735921830804 \\
0.015384615384615385 0.3016803805605554 \\
0.020512820512820513 0.4004879146310105 \\
0.02564102564102564 0.4844613011056827 \\
0.03076923076923077 0.5645950115710979 \\
0.035897435897435895 0.6353567755206994 \\
0.041025641025641026 0.6932032656209822 \\
0.046153846153846156 0.7420095140138853 \\
0.05128205128205128 0.7837258935458987 \\
0.05641025641025641 0.8194902288506043 \\
0.06153846153846154 0.8500758549755721 \\
0.06666666666666667 0.8759244021599383 \\
0.07179487179487179 0.8976028542041656 \\
0.07692307692307693 0.9156139110311134 \\
0.08205128205128205 0.9299607868346619 \\
0.08717948717948718 0.9414746721522242 \\
0.09230769230769231 0.9505406274106454 \\
0.09743589743589744 0.9572711493957315 \\
0.10256410256410256 0.9625649267163795 \\
0.1076923076923077 0.966580097711494 \\
0.11282051282051282 0.9697222936487528 \\
0.11794871794871795 0.97209629724865 \\
0.12307692307692308 0.9740029570583698 \\
0.1282051282051282 0.9757431216250965 \\
0.13333333333333333 0.9772319362303934 \\
0.13846153846153847 0.9784874003599897 \\
0.14358974358974358 0.9796020827976344 \\
0.14871794871794872 0.9805920545127282 \\
0.15384615384615385 0.9814637438930316 \\
0.15897435897435896 0.9822370789406017 \\
0.1641025641025641 0.9829461301105683 \\
0.16923076923076924 0.9835568269478014 \\
0.17435897435897435 0.9841064541013114 \\
0.1794871794871795 0.9846194394445873 \\
0.18461538461538463 0.9850681409102597 \\
0.18974358974358974 0.9855901259964001 \\
0.19487179487179487 0.9859880431987658 \\
0.2 0.9863743893031628 \\
    };
\addlegendentry{\textcolor{black}{ULRSSM: {0.82}}}    

\addplot [color=cPLOT1, smooth, line width=\pckLineWidth]
table[row sep=crcr]{%
0.0 0.13191823090768834 \\
0.005128205128205128 0.14485921830804835 \\
0.010256410256410256 0.23045127282077654 \\
0.015384615384615385 0.3667967343790177 \\
0.020512820512820513 0.48424016456672664 \\
0.02564102564102564 0.5758440473129339 \\
0.03076923076923077 0.656398817176652 \\
0.035897435897435895 0.7204557726922088 \\
0.041025641025641026 0.7688332476214965 \\
0.046153846153846156 0.8072904345590126 \\
0.05128205128205128 0.8391803805605554 \\
0.05641025641025641 0.8656897660066856 \\
0.06153846153846154 0.8882874775006429 \\
0.06666666666666667 0.9072139367446644 \\
0.07179487179487179 0.9226626382103369 \\
0.07692307692307693 0.935311776806377 \\
0.08205128205128205 0.9454525584983287 \\
0.08717948717948718 0.9537117510928259 \\
0.09230769230769231 0.9601555669838004 \\
0.09743589743589744 0.9654448444330161 \\
0.10256410256410256 0.9697698637181795 \\
0.1076923076923077 0.9732829776292106 \\
0.11282051282051282 0.9760343275906402 \\
0.11794871794871795 0.9782032656209823 \\
0.12307692307692308 0.9799845718693752 \\
0.1282051282051282 0.9814637438930316 \\
0.13333333333333333 0.982721136538956 \\
0.13846153846153847 0.9838538184623297 \\
0.14358974358974358 0.9848065055284134 \\
0.14871794871794872 0.9856344818719465 \\
0.15384615384615385 0.986377603497043 \\
0.15897435897435896 0.9870358704037028 \\
0.1641025641025641 0.9876182823347904 \\
0.16923076923076924 0.9881736950372847 \\
0.17435897435897435 0.9887014656724094 \\
0.1794871794871795 0.989134739007457 \\
0.18461538461538463 0.9895493700179995 \\
0.18974358974358974 0.989945358704037 \\
0.19487179487179487 0.9902995628696323 \\
0.2 0.9906299820005142 \\
    };
\addlegendentry{\textcolor{black}{HybridFMap: {0.85}}}  

\addplot [color=cPLOT5, smooth, line width=\pckLineWidth]
table[row sep=crcr]{%
0.0 0.15452622782206224 \\
0.005128205128205128 0.1682019799434302 \\
0.010256410256410256 0.27031884803291334 \\
0.015384615384615385 0.43024235021856516 \\
0.020512820512820513 0.5598463615325276 \\
0.02564102564102564 0.6604004885574698 \\
0.03076923076923077 0.7459237593211623 \\
0.035897435897435895 0.8097036513242479 \\
0.041025641025641026 0.8538557469786577 \\
0.046153846153846156 0.8853567755206994 \\
0.05128205128205128 0.9085066855232707 \\
0.05641025641025641 0.9262824633581898 \\
0.06153846153846154 0.9404178452044227 \\
0.06666666666666667 0.9511616096682952 \\
0.07179487179487179 0.9595930830547699 \\
0.07692307692307693 0.9660368989457444 \\
0.08205128205128205 0.9711326819233737 \\
0.08717948717948718 0.975146567240936 \\
0.09230769230769231 0.9782392645924403 \\
0.09743589743589744 0.9808498328619182 \\
0.10256410256410256 0.9829165595268706 \\
0.1076923076923077 0.9846052969915146 \\
0.11282051282051282 0.9860021856518385 \\
0.11794871794871795 0.9871760092568784 \\
0.12307692307692308 0.988180766263821 \\
0.1282051282051282 0.9890498842890203 \\
0.13333333333333333 0.9898251478529185 \\
0.13846153846153847 0.990514271020828 \\
0.14358974358974358 0.9911326819233737 \\
0.14871794871794872 0.9916643095911546 \\
0.15384615384615385 0.9921361532527642 \\
0.15897435897435896 0.992555284134739 \\
0.1641025641025641 0.9929352018513756 \\
0.16923076923076924 0.9933009771149396 \\
0.17435897435897435 0.9936326819233736 \\
0.1794871794871795 0.9939637438930317 \\
0.18461538461538463 0.9942555926973515 \\
0.18974358974358974 0.9945255849832861 \\
0.19487179487179487 0.9947550784263307 \\
0.2 0.9949974286448958 \\
    };
\addlegendentry{\textcolor{black}{\textbf{Ours: 0.88}}}  

\end{axis}
\end{tikzpicture}&
    \hspace{-1cm}
    \newcommand{\pckLineWidth}{1pt}
\newcommand{\plotWidth}{0.6\columnwidth}
\newcommand{\plotHeight}{0.4\columnwidth}
\newcommand{\pckTitle}{\textbf{DT4D-H inter-class}}
\definecolor{cPLOT0}{RGB}{28,213,227}
\definecolor{cPLOT1}{RGB}{80,150,80}
\definecolor{cPLOT2}{RGB}{90,130,213}
\definecolor{cPLOT3}{RGB}{247,179,43}
\definecolor{cPLOT4}{RGB}{124,42,43}
\definecolor{cPLOT5}{RGB}{242,64,0}

\pgfplotsset{%
    label style = {font=\normalsize},
    tick label style = {font=\normalsize},
    title style =  {font=\normalsize},
    legend style={  fill= gray!10,
                    fill opacity=0.6, 
                    font=\normalsize,
                    draw=gray!20, %
                    text opacity=1}
}
\begin{tikzpicture}[scale=0.55, transform shape]
	\begin{axis}[
		width=\plotWidth,
		height=\plotHeight,
		grid=major,
		title=\pckTitle,
		legend style={
			at={(0.97,0.03)},
			anchor=south east,
			legend columns=1},
		legend cell align={left},
        xlabel={\large Mean Geodesic Error},
		xmin=0,
        xmax=0.2,
        ylabel near ticks,
        xtick={0, 0.05, 0.1, 0.15, 0.2},
	ymin=0,
        ymax=1,
        ytick={0.20, 0.40, 0.60, 0.80, 1.0}
	]

\addplot [color=cPLOT3, smooth, line width=\pckLineWidth]
table[row sep=crcr]{%
0.0 0.04701138045313379 \\
0.005128205128205128 0.053861433603601724 \\
0.010256410256410256 0.1139123955228547 \\
0.015384615384615385 0.23107718281676637 \\
0.020512820512820513 0.3554117598015716 \\
0.02564102564102564 0.462032536423703 \\
0.03076923076923077 0.5528308356087291 \\
0.035897435897435895 0.6312898889051004 \\
0.041025641025641026 0.6968645392583946 \\
0.046153846153846156 0.7502370927736207 \\
0.05128205128205128 0.792261604518832 \\
0.05641025641025641 0.8256408278967005 \\
0.06153846153846154 0.8527653875815495 \\
0.06666666666666667 0.8751245388415282 \\
0.07179487179487179 0.893232069533318 \\
0.07692307692307693 0.9081525522646268 \\
0.08205128205128205 0.9205471371698939 \\
0.08717948717948718 0.9305300456468725 \\
0.09230769230769231 0.938700523167351 \\
0.09743589743589744 0.9453941471955312 \\
0.10256410256410256 0.9507756841820039 \\
0.1076923076923077 0.9553523563374117 \\
0.11282051282051282 0.9593297830210309 \\
0.11794871794871795 0.9626730516705921 \\
0.12307692307692308 0.9655646455593305 \\
0.1282051282051282 0.9680355170185714 \\
0.13333333333333333 0.9700320987139671 \\
0.13846153846153847 0.9717514642432833 \\
0.14358974358974358 0.9732730266586073 \\
0.14871794871794872 0.9746756779290077 \\
0.15384615384615385 0.9760236154824187 \\
0.15897435897435896 0.9772058486358046 \\
0.1641025641025641 0.9782753194238906 \\
0.16923076923076924 0.9791223919794901 \\
0.17435897435897435 0.9798745232090377 \\
0.1794871794871795 0.9804733518144111 \\
0.18461538461538463 0.9810215311503429 \\
0.18974358974358974 0.9814875878024887 \\
0.19487179487179487 0.9819249848885925 \\
0.2 0.9823433103362028 \\
    };
\addlegendentry{\textcolor{black}{ULRSSM: {0.76}}}    

\addplot [color=cPLOT1, smooth, line width=\pckLineWidth]
table[row sep=crcr]{%
0.0 0.06155439064551765 \\
0.005128205128205128 0.07071826083331596 \\
0.010256410256410256 0.14995622902640848 \\
0.015384615384615385 0.2938760239281322 \\
0.020512820512820513 0.4394918398399233 \\
0.02564102564102564 0.5550157367071722 \\
0.03076923076923077 0.6482182087250141 \\
0.035897435897435895 0.7227372490985263 \\
0.041025641025641026 0.7811561790024387 \\
0.046153846153846156 0.8253345352981637 \\
0.05128205128205128 0.8574770202388644 \\
0.05641025641025641 0.88130885632699 \\
0.06153846153846154 0.9000991099902036 \\
0.06666666666666667 0.9152065573087105 \\
0.07179487179487179 0.92723982324864 \\
0.07692307692307693 0.9368071159097068 \\
0.08205128205128205 0.9444119057048169 \\
0.08717948717948718 0.9504082164370428 \\
0.09230769230769231 0.9552934739562707 \\
0.09743589743589744 0.9594963211538863 \\
0.10256410256410256 0.9630867915876358 \\
0.1076923076923077 0.9661820247201784 \\
0.11282051282051282 0.9688538257915251 \\
0.11794871794871795 0.9711326260499823 \\
0.12307692307692308 0.9729538112012006 \\
0.1282051282051282 0.9743918961168894 \\
0.13333333333333333 0.9755593305125373 \\
0.13846153846153847 0.9765688767534444 \\
0.14358974358974358 0.9774401275611231 \\
0.14871794871794872 0.9781495299831169 \\
0.15384615384615385 0.9787743085228339 \\
0.15897435897435896 0.9793231131583884 \\
0.1641025641025641 0.9798330450007295 \\
0.16923076923076924 0.980279508931363 \\
0.17435897435897435 0.9806740729933093 \\
0.1794871794871795 0.9810522750484607 \\
0.18461538461538463 0.9813998374220981 \\
0.18974358974358974 0.9817439606478104 \\
0.19487179487179487 0.9820532755278572 \\
0.2 0.9823867686599829 \\
    };
\addlegendentry{\textcolor{black}{HybridFMap: {0.83}}}  

\addplot [color=cPLOT5, smooth, line width=\pckLineWidth]
table[row sep=crcr]{%
0.0 0.056241011317923174 \\
0.005128205128205128 0.06322310690539217 \\
0.010256410256410256 0.1304731433812035 \\
0.015384615384615385 0.2644273297621777 \\
0.020512820512820513 0.4064219313421014 \\
0.02564102564102564 0.5244795422806762 \\
0.03076923076923077 0.6243088354836692 \\
0.035897435897435895 0.7062113095858432 \\
0.041025641025641026 0.7707888154740813 \\
0.046153846153846156 0.8201173478958668 \\
0.05128205128205128 0.8559991245805282 \\
0.05641025641025641 0.8821035079308835 \\
0.06153846153846154 0.9021178898222064 \\
0.06666666666666667 0.9175645621860474 \\
0.07179487179487179 0.9296509785939095 \\
0.07692307692307693 0.9393347854180129 \\
0.08205128205128205 0.9472767159263814 \\
0.08717948717948718 0.9537962982262334 \\
0.09230769230769231 0.9591575129749672 \\
0.09743589743589744 0.9634976134397732 \\
0.10256410256410256 0.9670864164078621 \\
0.1076923076923077 0.9700290764324572 \\
0.11282051282051282 0.9724614919648998 \\
0.11794871794871795 0.9743974195968902 \\
0.12307692307692308 0.9760473768680826 \\
0.1282051282051282 0.9773637367905454 \\
0.13333333333333333 0.9784436292390104 \\
0.13846153846153847 0.979383141922171 \\
0.14358974358974358 0.9802091627238052 \\
0.14871794871794872 0.9809267982574984 \\
0.15384615384615385 0.9816204639723201 \\
0.15897435897435896 0.982257644287888 \\
0.1641025641025641 0.9829242136857244 \\
0.16923076923076924 0.9835815078058237 \\
0.17435897435897435 0.9842477645538488 \\
0.1794871794871795 0.984936219438481 \\
0.18461538461538463 0.9856608374846281 \\
0.18974358974358974 0.9863849344477562 \\
0.19487179487179487 0.9871249765512642 \\
0.2 0.9878479271317506 \\
    };
\addlegendentry{\textcolor{black}{\textbf{Ours: 0.84}}}  

\end{axis}
\end{tikzpicture}&
    \hspace{-1cm}
    \newcommand{\pckLineWidth}{1pt}
\newcommand{\plotWidth}{0.6\columnwidth}
\newcommand{\plotHeight}{0.4\columnwidth}
\newcommand{\pckTitle}{\textbf{TOPKIDS}}
\definecolor{cPLOT0}{RGB}{28,213,227}
\definecolor{cPLOT1}{RGB}{80,150,80}
\definecolor{cPLOT2}{RGB}{90,130,213}
\definecolor{cPLOT3}{RGB}{247,179,43}
\definecolor{cPLOT4}{RGB}{124,42,43}
\definecolor{cPLOT5}{RGB}{242,64,0}

\pgfplotsset{%
    label style = {font=\normalsize},
    tick label style = {font=\normalsize},
    title style =  {font=\normalsize},
    legend style={  fill= gray!10,
                    fill opacity=0.6, 
                    font=\normalsize,
                    draw=gray!20, %
                    text opacity=1}
}
\begin{tikzpicture}[scale=0.55, transform shape]
	\begin{axis}[
		width=\plotWidth,
		height=\plotHeight,
		grid=major,
		title=\pckTitle,
		legend style={
			at={(0.97,0.03)},
			anchor=south east,
			legend columns=1},
		legend cell align={left},
        xlabel={\large Mean Geodesic Error},
		xmin=0,
        xmax=0.2,
        ylabel near ticks,
        xtick={0, 0.05, 0.1, 0.15, 0.2},
	ymin=0,
        ymax=1,
        ytick={0.20, 0.40, 0.60, 0.80, 1.0}
	]

\addplot [color=cPLOT3, smooth, line width=\pckLineWidth]
table[row sep=crcr]{%
0.0 0.14454343501582903 \\
0.005128205128205128 0.18694511312532414 \\
0.010256410256410256 0.3017307439257545 \\
0.015384615384615385 0.3975021866509795 \\
0.020512820512820513 0.4773207526723584 \\
0.02564102564102564 0.5463066884428723 \\
0.03076923076923077 0.601294187765591 \\
0.035897435897435895 0.6439977398156247 \\
0.041025641025641026 0.6781636775986315 \\
0.046153846153846156 0.7064430735411361 \\
0.05128205128205128 0.7301440480521391 \\
0.05641025641025641 0.7500948193787589 \\
0.06153846153846154 0.7668217318275758 \\
0.06666666666666667 0.7816445163437648 \\
0.07179487179487179 0.7941490638037665 \\
0.07692307692307693 0.8053145294249688 \\
0.08205128205128205 0.8146029583646173 \\
0.08717948717948718 0.8225019931420433 \\
0.09230769230769231 0.8290696864381197 \\
0.09743589743589744 0.8347859404147283 \\
0.10256410256410256 0.8399874606209314 \\
0.1076923076923077 0.8450186929632411 \\
0.11282051282051282 0.8496667776117901 \\
0.11794871794871795 0.8539123636729544 \\
0.12307692307692308 0.8578096336488819 \\
0.1282051282051282 0.8614476016502443 \\
0.13333333333333333 0.8647875658897928 \\
0.13846153846153847 0.867817915831353 \\
0.14358974358974358 0.8707747323771412 \\
0.14871794871794872 0.8736657558846067 \\
0.15384615384615385 0.8762432949153592 \\
0.15897435897435896 0.8785189600055731 \\
0.1641025641025641 0.8803921265084022 \\
0.16923076923076924 0.8821220963984117 \\
0.17435897435897435 0.8837127398543265 \\
0.1794871794871795 0.8851872779484956 \\
0.18461538461538463 0.8866308546128661 \\
0.18974358974358974 0.8878538310899198 \\
0.19487179487179487 0.8891464707840208 \\
0.2 0.8902107699333556 \\
    };
\addlegendentry{\textcolor{black}{ULRSSM: {0.77}}}    

\addplot [color=cPLOT1, smooth, line width=\pckLineWidth]
table[row sep=crcr]{%
0.0 0.25525376761898866 \\
0.005128205128205128 0.2867686329754708 \\
0.010256410256410256 0.41845920444606133 \\
0.015384615384615385 0.5282252134403567 \\
0.020512820512820513 0.6109309327904763 \\
0.02564102564102564 0.6773431997089625 \\
0.03076923076923077 0.7268040838125904 \\
0.035897435897435895 0.7646970037076313 \\
0.041025641025641026 0.79246940623718 \\
0.046153846153846156 0.8138289226196466 \\
0.05128205128205128 0.830486171851416 \\
0.05641025641025641 0.8429133157369207 \\
0.06153846153846154 0.8534015000812738 \\
0.06666666666666667 0.8614205103991702 \\
0.07179487179487179 0.8687351481891433 \\
0.07692307692307693 0.8750706307617285 \\
0.08205128205128205 0.8808488075979348 \\
0.08717948717948718 0.8859497031572918 \\
0.09230769230769231 0.8905823070909414 \\
0.09743589743589744 0.894688566718011 \\
0.10256410256410256 0.8983304048980982 \\
0.1076923076923077 0.9011749862608656 \\
0.11282051282051282 0.9039576447640352 \\
0.11794871794871795 0.9063223239649207 \\
0.12307692307692308 0.9087605365615784 \\
0.1282051282051282 0.9107923803921265 \\
0.13333333333333333 0.9127661715418018 \\
0.13846153846153847 0.9146702994744297 \\
0.14358974358974358 0.9165279852623595 \\
0.14871794871794872 0.91809927782465 \\
0.15384615384615385 0.9196202580635173 \\
0.15897435897435896 0.9209632100810415 \\
0.1641025641025641 0.9223293831709148 \\
0.16923076923076924 0.9236258930437408 \\
0.17435897435897435 0.9247830764824719 \\
0.1794871794871795 0.925944130099928 \\
0.18461538461538463 0.9273257839047008 \\
0.18974358974358974 0.9286106832413521 \\
0.19487179487179487 0.9298530106120301 \\
0.2 0.9311843520933797 \\
    };
\addlegendentry{\textcolor{black}{HybridFMap: {0.82}}}  

\addplot [color=cPLOT5, smooth, line width=\pckLineWidth]
table[row sep=crcr]{%
0.0 0.25377922952481946 \\
0.005128205128205128 0.2865441626094293 \\
0.010256410256410256 0.42879645182014503 \\
0.015384615384615385 0.5508773695169243 \\
0.020512820512820513 0.6411531584528574 \\
0.02564102564102564 0.7172563528983769 \\
0.03076923076923077 0.7749413667923185 \\
0.035897435897435895 0.8173894870465118 \\
0.041025641025641026 0.8495158406415209 \\
0.046153846153846156 0.874215321263536 \\
0.05128205128205128 0.8941002995518333 \\
0.05641025641025641 0.9089230840680222 \\
0.06153846153846154 0.9209361188299676 \\
0.06666666666666667 0.9306386568931754 \\
0.07179487179487179 0.9383712739854326 \\
0.07692307692307693 0.9449234865666096 \\
0.08205128205128205 0.9503959192835525 \\
0.08717948717948718 0.9544092946212256 \\
0.09230769230769231 0.9580240415502388 \\
0.09743589743589744 0.9612091986407932 \\
0.10256410256410256 0.9638525307098682 \\
0.1076923076923077 0.9663604065235732 \\
0.11282051282051282 0.9684851346435178 \\
0.11794871794871795 0.9701299606015806 \\
0.12307692307692308 0.9716006285170249 \\
0.1282051282051282 0.972680408381259 \\
0.13333333333333333 0.9737060057433452 \\
0.13846153846153847 0.9745729257777124 \\
0.14358974358974358 0.9752424666971121 \\
0.14871794871794872 0.9759274883314112 \\
0.15384615384615385 0.976511885318864 \\
0.15897435897435896 0.9771427244510151 \\
0.1641025641025641 0.9776458476852461 \\
0.16923076923076924 0.9780096444853823 \\
0.17435897435897435 0.9783966623578677 \\
0.1794871794871795 0.978652094153708 \\
0.18461538461538463 0.978810771481427 \\
0.18974358974358974 0.9789617084516963 \\
0.19487179487179487 0.9790932945283414 \\
0.2 0.9792945438220337 \\
    };
\addlegendentry{\textcolor{black}{\textbf{Ours: 0.89}}}  

\end{axis}
\end{tikzpicture}
\end{tabular}
\caption{
PCK curves and AUC values. \textbf{Left}: Non-isometric matching on SMAL and DT4D-H. \textbf{Right}: Matching with topological noise on TOPKIDS. Our approach achieves strong performance in both settings, outperforming existing methods.
}
\label{fig:noniso-pck}
\end{figure}
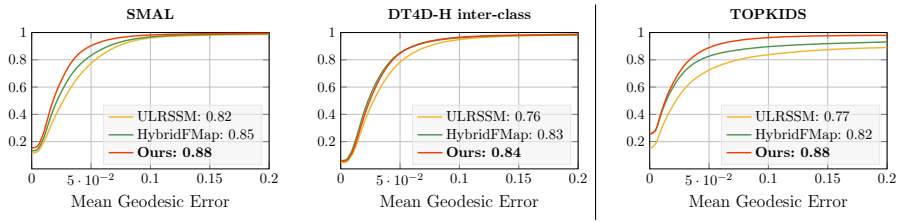

\subsubsection{Results}
\begin{figure}[t]
\centering
\includegraphics[width=1\linewidth]{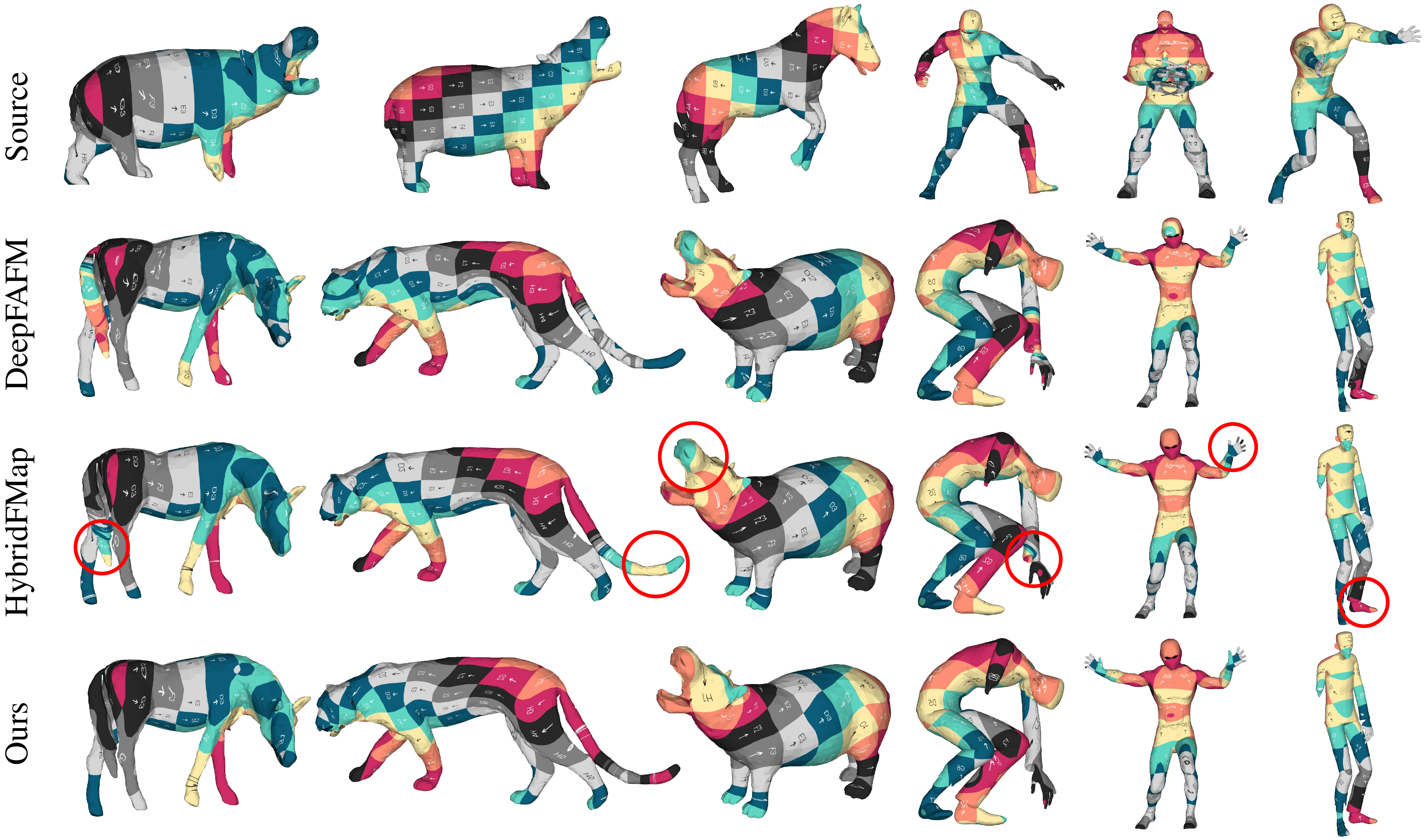}
\caption{
    \textbf{Qualitative Results on SMAL and DT4D-H.} Comparison of our method against DeepFAFM and HybridFMap, via texture transfer.}
\label{fig:qualitative_comparison}
\end{figure}

As shown in \cref{tab:non-iso}, our method achieves strong performance on non-isometric benchmarks, outperforming most existing approaches including supervised ones. On the SMAL dataset, we achieve the best overall performance, surpassing the previous state-of-the-art by 24\%. On DT4D-H, our approach attains the best intra-class result and competitive performance in the more challenging inter-class setting, where cross-category semantic consistency is harder to maintain through geometric descriptors alone. This highlights the limitations of relying solely on intrinsic geometric descriptors under large non-isometric deformations. In contrast, incorporating semantic features yields more stable cross-category correspondences. The PCK curves in \cref{fig:noniso-pck} (left) and qualitative comparisons in \cref{fig:qualitative_comparison} further confirm our method's robustness.

\begin{figure}[!ht]
\centering
\begin{tabular}{cc}
    \resizebox{0.38\linewidth}{!}{
\setlength{\tabcolsep}{4pt}
\small
\begin{tabular}{lc}
\toprule
\textbf{Geo.Err} & \textbf{TOPKIDS} \\
\midrule
\multicolumn{2}{c}{\textit{Axiomatic Methods}} \\
ZoomOut~\cite{melzi2019zoomout}            & 33.7 \\
Smooth Shells~\cite{eisenberger2020smooth} & 10.8 \\
DiscreteOp~\cite{ren2021discrete}          & 35.5 \\
\midrule
\multicolumn{2}{c}{\textit{Unsupervised Methods}} \\
Deep Shell~\cite{eisenberger2020deep}   & 13.7 \\
AttnFMaps~\cite{li2022learning}         & 23.4\\
ULRSSM~\cite{cao2023unsupervised}       & 9.2 \\
HybridFMap~\cite{bastian2024hybrid}     & 5.0 \\
DeepFAFM~\cite{luo2025deep}             & 6.2 \\
\textbf{Ours} & \textbf{2.9}\\
\bottomrule
\end{tabular}
}
&
\hspace{-1.0cm}
\resizebox{0.62\linewidth}{!}{
\def\topkidSrc{kid00}
\def\topkidA{kid17}
\def\topkidB{kid19}
\def\pathFAFM{figs/deepfafm/topkids/}
\def\pathHybFM{figs/hybridfmap/topkids/}
\def\pathOurs{figs/ours/topkids/}
\def\hspaceCols{0.5cm}
\def\hspaceRows{-1.0cm}
\def\wspaceRows{1.0cm}
\def\height{4.6cm}
\def\heightT{\height}
\begin{tabular}{cccc}%
    \setlength{\tabcolsep}{0pt} 
    {\Large Source} & {\Large DeepFAFM} & {\Large HybridFMap} & {\Large Ours}\\
    \vspace{\wspaceRows}
    \hspace{\hspaceCols}
    \includegraphics[height=\heightT]{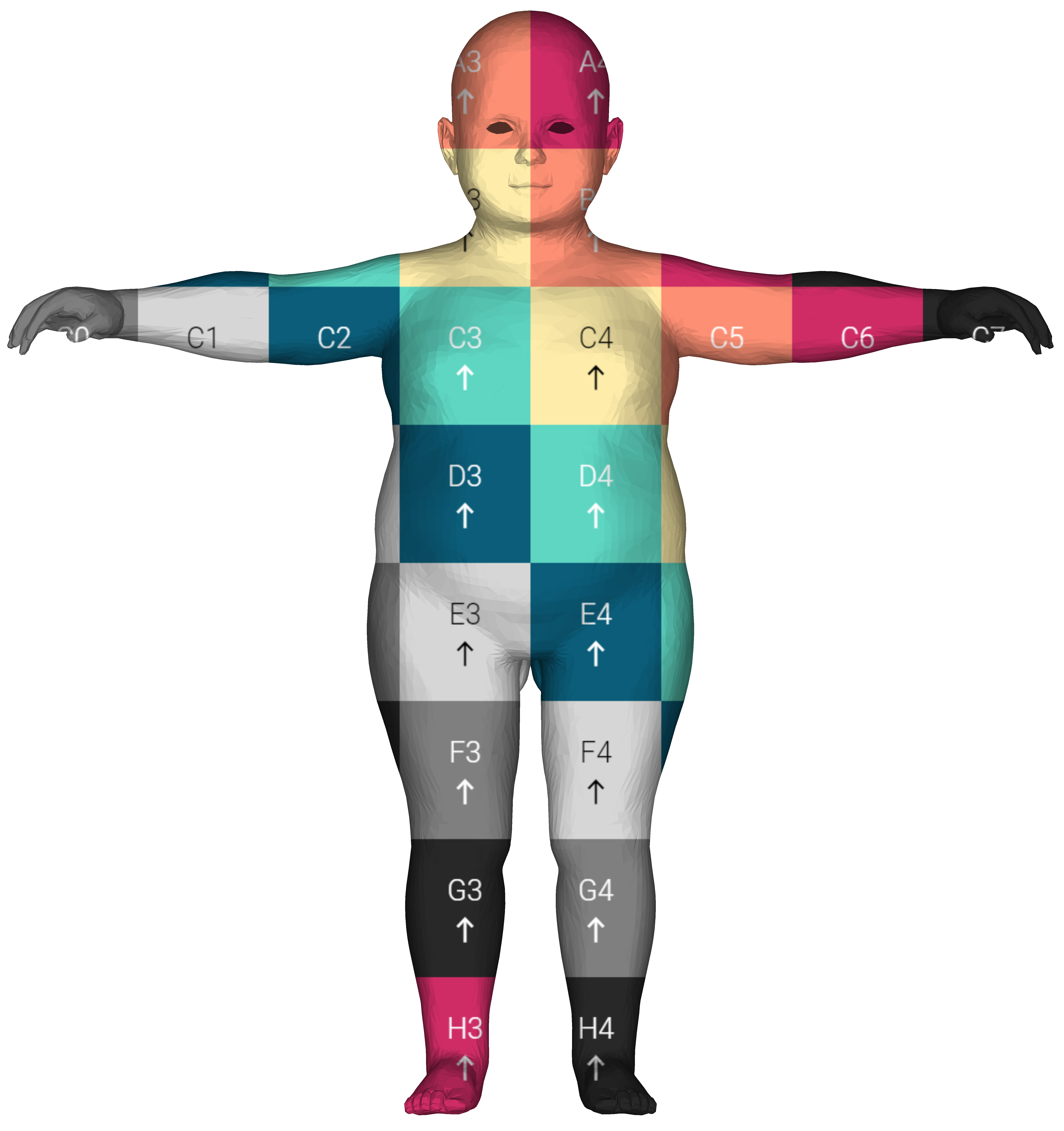}&
    \hspace{\hspaceCols}
    \includegraphics[height=\heightT]{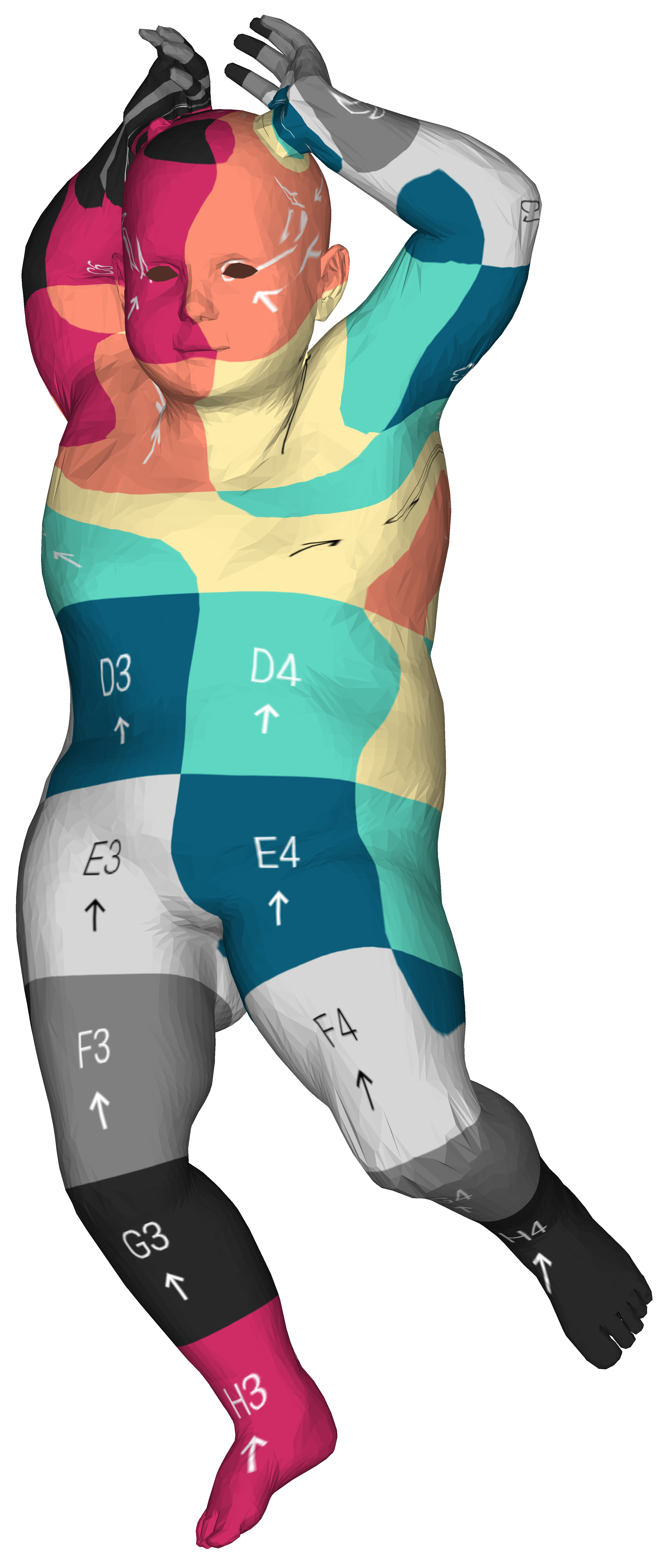}&
    \hspace{\hspaceCols}
    \includegraphics[height=\heightT]{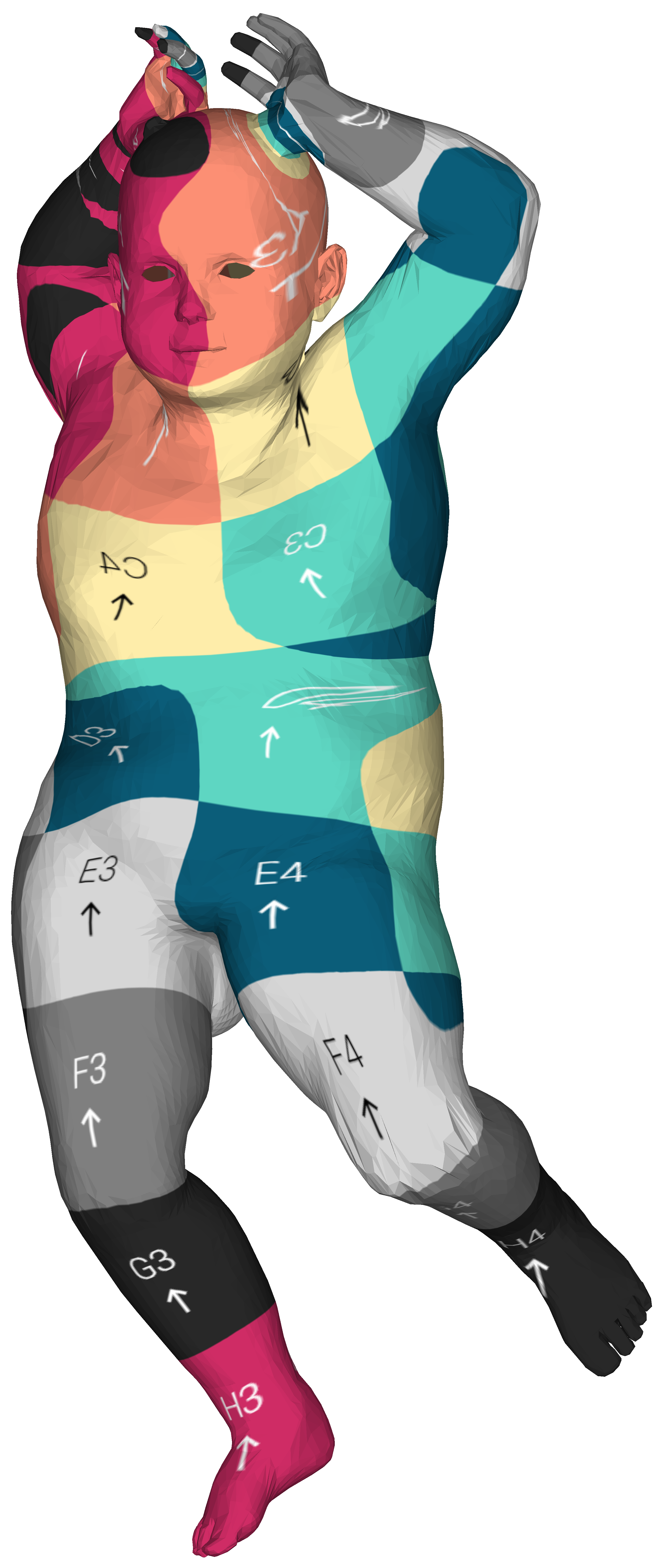}&
    \hspace{\hspaceCols}
    \includegraphics[height=\heightT]{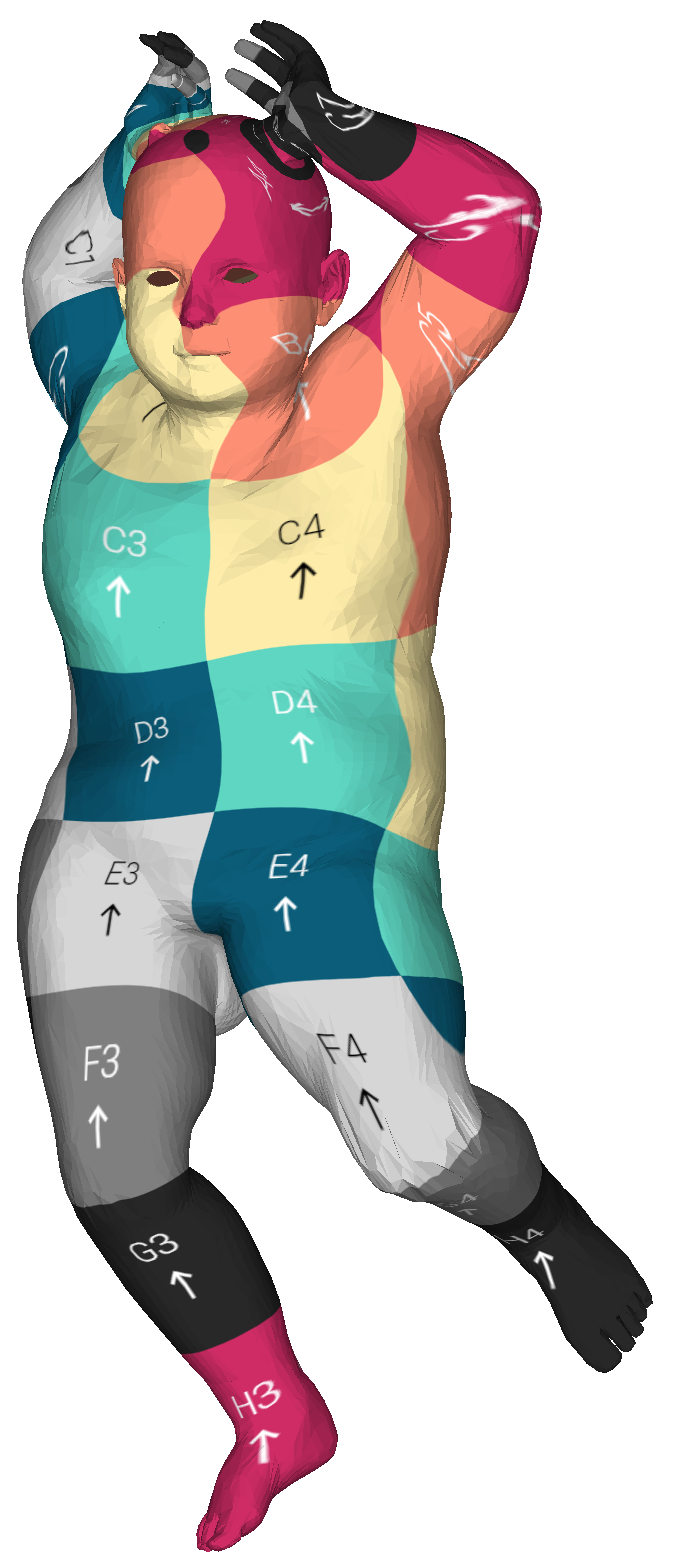}
    \\[\hspaceRows]
    \vspace{\wspaceRows}
    \hspace{\hspaceCols}
    \includegraphics[height=\heightT]{\pathOurs\topkidSrc}&
    \hspace{\hspaceCols}
    \includegraphics[height=\heightT]{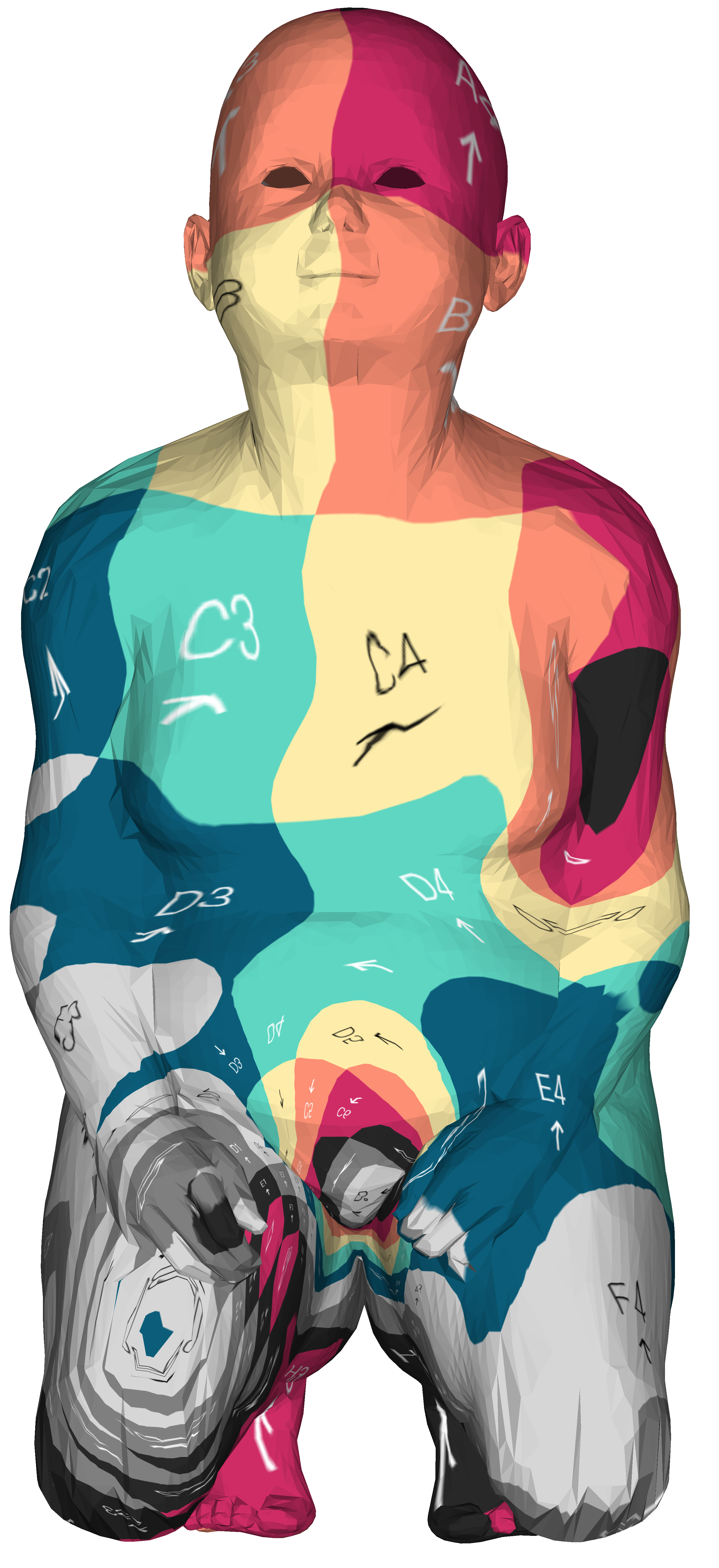}&
    \hspace{\hspaceCols}
    \includegraphics[height=\heightT]{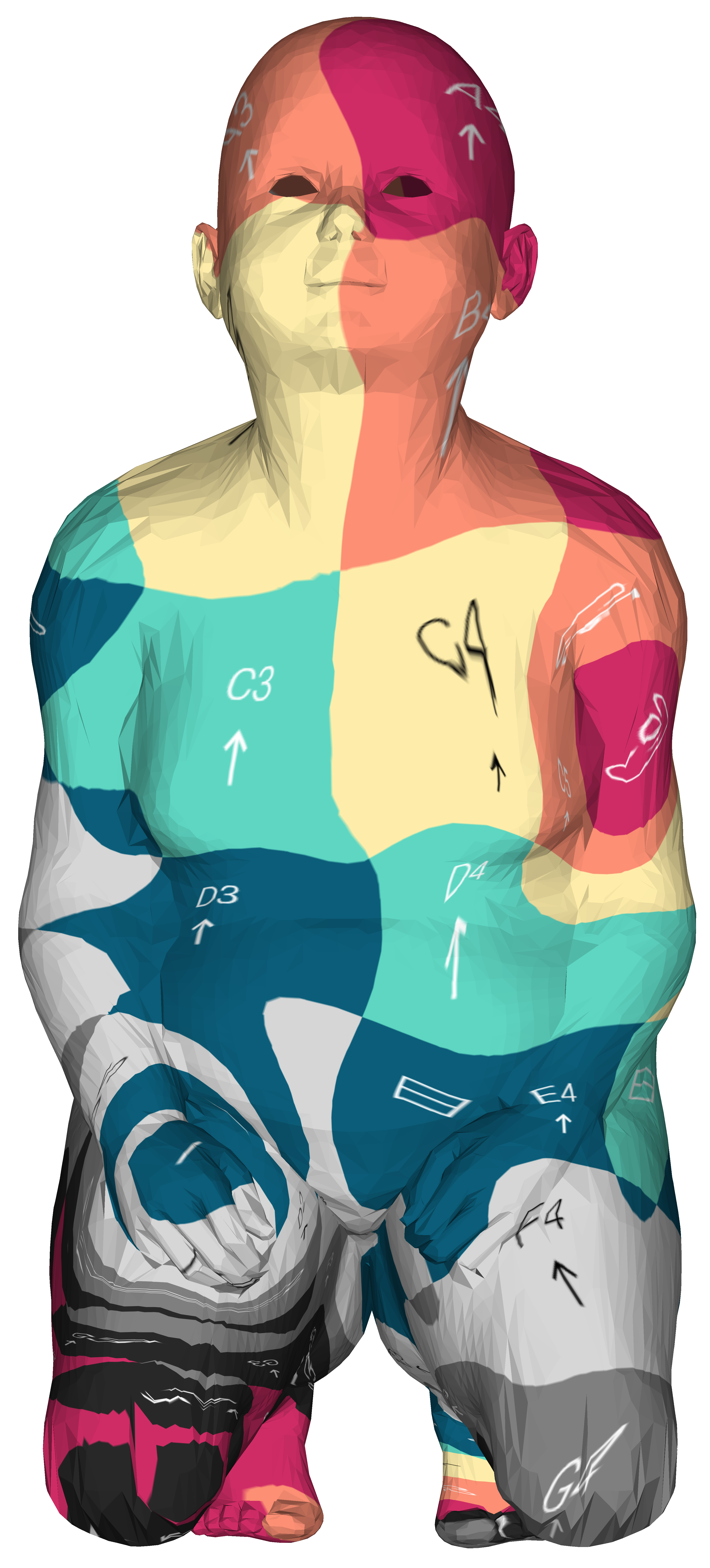}&
    \hspace{\hspaceCols}
    \includegraphics[height=\heightT]{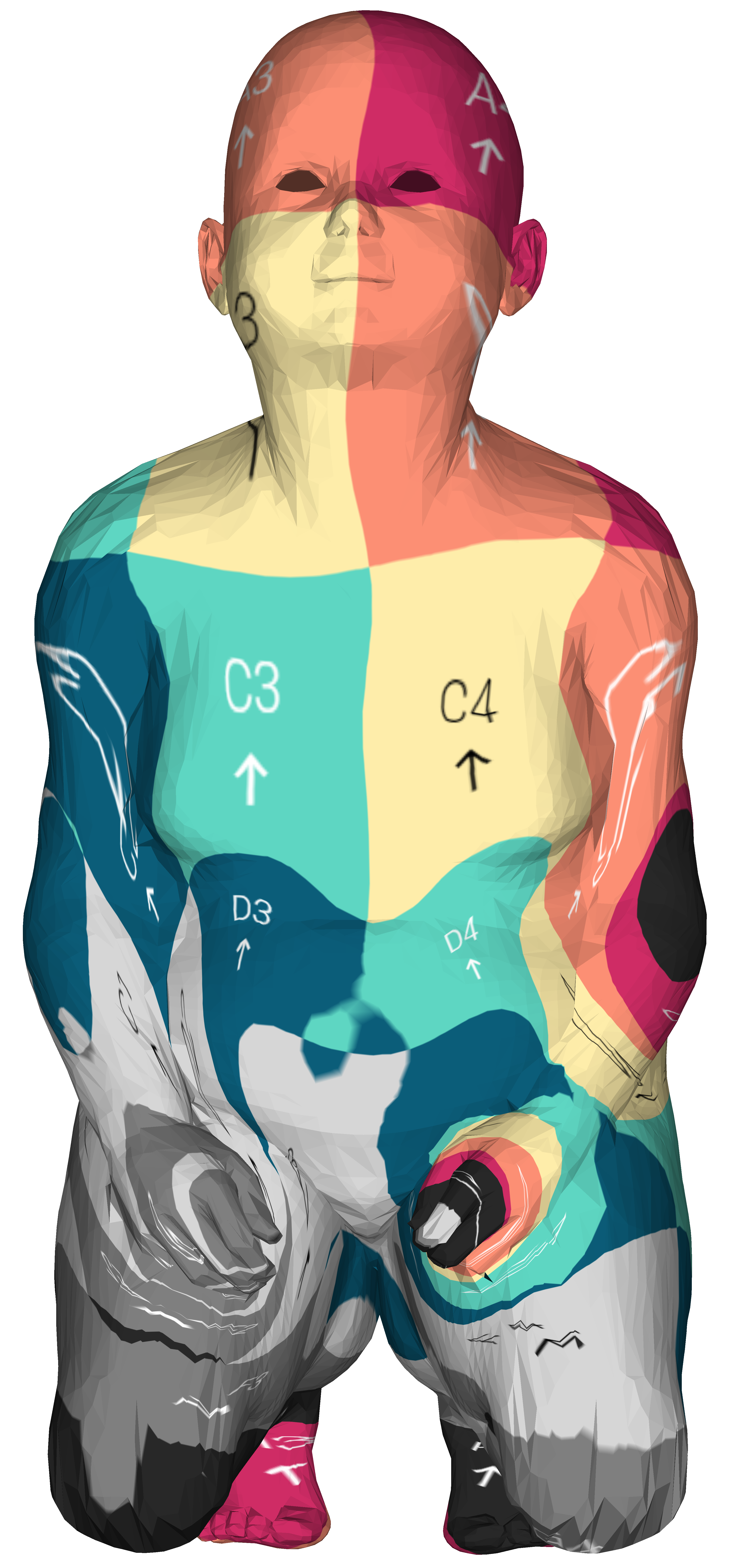}
\end{tabular}
} \\
\end{tabular}
\caption{
\textbf{Left: } Quantitative comparison with state-of-the-art methods on TOPKIDS under topological noise. \textbf{Right: } Qualitative comparison on TOPKIDS; our method produces more accurate, coherent correspondences than existing approaches.
}
\label{tab:topkids}
\end{figure}

\subsection{Matching with Topological Noise}
\subsubsection{Datasets}
Real-world scans often exhibit self-intersections and local geometric artifacts, leading to degraded mesh topology. Such topological noise distorts the intrinsic geometric structure and poses substantial challenges to functional map-based methods, which rely on stable Laplace-Beltrami operators. To assess robustness under severe topological perturbations, we conduct experiments on the TOPKIDS dataset~\cite{lahner2016shrec}. Owing to its limited training set (26 shapes), the comparison is restricted to axiomatic and unsupervised methods.

\subsubsection{Results}
As shown in \cref{tab:topkids}, our method achieves the best overall performance, yielding a 42\% improvement in mean geodesic error. The PCK curves in \cref{fig:noniso-pck} (right) further illustrate this advantage. This can be attributed to semantic features, which provide complementary high-level structural cues when intrinsic geometric descriptors become unreliable.

\subsection{Map Smoothness}
While average geodesic error quantifies overall correspondence accuracy, it does not explicitly capture the local smoothness of the point-wise maps. We therefore evaluate map smoothness using complementary metrics: conformal distortion~\cite{hormann2000mips}, surface coverage, bijectivity~\cite{ren2018continuous}, and Dirichlet energy~\cite{magnet2022smooth}.

\begin{table}[t]
\centering
\small
\caption{Map smoothness comparison with HybridFMap. We report conformal distortion, surface coverage, Dirichlet energy, and bijectivity to evaluate map smoothness beyond point-wise geodesic accuracy.}
\label{tab:smoothness_metrics}
\setlength{\tabcolsep}{5pt}
\resizebox{\linewidth}{!}{
\begin{tabular}{llcccccc}
\toprule
\textbf{Metric} & \textbf{Method} & \textbf{FAUST} & \textbf{SCAPE} & \textbf{SHREC'19} & \textbf{SMAL} & \textbf{DT4D-inter} & \textbf{TOPKIDS} \\
\midrule
\multirow{2}{*}{Conf. $\downarrow$} & HybridFMap & 0.654 & 0.779 & 1.030 & 2.473 & 1.866 & 3.201 \\
                                    & Ours       & \textbf{0.639} & \textbf{0.764} & \textbf{0.787} & \textbf{1.956} & \textbf{1.563} & \textbf{2.517} \\
\midrule
\multirow{2}{*}{Cov. (\%) $\uparrow$} & HybridFMap & 83.3 & \textbf{82.5} & 75.9 & 65.0 & \textbf{67.1} & 63.3 \\
                                      & Ours       & \textbf{83.5} & 81.9 & \textbf{76.1} & \textbf{71.2} & 66.3 & \textbf{63.9} \\
\midrule
\multirow{2}{*}{Dir.E $\downarrow$} & HybridFMap & 2.96 & \textbf{3.11} & 12.30 & 19.29 & \textbf{8.44} & 98.30 \\
                                   & Ours       & \textbf{2.95} & 3.27 & \textbf{6.30} & \textbf{10.20} & 8.70 & \textbf{48.33} \\
\midrule
\multirow{2}{*}{Bij. $\downarrow$} & HybridFMap & \textbf{0.0049} & \textbf{0.0060} & -- & 0.0285 & \textbf{0.0130} & -- \\
                                  & Ours       & 0.0052 & 0.0064 & -- & \textbf{0.0175} & 0.0166 & -- \\
\bottomrule
\end{tabular}
}
\end{table}

\begin{figure}[t]
\centering
\setlength{\tabcolsep}{1pt}
\begin{tabular}{@{}ccc@{}}
\resizebox{0.33\linewidth}{!}{\begingroup\newcommand{\pckLineWidth}{1pt}
\newcommand{\plotWidth}{\columnwidth}
\newcommand{\plotHeight}{0.7\columnwidth}
\newcommand{\pckTitle}{\textbf{SMAL}}
\definecolor{cPLOT0}{RGB}{28,213,227}
\definecolor{cPLOT1}{RGB}{80,150,80}
\definecolor{cPLOT2}{RGB}{90,130,213}
\definecolor{cPLOT3}{RGB}{247,179,43}
\definecolor{cPLOT4}{RGB}{124,42,43}
\definecolor{cPLOT5}{RGB}{242,64,0}

\pgfplotsset{%
    label style = {font=\LARGE},
    tick label style = {font=\large},
    title style =  {font=\LARGE},
    legend style={  fill= gray!10,
                    fill opacity=0.6, 
                    font=\large,
                    draw=gray!20, %
                    text opacity=1}
}
\begin{tikzpicture}[scale=0.55, transform shape]
	\begin{axis}[
        trim axis left,
        trim axis right,
		width=\plotWidth,
		height=\plotHeight,
		grid=major,
		title=\pckTitle,
		legend style={
			at={(0.97,0.03)},
			anchor=south east,
			legend columns=1},
		legend cell align={left},
        xlabel={\LARGE Conformal distortion},
		xmin=0,
        xmax=2.0,
        ylabel near ticks,
        xtick={0, 0.5, 1.0, 1.5, 2.0},
	    ymin=0,
        ymax=1,
        ytick={0, 0.2, 0.40, 0.60, 0.80, 1.0}
	]

\addplot [color=cPLOT1, smooth, line width=\pckLineWidth]
table[row sep=crcr]{%
0.0 0.07367435097694397 \\
0.02531645569620253 0.098990797996521 \\
0.05063291139240506 0.12292428314685822 \\
0.0759493670886076 0.14593961834907532 \\
0.10126582278481013 0.16780555248260498 \\
0.12658227848101267 0.1888226866722107 \\
0.1518987341772152 0.2090321183204651 \\
0.17721518987341772 0.22849170863628387 \\
0.20253164556962025 0.24726174771785736 \\
0.22784810126582278 0.2652311325073242 \\
0.25316455696202533 0.2824249565601349 \\
0.27848101265822783 0.29884764552116394 \\
0.3037974683544304 0.31485122442245483 \\
0.3291139240506329 0.32988566160202026 \\
0.35443037974683544 0.3444073498249054 \\
0.37974683544303794 0.35851234197616577 \\
0.4050632911392405 0.3722456991672516 \\
0.43037974683544306 0.38542959094047546 \\
0.45569620253164556 0.3982655107975006 \\
0.4810126582278481 0.41056036949157715 \\
0.5063291139240507 0.4223693907260895 \\
0.5316455696202531 0.43372008204460144 \\
0.5569620253164557 0.4448079764842987 \\
0.5822784810126582 0.45542842149734497 \\
0.6075949367088608 0.4658091068267822 \\
0.6329113924050633 0.4760783910751343 \\
0.6582278481012658 0.4856971204280853 \\
0.6835443037974683 0.49531853199005127 \\
0.7088607594936709 0.5044764280319214 \\
0.7341772151898734 0.5132884979248047 \\
0.7594936708860759 0.5219845771789551 \\
0.7848101265822784 0.5304487347602844 \\
0.810126582278481 0.5386607050895691 \\
0.8354430379746836 0.546726644039154 \\
0.8607594936708861 0.5544946789741516 \\
0.8860759493670886 0.5619748830795288 \\
0.9113924050632911 0.5692061185836792 \\
0.9367088607594937 0.5763577818870544 \\
0.9620253164556962 0.5833150148391724 \\
0.9873417721518987 0.5899734497070312 \\
1.0126582278481013 0.5965153574943542 \\
1.0379746835443038 0.6028447151184082 \\
1.0632911392405062 0.6090278625488281 \\
1.0886075949367089 0.6150269508361816 \\
1.1139240506329113 0.620856523513794 \\
1.139240506329114 0.6266254782676697 \\
1.1645569620253164 0.6320739984512329 \\
1.1898734177215189 0.6373612284660339 \\
1.2151898734177216 0.6425868272781372 \\
1.240506329113924 0.6477847695350647 \\
1.2658227848101267 0.6528101563453674 \\
1.2911392405063291 0.6577924489974976 \\
1.3164556962025316 0.6625968813896179 \\
1.3417721518987342 0.6672878265380859 \\
1.3670886075949367 0.6719198822975159 \\
1.3924050632911391 0.676317036151886 \\
1.4177215189873418 0.6806814074516296 \\
1.4430379746835442 0.6848098039627075 \\
1.4683544303797469 0.6890336871147156 \\
1.4936708860759493 0.6931120157241821 \\
1.5189873417721518 0.6970555186271667 \\
1.5443037974683544 0.7010029554367065 \\
1.5696202531645569 0.7048618197441101 \\
1.5949367088607596 0.7085537910461426 \\
1.620253164556962 0.7122172713279724 \\
1.6455696202531644 0.7158241868019104 \\
1.6708860759493671 0.7192823886871338 \\
1.6962025316455696 0.7227402329444885 \\
1.7215189873417722 0.7261481881141663 \\
1.7468354430379747 0.7293304204940796 \\
1.7721518987341771 0.7324411273002625 \\
1.7974683544303798 0.7354795336723328 \\
1.8227848101265822 0.7385309338569641 \\
1.8481012658227849 0.7416505217552185 \\
1.8734177215189873 0.7446730136871338 \\
1.8987341772151898 0.7474775314331055 \\
1.9240506329113924 0.7502674460411072 \\
1.9493670886075949 0.7531158328056335 \\
1.9746835443037973 0.7558263540267944 \\
2.0 0.7584949135780334 \\
    };
\addlegendentry{\textcolor{black}{HybridFMap (AUC: {0.54})}}

\addplot [color=cPLOT5, smooth, line width=\pckLineWidth]
table[row sep=crcr]{%
0.0 0.0724034383893013 \\
0.02531645569620253 0.09933581203222275 \\
0.05063291139240506 0.1250745803117752 \\
0.0759493670886076 0.14970429241657257 \\
0.10126582278481013 0.17305448651313782 \\
0.12658227848101267 0.1954198032617569 \\
0.1518987341772152 0.21701496839523315 \\
0.17721518987341772 0.23768927156925201 \\
0.20253164556962025 0.2575952112674713 \\
0.22784810126582278 0.27650824189186096 \\
0.25316455696202533 0.2948586940765381 \\
0.27848101265822783 0.3121562898159027 \\
0.3037974683544304 0.3288375437259674 \\
0.3291139240506329 0.34496957063674927 \\
0.35443037974683544 0.3606016933917999 \\
0.37974683544303794 0.3755766749382019 \\
0.4050632911392405 0.39003080129623413 \\
0.43037974683544306 0.4039103090763092 \\
0.45569620253164556 0.41734206676483154 \\
0.4810126582278481 0.43002450466156006 \\
0.5063291139240507 0.44238197803497314 \\
0.5316455696202531 0.4541949927806854 \\
0.5569620253164557 0.4656226634979248 \\
0.5822784810126582 0.47668659687042236 \\
0.6075949367088608 0.48754191398620605 \\
0.6329113924050633 0.49806174635887146 \\
0.6582278481012658 0.5080471634864807 \\
0.6835443037974683 0.5177887678146362 \\
0.7088607594936709 0.5271121263504028 \\
0.7341772151898734 0.5362592935562134 \\
0.7594936708860759 0.545099675655365 \\
0.7848101265822784 0.5536145567893982 \\
0.810126582278481 0.561926007270813 \\
0.8354430379746836 0.5699108839035034 \\
0.8607594936708861 0.5779135823249817 \\
0.8860759493670886 0.585607647895813 \\
0.9113924050632911 0.5930662751197815 \\
0.9367088607594937 0.6001026630401611 \\
0.9620253164556962 0.6071873903274536 \\
0.9873417721518987 0.6139966249465942 \\
1.0126582278481013 0.6206568479537964 \\
1.0379746835443038 0.6269614696502686 \\
1.0632911392405062 0.6331392526626587 \\
1.0886075949367089 0.6391716003417969 \\
1.1139240506329113 0.6450938582420349 \\
1.139240506329114 0.6508686542510986 \\
1.1645569620253164 0.6563482284545898 \\
1.1898734177215189 0.6618244647979736 \\
1.2151898734177216 0.6671819090843201 \\
1.240506329113924 0.6722927689552307 \\
1.2658227848101267 0.6773631572723389 \\
1.2911392405063291 0.6821764707565308 \\
1.3164556962025316 0.6869850754737854 \\
1.3417721518987342 0.6916252970695496 \\
1.3670886075949367 0.6961800456047058 \\
1.3924050632911391 0.7006884813308716 \\
1.4177215189873418 0.7049587368965149 \\
1.4430379746835442 0.7093063592910767 \\
1.4683544303797469 0.7135393023490906 \\
1.4936708860759493 0.717659056186676 \\
1.5189873417721518 0.7216504216194153 \\
1.5443037974683544 0.7254043221473694 \\
1.5696202531645569 0.7291187047958374 \\
1.5949367088607596 0.7327741980552673 \\
1.620253164556962 0.7364317178726196 \\
1.6455696202531644 0.7399353384971619 \\
1.6708860759493671 0.7433377504348755 \\
1.6962025316455696 0.7466610074043274 \\
1.7215189873417722 0.749866247177124 \\
1.7468354430379747 0.753023624420166 \\
1.7721518987341771 0.756156861782074 \\
1.7974683544303798 0.7592369914054871 \\
1.8227848101265822 0.7622770071029663 \\
1.8481012658227849 0.7651816606521606 \\
1.8734177215189873 0.7681170105934143 \\
1.8987341772151898 0.7709223628044128 \\
1.9240506329113924 0.7737748026847839 \\
1.9493670886075949 0.7764970660209656 \\
1.9746835443037973 0.7792308926582336 \\
2.0 0.7818278670310974 \\
    };
\addlegendentry{\textcolor{black}{\textbf{Ours (AUC: 0.56)}}}

\end{axis}
\end{tikzpicture}\endgroup} &
\resizebox{0.33\linewidth}{!}{\begingroup\newcommand{\pckLineWidth}{1pt}
\newcommand{\plotWidth}{\columnwidth}
\newcommand{\plotHeight}{0.7\columnwidth}
\newcommand{\pckTitle}{\textbf{DT4D-H inter-class}}
\definecolor{cPLOT0}{RGB}{28,213,227}
\definecolor{cPLOT1}{RGB}{80,150,80}
\definecolor{cPLOT2}{RGB}{90,130,213}
\definecolor{cPLOT3}{RGB}{247,179,43}
\definecolor{cPLOT4}{RGB}{124,42,43}
\definecolor{cPLOT5}{RGB}{242,64,0}

\pgfplotsset{%
    label style = {font=\LARGE},
    tick label style = {font=\large},
    title style =  {font=\LARGE},
    legend style={  fill= gray!10,
                    fill opacity=0.6, 
                    font=\large,
                    draw=gray!20, %
                    text opacity=1}
}
\begin{tikzpicture}[scale=0.55, transform shape]
	\begin{axis}[
        trim axis left,
        trim axis right,
		width=\plotWidth,
		height=\plotHeight,
		grid=major,
		title=\pckTitle,
		legend style={
			at={(0.97,0.03)},
			anchor=south east,
			legend columns=1},
		legend cell align={left},
        xlabel={\LARGE Conformal distortion},
		xmin=0,
        xmax=2.0,
        ylabel near ticks,
        xtick={0, 0.5, 1.0, 1.5, 2.0},
	    ymin=0,
        ymax=1,
        ytick={0, 0.2, 0.40, 0.60, 0.80, 1.0}
	]

\addplot [color=cPLOT1, smooth, line width=\pckLineWidth]
table[row sep=crcr]{%
0.0 0.0 \\
0.02531645569620253 0.03274094685912132 \\
0.05063291139240506 0.06374422460794449 \\
0.0759493670886076 0.09330937266349792 \\
0.10126582278481013 0.12139475345611572 \\
0.12658227848101267 0.14795146882534027 \\
0.1518987341772152 0.17335833609104156 \\
0.17721518987341772 0.19763065874576569 \\
0.20253164556962025 0.22090469300746918 \\
0.22784810126582278 0.2430759072303772 \\
0.25316455696202533 0.2642838656902313 \\
0.27848101265822783 0.2844858169555664 \\
0.3037974683544304 0.30385127663612366 \\
0.3291139240506329 0.3224237263202667 \\
0.35443037974683544 0.34018298983573914 \\
0.37974683544303794 0.35718223452568054 \\
0.4050632911392405 0.3735203742980957 \\
0.43037974683544306 0.3892887532711029 \\
0.45569620253164556 0.40444496273994446 \\
0.4810126582278481 0.4191051125526428 \\
0.5063291139240507 0.43304455280303955 \\
0.5316455696202531 0.4465591609477997 \\
0.5569620253164557 0.4595739543437958 \\
0.5822784810126582 0.4722025990486145 \\
0.6075949367088608 0.4843846261501312 \\
0.6329113924050633 0.4960203468799591 \\
0.6582278481012658 0.5073056221008301 \\
0.6835443037974683 0.5182291865348816 \\
0.7088607594936709 0.5288000106811523 \\
0.7341772151898734 0.5390540957450867 \\
0.7594936708860759 0.5488771796226501 \\
0.7848101265822784 0.5583542585372925 \\
0.810126582278481 0.5674586892127991 \\
0.8354430379746836 0.5763702392578125 \\
0.8607594936708861 0.5849947929382324 \\
0.8860759493670886 0.5933849215507507 \\
0.9113924050632911 0.6014697551727295 \\
0.9367088607594937 0.6093775033950806 \\
0.9620253164556962 0.616991400718689 \\
0.9873417721518987 0.6243941187858582 \\
1.0126582278481013 0.6315865516662598 \\
1.0379746835443038 0.6385625004768372 \\
1.0632911392405062 0.645313560962677 \\
1.0886075949367089 0.6516879796981812 \\
1.1139240506329113 0.6578909158706665 \\
1.139240506329114 0.6639841794967651 \\
1.1645569620253164 0.6698868870735168 \\
1.1898734177215189 0.675571858882904 \\
1.2151898734177216 0.6811424493789673 \\
1.240506329113924 0.6865143179893494 \\
1.2658227848101267 0.6917683482170105 \\
1.2911392405063291 0.6968950033187866 \\
1.3164556962025316 0.7018701434135437 \\
1.3417721518987342 0.706744909286499 \\
1.3670886075949367 0.7114332318305969 \\
1.3924050632911391 0.7160029411315918 \\
1.4177215189873418 0.7204685807228088 \\
1.4430379746835442 0.7248235940933228 \\
1.4683544303797469 0.7290646433830261 \\
1.4936708860759493 0.7332406044006348 \\
1.5189873417721518 0.7373181581497192 \\
1.5443037974683544 0.7412675619125366 \\
1.5696202531645569 0.7451431155204773 \\
1.5949367088607596 0.7489181160926819 \\
1.620253164556962 0.7525299787521362 \\
1.6455696202531644 0.7561229467391968 \\
1.6708860759493671 0.7596269249916077 \\
1.6962025316455696 0.7630133032798767 \\
1.7215189873417722 0.7663495540618896 \\
1.7468354430379747 0.769608736038208 \\
1.7721518987341771 0.7727413177490234 \\
1.7974683544303798 0.7758277058601379 \\
1.8227848101265822 0.7788580060005188 \\
1.8481012658227849 0.7818282842636108 \\
1.8734177215189873 0.7847446799278259 \\
1.8987341772151898 0.787563145160675 \\
1.9240506329113924 0.7903578281402588 \\
1.9493670886075949 0.7930542230606079 \\
1.9746835443037973 0.7957128286361694 \\
2.0 0.7982920408248901 \\
    };
\addlegendentry{\textcolor{black}{HybridFMap (AUC: 0.56)}}

\addplot [color=cPLOT5, smooth, line width=\pckLineWidth]
table[row sep=crcr]{%
0.0 1.916344842811668e-07 \\
0.02531645569620253 0.036707863211631775 \\
0.05063291139240506 0.07125509530305862 \\
0.0759493670886076 0.10389871150255203 \\
0.10126582278481013 0.13489095866680145 \\
0.12658227848101267 0.16438689827919006 \\
0.1518987341772152 0.19253626465797424 \\
0.17721518987341772 0.21931935846805573 \\
0.20253164556962025 0.24482186138629913 \\
0.22784810126582278 0.2690995931625366 \\
0.25316455696202533 0.2922584116458893 \\
0.27848101265822783 0.31434890627861023 \\
0.3037974683544304 0.33533087372779846 \\
0.3291139240506329 0.35537227988243103 \\
0.35443037974683544 0.3747132420539856 \\
0.37974683544303794 0.39312899112701416 \\
0.4050632911392405 0.4106765389442444 \\
0.43037974683544306 0.4274725317955017 \\
0.45569620253164556 0.4436118006706238 \\
0.4810126582278481 0.45913776755332947 \\
0.5063291139240507 0.47401848435401917 \\
0.5316455696202531 0.4883117973804474 \\
0.5569620253164557 0.5019951462745667 \\
0.5822784810126582 0.5151545405387878 \\
0.6075949367088608 0.5277687907218933 \\
0.6329113924050633 0.5399959683418274 \\
0.6582278481012658 0.5517216324806213 \\
0.6835443037974683 0.5630841255187988 \\
0.7088607594936709 0.5738774538040161 \\
0.7341772151898734 0.5843684077262878 \\
0.7594936708860759 0.5945112705230713 \\
0.7848101265822784 0.6042524576187134 \\
0.810126582278481 0.6136153936386108 \\
0.8354430379746836 0.6227044463157654 \\
0.8607594936708861 0.6314501166343689 \\
0.8860759493670886 0.6397684216499329 \\
0.9113924050632911 0.6477981209754944 \\
0.9367088607594937 0.6554968953132629 \\
0.9620253164556962 0.6629692912101746 \\
0.9873417721518987 0.6702369451522827 \\
1.0126582278481013 0.6772509813308716 \\
1.0379746835443038 0.684097170829773 \\
1.0632911392405062 0.6906816959381104 \\
1.0886075949367089 0.6970273852348328 \\
1.1139240506329113 0.703258216381073 \\
1.139240506329114 0.7092229723930359 \\
1.1645569620253164 0.7150229215621948 \\
1.1898734177215189 0.7206273674964905 \\
1.2151898734177216 0.7260286211967468 \\
1.240506329113924 0.7313198447227478 \\
1.2658227848101267 0.7364164590835571 \\
1.2911392405063291 0.741336464881897 \\
1.3164556962025316 0.7461447715759277 \\
1.3417721518987342 0.750796377658844 \\
1.3670886075949367 0.7553399205207825 \\
1.3924050632911391 0.7597207427024841 \\
1.4177215189873418 0.7640394568443298 \\
1.4430379746835442 0.7682652473449707 \\
1.4683544303797469 0.7722907066345215 \\
1.4936708860759493 0.7762078046798706 \\
1.5189873417721518 0.780049741268158 \\
1.5443037974683544 0.783757746219635 \\
1.5696202531645569 0.787360429763794 \\
1.5949367088607596 0.7909228205680847 \\
1.620253164556962 0.794363260269165 \\
1.6455696202531644 0.797741711139679 \\
1.6708860759493671 0.8009899854660034 \\
1.6962025316455696 0.8041825890541077 \\
1.7215189873417722 0.8073240518569946 \\
1.7468354430379747 0.8103232383728027 \\
1.7721518987341771 0.8132795691490173 \\
1.7974683544303798 0.8161539435386658 \\
1.8227848101265822 0.8189792633056641 \\
1.8481012658227849 0.821704626083374 \\
1.8734177215189873 0.8243488669395447 \\
1.8987341772151898 0.8269519209861755 \\
1.9240506329113924 0.8294872641563416 \\
1.9493670886075949 0.8319876790046692 \\
1.9746835443037973 0.834398627281189 \\
2.0 0.8367895483970642 \\
    };
\addlegendentry{\textcolor{black}{\textbf{Ours: (AUC: {0.60})}}}

\end{axis}
\end{tikzpicture}\endgroup} &
\resizebox{0.33\linewidth}{!}{\begingroup\newcommand{\pckLineWidth}{1pt}
\newcommand{\plotWidth}{\columnwidth}
\newcommand{\plotHeight}{0.7\columnwidth}
\newcommand{\pckTitle}{\textbf{TOPKIDS}}
\definecolor{cPLOT0}{RGB}{28,213,227}
\definecolor{cPLOT1}{RGB}{80,150,80}
\definecolor{cPLOT2}{RGB}{90,130,213}
\definecolor{cPLOT3}{RGB}{247,179,43}
\definecolor{cPLOT4}{RGB}{124,42,43}
\definecolor{cPLOT5}{RGB}{242,64,0}

\pgfplotsset{%
    label style = {font=\LARGE},
    tick label style = {font=\large},
    title style =  {font=\LARGE},
    legend style={  fill= gray!10,
                    fill opacity=0.6, 
                    font=\large,
                    draw=gray!20, %
                    text opacity=1}
}
\begin{tikzpicture}[scale=0.55, transform shape]
	\begin{axis}[
        trim axis left,
        trim axis right,
		width=\plotWidth,
		height=\plotHeight,
		grid=major,
		title=\pckTitle,
		legend style={
			at={(0.97,0.03)},
			anchor=south east,
			legend columns=1},
		legend cell align={left},
        xlabel={\LARGE Conformal distortion},
		xmin=0,
        xmax=2.0,
        ylabel near ticks,
        xtick={0, 0.5, 1.0, 1.5, 2.0},
	    ymin=0,
        ymax=1,
        ytick={0, 0.2, 0.40, 0.60, 0.80, 1.0}
	]
\addplot [color=cPLOT1, smooth, line width=\pckLineWidth]
table[row sep=crcr]{%
0.0 0.004938903730362654 \\
0.02531645569620253 0.1757378876209259 \\
0.05063291139240506 0.21181704103946686 \\
0.0759493670886076 0.23938456177711487 \\
0.10126582278481013 0.2637030780315399 \\
0.12658227848101267 0.2849236726760864 \\
0.1518987341772152 0.3049609363079071 \\
0.17721518987341772 0.3233189284801483 \\
0.20253164556962025 0.3407827615737915 \\
0.22784810126582278 0.35687655210494995 \\
0.25316455696202533 0.37234053015708923 \\
0.27848101265822783 0.38740357756614685 \\
0.3037974683544304 0.4010189175605774 \\
0.3291139240506329 0.4150642454624176 \\
0.35443037974683544 0.42787882685661316 \\
0.37974683544303794 0.4408748745918274 \\
0.4050632911392405 0.45291849970817566 \\
0.43037974683544306 0.4649893641471863 \\
0.45569620253164556 0.4765799343585968 \\
0.4810126582278481 0.4876977503299713 \\
0.5063291139240507 0.49844157695770264 \\
0.5316455696202531 0.5088657140731812 \\
0.5569620253164557 0.5186998844146729 \\
0.5822784810126582 0.5283938646316528 \\
0.6075949367088608 0.5376816987991333 \\
0.6329113924050633 0.5470530390739441 \\
0.6582278481012658 0.5557247400283813 \\
0.6835443037974683 0.5638969540596008 \\
0.7088607594936709 0.5719779133796692 \\
0.7341772151898734 0.57987380027771 \\
0.7594936708860759 0.5870233178138733 \\
0.7848101265822784 0.5939636826515198 \\
0.810126582278481 0.6009550094604492 \\
0.8354430379746836 0.6080524921417236 \\
0.8607594936708861 0.6146326661109924 \\
0.8860759493670886 0.620901882648468 \\
0.9113924050632911 0.627285361289978 \\
0.9367088607594937 0.6332082748413086 \\
0.9620253164556962 0.6391209959983826 \\
0.9873417721518987 0.6449318528175354 \\
1.0126582278481013 0.6506702899932861 \\
1.0379746835443038 0.6559750437736511 \\
1.0632911392405062 0.6612461805343628 \\
1.0886075949367089 0.6664786338806152 \\
1.1139240506329113 0.671680748462677 \\
1.139240506329114 0.6764622926712036 \\
1.1645569620253164 0.681257963180542 \\
1.1898734177215189 0.6858209371566772 \\
1.2151898734177216 0.6904473304748535 \\
1.240506329113924 0.6950247287750244 \\
1.2658227848101267 0.6993535161018372 \\
1.2911392405063291 0.70381760597229 \\
1.3164556962025316 0.7080004811286926 \\
1.3417721518987342 0.7121146321296692 \\
1.3670886075949367 0.715890109539032 \\
1.3924050632911391 0.7195627093315125 \\
1.4177215189873418 0.7237663865089417 \\
1.4430379746835442 0.7274284362792969 \\
1.4683544303797469 0.7309391498565674 \\
1.4936708860759493 0.7344943284988403 \\
1.5189873417721518 0.7380553483963013 \\
1.5443037974683544 0.7415000200271606 \\
1.5696202531645569 0.7446002960205078 \\
1.5949367088607596 0.7478808760643005 \\
1.620253164556962 0.7510770559310913 \\
1.6455696202531644 0.7538937330245972 \\
1.6708860759493671 0.7568690180778503 \\
1.6962025316455696 0.7596294283866882 \\
1.7215189873417722 0.7624660134315491 \\
1.7468354430379747 0.7651635408401489 \\
1.7721518987341771 0.7681592702865601 \\
1.7974683544303798 0.7708362340927124 \\
1.8227848101265822 0.7734546065330505 \\
1.8481012658227849 0.7760469913482666 \\
1.8734177215189873 0.7784927487373352 \\
1.8987341772151898 0.780979573726654 \\
1.9240506329113924 0.7833769917488098 \\
1.9493670886075949 0.7855141162872314 \\
1.9746835443037973 0.7878464460372925 \\
2.0 0.7900725603103638 \\
    };
\addlegendentry{\textcolor{black}{HybridFMap (AUC: {0.59})}}

\addplot [color=cPLOT5, smooth, line width=\pckLineWidth]
table[row sep=crcr]{%
0.0 0.0034060648176819086 \\
0.02531645569620253 0.13011816143989563 \\
0.05063291139240506 0.1694672405719757 \\
0.0759493670886076 0.2028014361858368 \\
0.10126582278481013 0.23265673220157623 \\
0.12658227848101267 0.25901302695274353 \\
0.1518987341772152 0.28368282318115234 \\
0.17721518987341772 0.3063332736492157 \\
0.20253164556962025 0.3284175992012024 \\
0.22784810126582278 0.34913453459739685 \\
0.25316455696202533 0.3686974346637726 \\
0.27848101265822783 0.3870926797389984 \\
0.3037974683544304 0.4040910601615906 \\
0.3291139240506329 0.4208691716194153 \\
0.35443037974683544 0.436421662569046 \\
0.37974683544303794 0.4515388011932373 \\
0.4050632911392405 0.4663015604019165 \\
0.43037974683544306 0.48024025559425354 \\
0.45569620253164556 0.49334394931793213 \\
0.4810126582278481 0.5060240626335144 \\
0.5063291139240507 0.5183156132698059 \\
0.5316455696202531 0.5301357507705688 \\
0.5569620253164557 0.5413326025009155 \\
0.5822784810126582 0.5522223114967346 \\
0.6075949367088608 0.5620368719100952 \\
0.6329113924050633 0.5721414089202881 \\
0.6582278481012658 0.5819132924079895 \\
0.6835443037974683 0.5906693339347839 \\
0.7088607594936709 0.5995997786521912 \\
0.7341772151898734 0.6080589294433594 \\
0.7594936708860759 0.61612868309021 \\
0.7848101265822784 0.6240532994270325 \\
0.810126582278481 0.6314826011657715 \\
0.8354430379746836 0.6390403509140015 \\
0.8607594936708861 0.6462010145187378 \\
0.8860759493670886 0.6530924439430237 \\
0.9113924050632911 0.6599744558334351 \\
0.9367088607594937 0.6661155819892883 \\
0.9620253164556962 0.6726237535476685 \\
0.9873417721518987 0.6784862875938416 \\
1.0126582278481013 0.6843080520629883 \\
1.0379746835443038 0.6899078488349915 \\
1.0632911392405062 0.6956289410591125 \\
1.0886075949367089 0.7010449171066284 \\
1.1139240506329113 0.7063441276550293 \\
1.139240506329114 0.711726188659668 \\
1.1645569620253164 0.7166305184364319 \\
1.1898734177215189 0.7211223840713501 \\
1.2151898734177216 0.7258199453353882 \\
1.240506329113924 0.7302731275558472 \\
1.2658227848101267 0.7345548868179321 \\
1.2911392405063291 0.7389596700668335 \\
1.3164556962025316 0.7428778409957886 \\
1.3417721518987342 0.746584415435791 \\
1.3670886075949367 0.7506034970283508 \\
1.3924050632911391 0.7543914318084717 \\
1.4177215189873418 0.7579189538955688 \\
1.4430379746835442 0.7615126371383667 \\
1.4683544303797469 0.7649039030075073 \\
1.4936708860759493 0.7685193419456482 \\
1.5189873417721518 0.7717331051826477 \\
1.5443037974683544 0.775077223777771 \\
1.5696202531645569 0.7780293822288513 \\
1.5949367088607596 0.7809998989105225 \\
1.620253164556962 0.7840025424957275 \\
1.6455696202531644 0.7869120240211487 \\
1.6708860759493671 0.789473295211792 \\
1.6962025316455696 0.7922406792640686 \\
1.7215189873417722 0.7949410080909729 \\
1.7468354430379747 0.7976735830307007 \\
1.7721518987341771 0.8002590537071228 \\
1.7974683544303798 0.8029475212097168 \\
1.8227848101265822 0.8053112030029297 \\
1.8481012658227849 0.8076750636100769 \\
1.8734177215189873 0.8099522590637207 \\
1.8987341772151898 0.812364399433136 \\
1.9240506329113924 0.8145538568496704 \\
1.9493670886075949 0.8168721199035645 \\
1.9746835443037973 0.8189916014671326 \\
2.0 0.8210799098014832 \\
    };
\addlegendentry{\textcolor{black}{\textbf{Ours (AUC: 0.62)}}}

\end{axis}
\end{tikzpicture}\endgroup}
\end{tabular}
\caption{\textbf{Conformal-distortion-based smoothness comparison.} Our method achieves competitive or lower average distortion, indicating smoother correspondences under non-isometric and topological-noise settings.}
\label{fig:smoothness}
\end{figure}
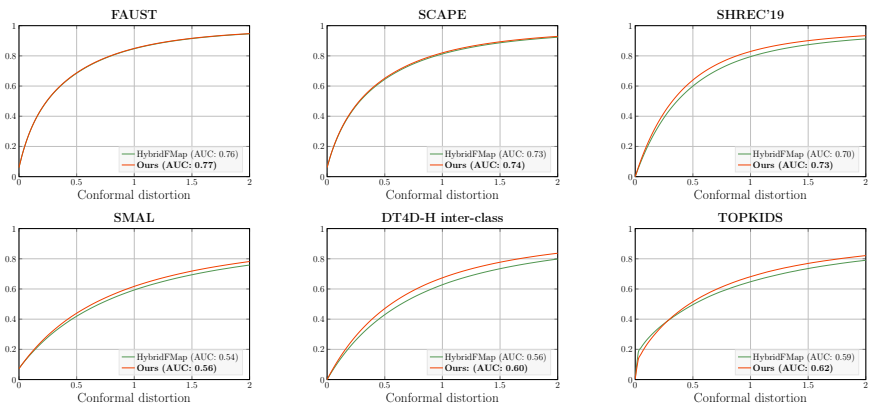

As shown in \cref{tab:smoothness_metrics}, SGMatch improves most smoothness metrics, with the clearest gains on challenging non-isometric and topology-noisy datasets. The conformal-distortion PCK curves in \cref{fig:smoothness} further show that these gains hold across distortion thresholds. On SMAL and TOPKIDS, where intrinsic geometry is less reliable, our method substantially reduces conformal distortion and Dirichlet energy while improving coverage. These results indicate that SGMatch improves not only point-wise accuracy but also local map smoothness. Some near-isometric metrics on SCAPE and DT4D-inter remain comparable rather than strictly better, suggesting that the main benefit appears under stronger ambiguity, deformation, or topological noise.

\subsection{Comparison with Alternative Smoothness Regularizers}
We further isolate the role of CFM by comparing it with alternative regularization objectives under the same SGMatch backbone. This comparison tests whether trajectory-level feature transport supervision is redundant with endpoint matching, cycle-consistency-style constraints~\cite{ren2019structured}, or synchronous-diffusion-style constraints~\cite{cao2024synchronous}.

\begin{table}[t]
\centering
\small
\caption{Comparison with alternative smoothness regularizers under the same SGMatch backbone.}
\label{tab:cfm_alternatives}
\setlength{\tabcolsep}{3pt}
\resizebox{\linewidth}{!}{
\begin{tabular}{l ccccc cccc}
\toprule
\multirow{2}{*}{\textbf{Variant}} &
\multicolumn{5}{c}{\textbf{SMAL}} &
\multicolumn{4}{c}{\textbf{TOPKIDS}} \\
\cmidrule(lr){2-6}\cmidrule(lr){7-10}
& \textbf{Geo. $\downarrow$} & \textbf{Conf. $\downarrow$} & \textbf{Dir.E $\downarrow$} & \textbf{Bij. $\downarrow$} & \textbf{Cov. $\uparrow$}
& \textbf{Geo. $\downarrow$} & \textbf{Conf. $\downarrow$} & \textbf{Dir.E $\downarrow$} & \textbf{Cov. $\uparrow$} \\
\midrule
w/o CFM           & 2.7 & 2.32 & 10.55 & 0.0210 & 66.8 & 3.4 & 2.92 & 45.64 & 63.5 \\
Endpoint-only     & 2.7 & 2.46 & 14.37 & 0.0225 & 65.0 & 3.4 & 2.76 & 50.92 & 63.8 \\
Cycle-consistency & 2.8 & 2.29 & 10.53 & 0.0201 & 67.2 & 3.4 & 2.75 & \textbf{42.42} & 63.7 \\
Sync. Diff.       & 2.6 & 2.45 & 13.48 & 0.0221 & 64.7 & 3.5 & 2.79 & 50.14 & 63.8 \\
\textbf{Ours}     & \textbf{2.5} & \textbf{1.96} & \textbf{10.20} & \textbf{0.0175} & \textbf{71.2} & \textbf{2.9} & \textbf{2.52} & 48.33 & \textbf{63.9} \\
\bottomrule
\end{tabular}
}
\end{table}

\begin{table}[t]
\centering
\small
\caption{Quantitative results of the ablation experiments on the SMAL dataset.}
\label{tab:ablation}
\setlength{\tabcolsep}{3pt}
\resizebox{\linewidth}{!}{
\begin{tabular}{lccccccc}
\toprule
\textbf{Variant} & \textbf{Geo.} & \textbf{Sem.} & \textbf{Gating} & \textbf{Attention} & \textbf{Heat Diff.} & \textbf{CFM} & \textbf{Geo.Err $\downarrow$} \\
\midrule
Geo. only                 & \checkmark & \ding{55}  & --         & --     & \checkmark & \checkmark & 3.2  \\
Sem. only                 & \ding{55}  & \checkmark & --         & --     & \checkmark & \checkmark & 21.2 \\
Gating only               & \checkmark & \checkmark & \checkmark & none   & \checkmark & \checkmark & 3.6  \\
Local attention only      & \checkmark & \checkmark & \ding{55}  & local  & \checkmark & \checkmark & 2.6  \\
Gating + global attention & \checkmark & \checkmark & \checkmark & global & \checkmark & \checkmark & 2.6  \\
w/o heat diffusion        & \checkmark & \checkmark & \checkmark & local  & \ding{55}  & \checkmark & 3.0  \\
\textbf{Ours}             & \checkmark & \checkmark & \checkmark & local  & \checkmark & \checkmark & \textbf{2.5}  \\
\bottomrule
\end{tabular}
}
\end{table}

Endpoint-only removes the velocity MLP and trajectory sampling, and instead directly matches the heat-diffused source features with the transported target features in both directions. This variant isolates whether the gain comes from trajectory-level supervision rather than endpoint feature matching alone. Cycle-consistency and Sync. Diff. provide time-free alternatives based on round-trip consistency and diffusion-style smoothness constraints. As shown in \cref{tab:cfm_alternatives}, CFM gives the best geodesic error and conformal distortion on both datasets, and improves most map regularity metrics on SMAL. This indicates that the trajectory-level transport objective is not redundant with endpoint matching or diffusion-based smoothing.

On TOPKIDS, cycle-consistency gives a lower Dirichlet energy than CFM, but it sacrifices geodesic accuracy, conformal distortion, and coverage. This suggests over-smoothing rather than a better correspondence, and supports using multiple regularity metrics rather than Dirichlet energy alone.

\subsection{Ablation Study}
After isolating the CFM regularizer above, we conduct ablation experiments on the SMAL dataset to evaluate the main components and the design of SGLCA.

Results in \cref{tab:ablation} show that both modalities are important: geometric features provide the structural basis, while semantic features alone are insufficient for accurate correspondence. The SGLCA variants further disentangle semantic gating from attention locality. Local attention provides the main gain by restricting semantic interactions to mesh neighborhoods, and semantic gating gives an additional improvement by allowing semantic cues to modulate geometric channels rather than replacing geometry. Global attention does not improve over local attention, suggesting that unrestricted semantic interactions can introduce irrelevant long-range responses. Finally, removing spectral heat diffusion degrades performance, confirming that stabilized feature endpoints complement the trajectory-level transport regularization analyzed above.

\section{Limitations}
Despite its effectiveness, SGMatch has two main limitations. First, the current framework is designed for full-shape correspondence and does not explicitly handle partial matching scenarios, which remain an important and actively studied problem in non-rigid shape analysis~\cite{xie2025echomatch, ehm2025beyond}. Second, its semantic features are derived from pretrained 2D visual models, so the performance of SGMatch depends on the quality and domain generalization capability of these external representations. Domains where such visual priors are weak or unavailable~\cite{bastian2023s3m} may therefore require domain-specific encoders or task-specific semantic pretraining.

\section{Conclusion}
We presented SGMatch for robust non-rigid shape correspondence under challenging deformations and topology changes. The results show that semantic guidance is most beneficial when geometric descriptors become ambiguous, while flow-based transport regularization improves the spatial coherence of point-wise recovery from truncated spectral representations. Experiments across near-isometric, non-isometric, and topology-noisy benchmarks demonstrate improved correspondence accuracy and map regularity, with the clearest gains under large deformation and topological noise.

\section*{Acknowledgements}
This work was supported by the National Key R\&D Program of China under Grant 2024YFE0202500, and the National Natural Science Foundation of China under Grants 624B2107 and U23B2050.

\bibliographystyle{splncs04}
\bibliography{main}

\newpage
\appendix

\section{Implementation Details}
\label{sec:supp_impl}

\subsection{Experimental Settings}
All learning-based methods are implemented using PyTorch 2.1.0 and CUDA 12.1, while axiomatic approaches are executed in MATLAB 2018a. All experiments are conducted on a single NVIDIA GeForce RTX 3090 GPU and an Intel Xeon(R) Platinum 8365A CPU (2.60GHz). Regarding geometric feature descriptors, we follow established practices and use WKS as geometric feature descriptors. For the SMAL dataset, we instead use raw XYZ coordinates of the 3D vertices and apply random rotation augmentation.

\subsection{Technical Details}
In the feature solver, the geometric feature dimension $D^g$ is 256, and the semantic feature dimension $D^s$ is 768. We adopt the pre-trained DINOv2-ViT-B/14 model. The number of viewpoints $\{\xi\}$ is set to 100, and the rendered image resolution is $H = 512$ and $W = 512$. We include normal maps to improve the representation of 3D shape details. For the SGLCA module, the attention embedding dimension $d$ is 256. Neighborhoods use a 1-ring structure with self-loops to maintain a fixed size of 32. In the functional map module, the eigenbasis dimensions are hybridly set to $k_{\mathrm{LB}}=140$ and $k_{\mathrm{Elas}}=60$ (see \cite{bastian2024hybrid}), and the temperature $\tau_T$ is set to 0.07. For CFM, the diffusion time $\tau$ in spectral heat diffusion varies by dataset. Following Cao et al.~\cite{cao2024synchronous}, we set $\tau$ to $10^{-2}$ for near-isometric cases, $10^{-4}$ for non-isometric datasets and $10^{-1}$ for topology-noisy datasets. Concurrently, the confidence sampling parameter $\alpha$ is set to 2, and the constant in the CFM loss is $10^{-3}$.

The loss weights are $\lambda_\mathrm{bij} = \lambda_\mathrm{orth} = \lambda_\mathrm{couple} = 1.0$ and $\lambda_\mathrm{cfm} = 100$. We train the model end-to-end using the Adam optimizer with a learning rate of 0.001 and a cosine annealing schedule.

\section{Additional Ablation Experiments}
\label{sec:supp_ablation}
\subsection{SGLCA Module Design}
We investigate alternative fusion strategies within the SGLCA module to validate architectural choices. Three variants are compared against our full design: (1) \textbf{element-wise addition}, where linear projection aligns feature dimensions before summation, treating modalities equally without channel modulation; (2) \textbf{raw concatenation}, which preserves cross-modal information but causes dimensional mismatch; and (3) \textbf{concatenation with MLP}, restoring original dimensions while lacking channel-wise selectivity. The main paper further disentangles semantic gating, local attention, and global attention in the unified ablation table.

\cref{tab:sglca} shows our full SGLCA design outperforms all alternatives. The gap between element-wise addition and our method indicates simple summation fails to capture geometry-semantic relationships, as it treats modalities uniformly regardless of local reliability. While concatenation variants improve upon addition, they remain inferior, suggesting information preservation is insufficient without fine-grained selectivity. Together with the main ablation table, these results show that semantic context is most effective when it modulates geometric channels through the proposed gated local attention design.

\begin{table}[!t]
\centering
\small
\caption{Ablation of feature fusion strategies and gating mechanism in the SGLCA module on the SMAL dataset.}
\label{tab:sglca}
\setlength{\tabcolsep}{2pt}
\begin{tabular}{lcccc}
\toprule
\textbf{Settings} & (1) Add & (2) Raw concat & (3) Concat+MLP & 
\textbf{Ours} \\
\midrule
\textbf{Geo. Err ($\times 100$)} & 5.3 & 3.5 & 3.6 & \textbf{2.5} \\
\bottomrule
\end{tabular}
\end{table}

\subsection{Conditional Flow Matching Regularization}
\label{subsec:cfm}

\begin{table}[!t]
\centering
\small
\caption{Comparison of CFM regularization variants on the SMAL dataset.}
\label{tab:cfm}
\setlength{\tabcolsep}{4pt}
\begin{tabular}{lccc}
\toprule
\textbf{Settings} & (1) MSE & (2) Laplacian & \textbf{Ours} \\
\midrule
\textbf{Geo. Err ($\times 100$)} & 2.6 & 4.1 & \textbf{2.5} \\
\bottomrule
\end{tabular}
\end{table}

We assess CFM regularization design choices by comparing two alternatives on SMAL: (1) \textbf{MSE loss.} Replacing the Charbonnier loss with a standard MSE objective:
\begin{equation}
\mathcal{L}_{\mathrm{mse}} = \mathbb{E}_{t, i \in \mathcal{S}} \left[\left| \mathbf{v}_{\theta}(\mathbf{z}_{t,i}, t) -\mathbf{v}_{\mathrm{target},i} \right|^2 \right].
\label{eq:l_cfm_mse}
\end{equation}
Although MSE and Charbonnier losses behave similarly under accurate correspondences, MSE is more sensitive to outliers from early-stage unreliable soft correspondences, potentially destabilizing optimization and hindering convergence.

(2) \textbf{Laplacian smoothing.} We replace CFM regularization with static Laplacian smoothing on transported target features:
\begin{equation}
\mathcal{L}_{\mathrm{lap}} = \mathrm{tr}(\mathbf{z}_1^{\top} \mathbf{L}_{\mathcal{X}} \mathbf{z}_1),
\label{eq:l_laplacian}
\end{equation}
where $\mathbf{z_1} = \mathbf{\Pi}_{\mathcal{XY}} \mathbf{Z}_{\mathcal{Y}}$ denotes transported target features and $\mathbf{L}_{\mathcal{X}}$ is the source shape's Laplace--Beltrami operator. This approach enforces smoothness as a static pairwise constraint rather than through dynamic trajectories, encouraging nearby source vertices to match similar target features.

\cref{tab:cfm} shows both alternatives reduce performance. The small gap between MSE and Charbonnier losses suggests CFM's primary benefit is trajectory-level regularization rather than the specific loss function, though Charbonnier loss adds noise robustness. The larger drop with Laplacian smoothing indicates static pairwise constraints cannot fully capture the dynamic feature transport required for non-isometric deformations. While Laplacian smoothing enforces final-state smoothness, it lacks transport process guidance, making optimization prone to locally inconsistent solutions a coherent velocity field prevents.

\subsection{Importance Sampling Strategy}
We evaluate the confidence-based sampling strategy used in the importance-weighted CFM objective (Eq.~\eqref{eq:l_cfm}), where vertices are sampled according to the confidence weights defined in Eq.~\eqref{eq:confidence_weight}. We compare it against a uniform sampling baseline ($w_i = 1\ \forall i$) with equal subset size $|\mathcal{S}|$.

As shown in \cref{tab:importance_sampling}, confidence-based sampling yields small but steady gains over uniform sampling. This limited gap is expected as the Charbonnier loss inherently handles outliers, partially mitigating the lack of weighting in the baseline. Nonetheless, confidence weighting provides further benefits by filtering unreliable transport targets during sampling, particularly in early training. 

\begin{figure}[t]
\centering
\includegraphics[width=\linewidth]{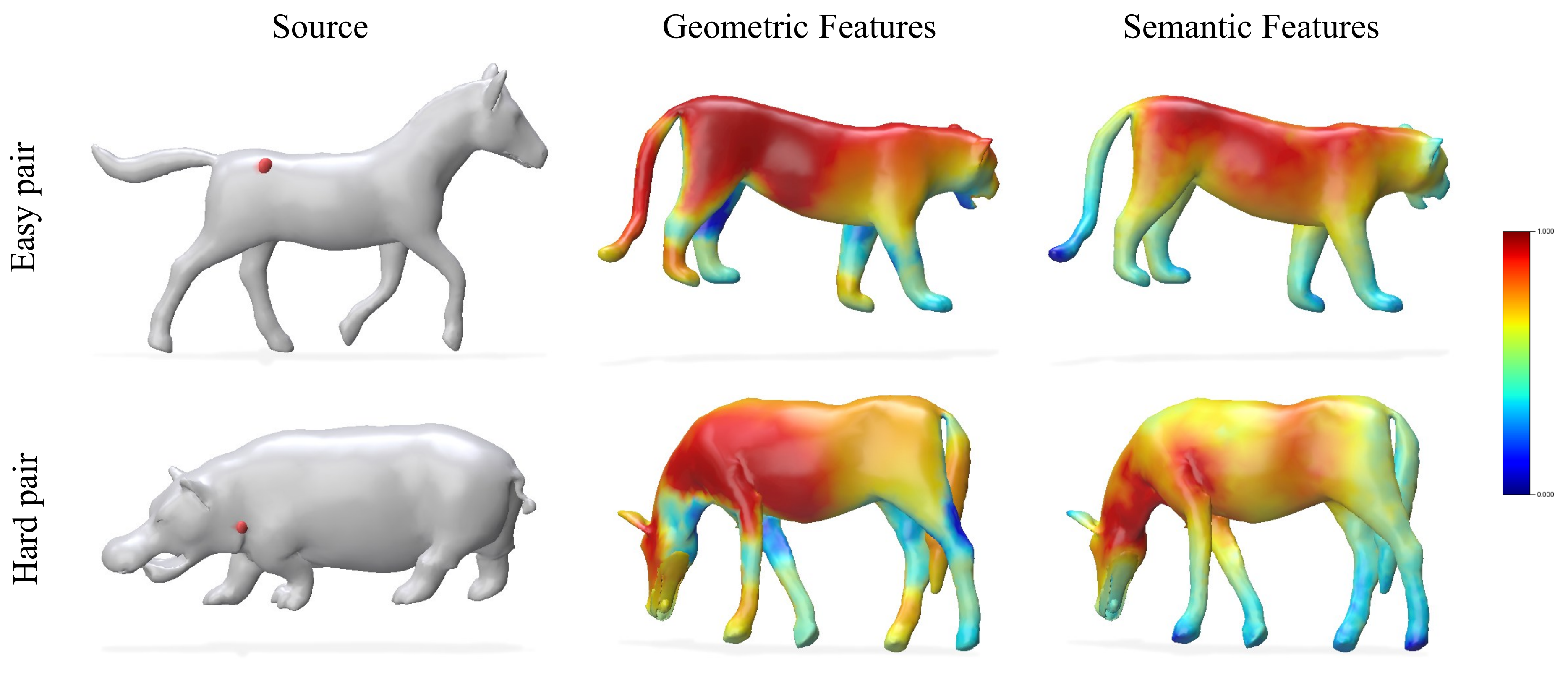}
\caption{\textbf{Feature similarity heatmaps on SMAL cross-species pairs.} For each red-marked source query point, we visualize cosine similarity to all target vertices using geometric (middle) and semantic (right) features. Semantic features produce more localized and semantically consistent responses.}
\label{fig:feat_sim}
\end{figure}

\begin{table}[!t]
\centering
\small
\caption{Comparison of sampling strategies in the CFM objective on SMAL.}
\label{tab:importance_sampling}
\setlength{\tabcolsep}{4pt}
\begin{tabular}{lcc}
\toprule
\textbf{Settings} & Uniform sampling & Confidence sampling (Ours) \\
\midrule
\textbf{Geo. Err ($\times 100$)} & 2.6 & \textbf{2.5} \\
\bottomrule
\end{tabular}
\end{table}

\section{Semantic Feature Analysis}
\label{sec:supp_sem_feat}

We investigate whether 2D-pretrained semantic features maintain cross-species discriminability on 3D surfaces without geometric supervision. We evaluate this by comparing per-vertex similarity between geometric and semantic descriptors using challenging SMAL shape pairs.

For a source query vertex, we visualize its cosine similarity to all target vertices as a heatmap. Two representative pairs are selected to evaluate feature behavior across matching difficulties: an easy case with low geodesic error (Horse $\to$ Cougar) and a hard case with high geodesic error (Hippo $\to$ Horse).

\cref{fig:feat_sim} shows that the two descriptors behave differently. In the easy pair (top row), geometric features produce a false high response near the target tail root, while semantic features correctly ignore this region and focus on the torso. In the hard pair (bottom row), geometric features yield diffuse responses under large deformations, whereas semantic features maintain a localized distribution, showing that DINOv2 captures stable cross-species part correspondences. These results suggest semantic features provide the discriminability geometric descriptors lack, particularly in resolving ambiguities from symmetry or self-similarity.

\section{CFM Regularization Analysis}
\label{sec:supp_cfm_analysis}
We further analyze CFM from transport-field coherence and qualitative map smoothness.

\subsection{Transport-Field Coherence}
\label{subsec:supp_transport_coherence}
To directly assess the local coherence of the transport field, we compute the neighbor velocity deviation
\begin{equation}
d_i = \left\| \mathbf{u}_i - \frac{1}{|\mathcal{N}(i)|}\sum_{j \in \mathcal{N}(i)} \mathbf{u}_j \right\|_2,
\end{equation}
where $\mathcal{N}(i)$ denotes the mesh neighborhood. For SGMatch, $\mathbf{u}_i = \mathbf{v}_{\theta}(\mathbf{z}_{t,i}, t)$ is the learned velocity; for the variant without CFM, $\mathbf{u}_i = \mathbf{z}_{1,i} - \mathbf{z}_{0,i}$ is the endpoint displacement. As shown in \cref{tab:vel_deviation}, CFM substantially reduces the mean deviation on both SMAL and TOPKIDS, supporting its role as a smoothness-promoting transport prior.

\begin{table}[!t]
\centering
\small
\caption{Mean neighbor velocity deviation. Lower is better.}
\label{tab:vel_deviation}
\setlength{\tabcolsep}{8pt}
\begin{tabular}{lcc}
\toprule
\textbf{Dataset} & \textbf{w/o CFM} & \textbf{w/ CFM} \\
\midrule
SMAL & 26.65 & \textbf{1.59} \\
TOPKIDS & 15.72 & \textbf{0.015} \\
\bottomrule
\end{tabular}
\end{table}

\subsection{Qualitative Smoothness Analysis}
\label{subsec:supp_qual_smoothness}

\begin{figure}[t]
\centering
\includegraphics[width=\linewidth]{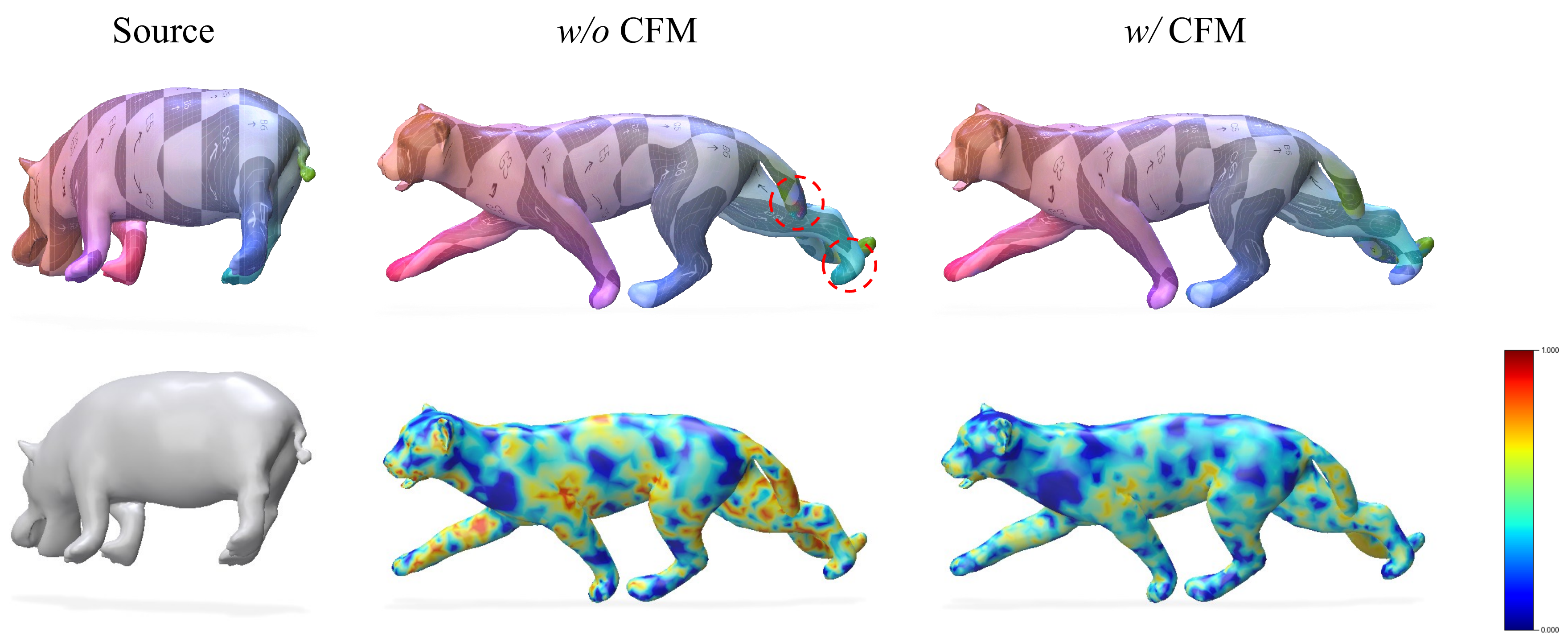}
\caption{\textbf{Effect of CFM regularization on correspondence smoothness for a challenging SMAL cross-species pair.} Top row shows texture transfer, and bottom row shows per-face conformal distortion. CFM suppresses local correspondence artifacts and reduces high-distortion regions.}
\label{fig:cfm_vis}
\end{figure}

\cref{fig:cfm_vis} displays texture transfer and per-face conformal distortion heatmaps on a hippo $\to$ cougar pair. In texture transfer (top row), omitting CFM causes local inconsistencies at tail and paw regions (red circles), where spatially adjacent vertices map to distant source regions, creating color discontinuities. With CFM, these artifacts are suppressed, and texture remains spatially coherent. 

Conformal distortion heatmaps (bottom row) offer a quantitative perspective on these results. Without CFM, high-distortion regions appear across the surface, with severe distortions concentrated around the limbs and tail, matching the correspondence artifacts in the texture transfer. With CFM, the share of high-distortion regions is reduced, and the overall distortion distribution shifts toward lower values.

\section{Semantic Backbone and View Sensitivity}
\label{sec:supp_backbone_view}
We evaluate how SGMatch depends on the quality of the lifted semantic features. We replace DINOv2 with weaker visual backbones while keeping the downstream matching architecture fixed. As shown in \cref{tab:backbone}, stronger visual backbones improve correspondence quality under large non-isometric deformation.

\begin{table}[H]
\centering
\small
\caption{Semantic backbone ablation on SMAL.}
\label{tab:backbone}
\setlength{\tabcolsep}{14pt}
\begin{tabular}{lccc}
\toprule
\textbf{Backbone} & \textbf{CLIP} & \textbf{DINOv1} & \textbf{DINOv2} \\
\midrule
\textbf{Geo.Err ($\times 100$) $\downarrow$} & 5.8 & 4.1 & \textbf{2.5} \\
\bottomrule
\end{tabular}
\end{table}

We also vary the number of rendered DINOv2 views. \cref{tab:views} shows that 50 views preserve the final accuracy of the 100-view setting, while fewer views gradually degrade performance.

\begin{table}[H]
\centering
\small
\caption{View number ablation on SMAL using DINOv2 features.}
\label{tab:views}
\setlength{\tabcolsep}{8pt}
\begin{tabular}{lcccc}
\toprule
\textbf{Views} & 12 & 25 & 50 & 100 \\
\midrule
\textbf{Geo.Err ($\times 100$) $\downarrow$} & 2.8 & 2.7 & \textbf{2.5} & \textbf{2.5} \\
\bottomrule
\end{tabular}
\end{table}

\section{Parameter Analysis}
\label{sec:supp_params}

\subsection{Neighborhood Size}
\label{sec:neigh}
We analyze SGLCA module sensitivity to neighborhood size $k$ on SMAL. As shown in \cref{tab:neigh}, performance is stable across $k$ values, with $k=16$ and $k=32$ yielding lowest errors. Small sizes ($k=8$) provide insufficient structural context for effective semantic-geometric fusion, degrading accuracy. Conversely, $k=64$ extends the receptive field beyond local geometry. Since neighbor lists derive from direct adjacency, vertices with fewer than $k$ neighbors are padded with self-loops, introducing redundant attention weights that dilute actual neighbor influence. We set $k=32$ as the default to balance local context coverage with attention focus.

\begin{table}[H]
\centering
\small
\caption{Effect of neighborhood size $k$ in the SGLCA module on the SMAL dataset.}
\label{tab:neigh}
\setlength{\tabcolsep}{4pt}
\begin{tabular}{lcccc}
\toprule
\textbf{Size} & $k=8$ & $k=16$ & $k=32$ (Ours) & $k=64$ \\
\midrule
\textbf{Geo. Err ($\times 100$)} & 2.7 & \textbf{2.5} & \textbf{2.5} & 2.6 \\
\bottomrule
\end{tabular}
\end{table}

\subsection{Diffusion Time}
\label{sec:diff.time}
Following~\cite{cao2024synchronous}, we adopt a diffusion time of $\tau = 10^{-2}$ for near-isometric matching, $\tau = 10^{-4}$ for non-isometric matching and $\tau = 10^{-1}$ for topology-noisy matching. We further examine the sensitivity of SGMatch to this parameter using the SMAL and TOPKIDS datasets.

\cref{tab:diffusion_smal} demonstrates that SMAL performance remains stable for $\tau$ between $10^{-5}$ and $10^{-3}$, with degradation occurring only at $\tau = 10^{-2}$. For non-isometric deformations, excessive diffusion likely smooths out the local geometric details required for cross-category correspondence. Conversely, a moderate-to-small diffusion scale suppresses noise while preserving feature discriminability. The marginal fluctuation at $\tau = 10^{-5}$ is consistent with spectral heat diffusion approaching the identity operator as $\tau \to 0$: at extremely small scales, the smoothing effect becomes negligible, slightly reducing the stabilization benefit for CFM transport without the catastrophic detail loss seen at large $\tau$. These results confirm the robustness of our method across a broad range of $\tau$ values.

On TOPKIDS, performance is consistent for $\tau \in \{10^{-2}, 10^{-1}\}$, while $\tau = 10^{-1}$ gives the best result, as shown in \cref{tab:diffusion_topkids}. We therefore use $\tau = 10^{-1}$ for TOPKIDS. The limited gap between $10^{-2}$ and $10^{-1}$ also indicates that the improvements from SGMatch are robust and do not rely on intensive hyperparameter tuning.

\begin{table}[H]
\centering
\small
\caption{Effect of diffusion time $\tau$ on the SMAL dataset.}
\label{tab:diffusion_smal}
\setlength{\tabcolsep}{4pt}
\begin{tabular}{lccccc}
\toprule
\textbf{Diff. Time} & $10^{-2}$ & $10^{-3}$ & $10^{-4}$ (Ours) & $10^{-5}$ & $10^{-6}$ \\
\midrule
\textbf{Geo. Err ($\times 100$)} & 3.3 & \textbf{2.5} & \textbf{2.5} & 2.6 & \textbf{2.5} \\
\bottomrule
\end{tabular}
\end{table}

\begin{table}[H]
\centering
\small
\caption{Effect of diffusion time $\tau$ on the TOPKIDS dataset.}
\label{tab:diffusion_topkids}
\setlength{\tabcolsep}{4pt}
\begin{tabular}{lccccc}
\toprule
\textbf{Diff. Time} & $1$ & $10^{-1}$ (Ours) & $10^{-2}$ & $10^{-3}$ & $10^{-4}$ \\
\midrule
\textbf{Geo. Err ($\times 100$)} & 3.3 & \textbf{2.9} & 3.3 & 3.5 & 3.5 \\
\bottomrule
\end{tabular}
\end{table}

\subsection{Confidence Weight Parameter}
\label{sec:param_alpha}
In the importance-weighted CFM objective, $\alpha$ controls the concentration of the confidence weights defined in Eq.~\eqref{eq:confidence_weight}. Larger $\alpha$ values prioritize vertices with reliable soft correspondences, while smaller $\alpha$ approaches uniform sampling. We evaluate $\alpha$ sensitivity on SMAL.

As shown in \cref{tab:sampling}, performance is stable for $\alpha \in \{0.5, 1.0, 2.0, 5.0\}$, indicating robustness to moderate variations. Excessively large $\alpha$ (e.g., $10.0$) concentrates weights on a tiny vertex subset with highest similarity, reducing training diversity and degrading performance. The partial recovery at $\alpha = 20.0$ likely stems from the interaction between extreme concentration and the Charbonnier loss, which limits outlier influence. We set $\alpha = 2.0$ as default to balance effective weighting with sampling distribution diversity.

\begin{table}[H]
\centering
\small
\caption{Sensitivity analysis of the confidence weight parameter $\alpha$ in importance sampling on the SMAL dataset.}
\label{tab:sampling}
\setlength{\tabcolsep}{4pt}
\begin{tabular}{lcccccc}
\toprule
$\alpha$ & 0.5 & 1.0 & 2.0 (Ours) & 5.0 & 10.0 & 20.0 \\
\midrule
\textbf{Geo.Err ($\times 100$)} & 2.6 & 2.6 & \textbf{2.5} & \textbf{2.5} & 3.6 & 2.7 \\
\bottomrule
\end{tabular}
\end{table}

\section{Statistical Analysis}
\label{sec:supp_stats}
We assess the robustness and stability of our approach by retraining on SHREC'19, SMAL, and TOPKIDS using three random seeds (42, 1234, and 3407). The mean geodesic error and standard deviation are reported and compared with HybridFMap under identical settings.

Results in \cref{tab:statistical} show that SGMatch yields lower mean errors than HybridFMap across all tested datasets. SGMatch also maintains much lower standard deviation (e.g., $0.01$ vs. $0.17$ on SMAL; $0.06$ vs. $1.02$ on TOPKIDS), suggesting greater stability regarding random initialization. This performance gain likely stems from the SGLCA module's semantic guidance and the regularization of CFM, which together lead to a more stable optimization process. In comparison, the higher variance of HybridFMap reflects a heavier reliance on initialization, potentially limiting its practical reliability.

\begin{table}[H]
\centering
\small
\caption{Mean geodesic error ($\times 100$) with standard deviation over three random seeds on SHREC'19, SMAL, and TOPKIDS.}
\label{tab:statistical}
\setlength{\tabcolsep}{6pt}
\begin{tabular}{lccc}
\toprule
\textbf{Method} & SHREC'19 & SMAL & TOPKIDS \\
\midrule
HybridFMap & $4.17 \pm 0.68$ & $3.52 \pm 0.17$ & $5.38 \pm 1.02$ \\
Ours       & $3.33 \pm 0.07$ & $2.51 \pm 0.01$ & $3.36 \pm 0.06$ \\
\bottomrule
\end{tabular}
\end{table}

\section{Runtime and Memory Analysis}
\label{sec:supp_runtime}
We evaluate the computational efficiency of SGMatch by reporting the average inference time per pair and peak memory usage across several datasets, using HybridFMaps as a baseline. All measurements are conducted on the same hardware. Our model contains 2.8M parameters, while HybridFMaps contains 0.5M. Although our parameter count is higher due to the SGLCA module and CFM regularization, this trade-off yields significant performance improvements, particularly in non-isometric settings and those with topological noise.

\begin{table}[H]
\centering
\small
\caption{Inference time (seconds per shape pair) and peak GPU memory (GB) comparison on each benchmark. }
\label{tab:cost}
\begin{tabular}{llcccccc}
\toprule
\multirow{2}{*}{\textbf{Method}} & \multirow{2}{*}{\textbf{Metric}} & 
\multicolumn{6}{c}{\textbf{Dataset}} \\
\cmidrule(lr){3-8}
& & FAUST & SCAPE & SHREC'19 & SMAL & DT4D (inter) & TOPKIDS \\
\midrule
\multirow{2}{*}{HybridFMap} 
  & Time (s) &  0.11 & 0.10 & 5.42 & 6.98 & 10.79 & 11.93 \\
  & Mem. (GB) & 3.3 & 3.4 & 3.3 & 3.7 & 5.5 & 18.7 \\
\midrule
\multirow{2}{*}{Ours} 
  & Time (s)  & 0.08 & 0.07 & 6.27 & 7.11 & 8.26 & 10.93 \\
  & Mem. (GB) & 3.9  & 4.0   & 4.0   & 3.2   & 6.3   & 16.3 \\
\bottomrule
\end{tabular}
\end{table}

As shown in \cref{tab:cost}, our method achieves inference times similar to HybridFMap on near-isometric datasets (e.g., 0.08s vs. 0.11s for FAUST). We observe a slight increase in runtime on larger benchmarks such as SHREC'19 and DT4D, reflecting the added complexity of our modules. Regarding peak GPU memory, our method requires roughly 0.6 GB more than HybridFMap on near-isometric data but remains within a similar range on non-isometric and topological-noise datasets. These results indicate that the additional computational overhead is manageable and well-compensated by the accuracy gains under challenging deformations.

Semantic feature extraction is performed once per shape and can be cached for repeated pairwise evaluations. We report this one-time preprocessing cost in \cref{tab:preprocess_time}.

\begin{table}[t]
\centering
\small
\caption{Semantic feature extraction time per shape.}
\label{tab:preprocess_time}
\setlength{\tabcolsep}{8pt}
\begin{tabular}{lcccccc}
\toprule
\textbf{Dataset} & FAUST & SCAPE & SMAL & SHREC'19 & DT4D & TOPKIDS \\
\midrule
\textbf{Time (s)} & 34.7 & 42.1 & 38.5 & 36.2 & 36.2 & 109.4 \\
\bottomrule
\end{tabular}
\end{table}

\FloatBarrier

\section{More Qualitative Results}
\label{sec:supp_qual}
In the figures below, we provide additional qualitative results of our method on SHREC'19, SMAL, DT4D-H inter class, and TOPKIDS datasets.

\FloatBarrier

\begin{figure*}[!htbp]
\centering
\def\imgOne{43}
\def\imgTwo{43-23}
\def\imgThree{43-20}
\def\imgFour{43-6}
\def\imgFive{43-31}
\def\pathShrecNT{figs/ours/additional_shrec19/}
\def\hspaceCols{0.15cm}
\def\wspaceRows{0.15cm}
\def\height{3.4cm}
\begin{tabular}{ccccc}
    \setlength{\tabcolsep}{3pt}
    {\small Source} & & & & \\
    \vspace{\wspaceRows}
    \hspace{\hspaceCols}\includegraphics[height=\height]{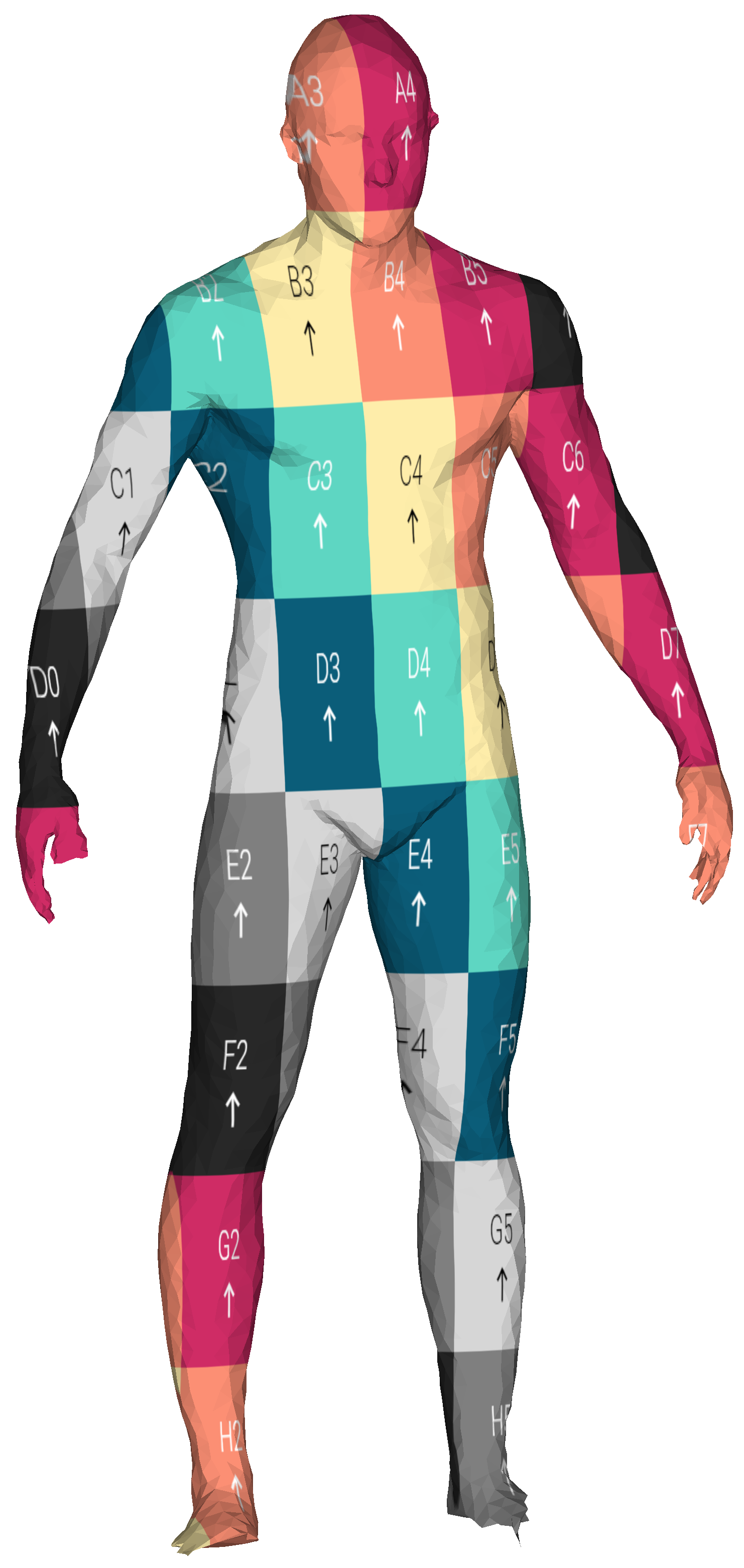} &
    \hspace{\hspaceCols}\includegraphics[height=\height]{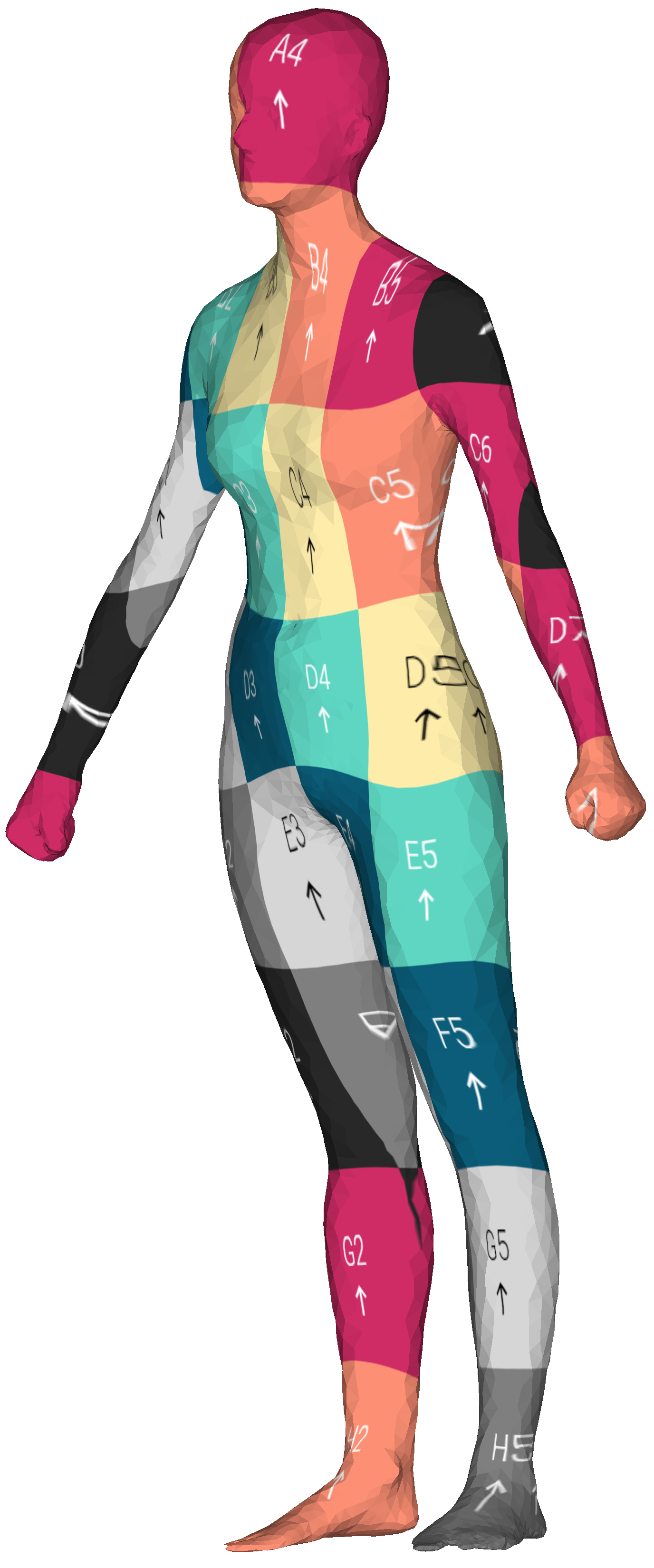} &
    \hspace{\hspaceCols}\includegraphics[height=\height]{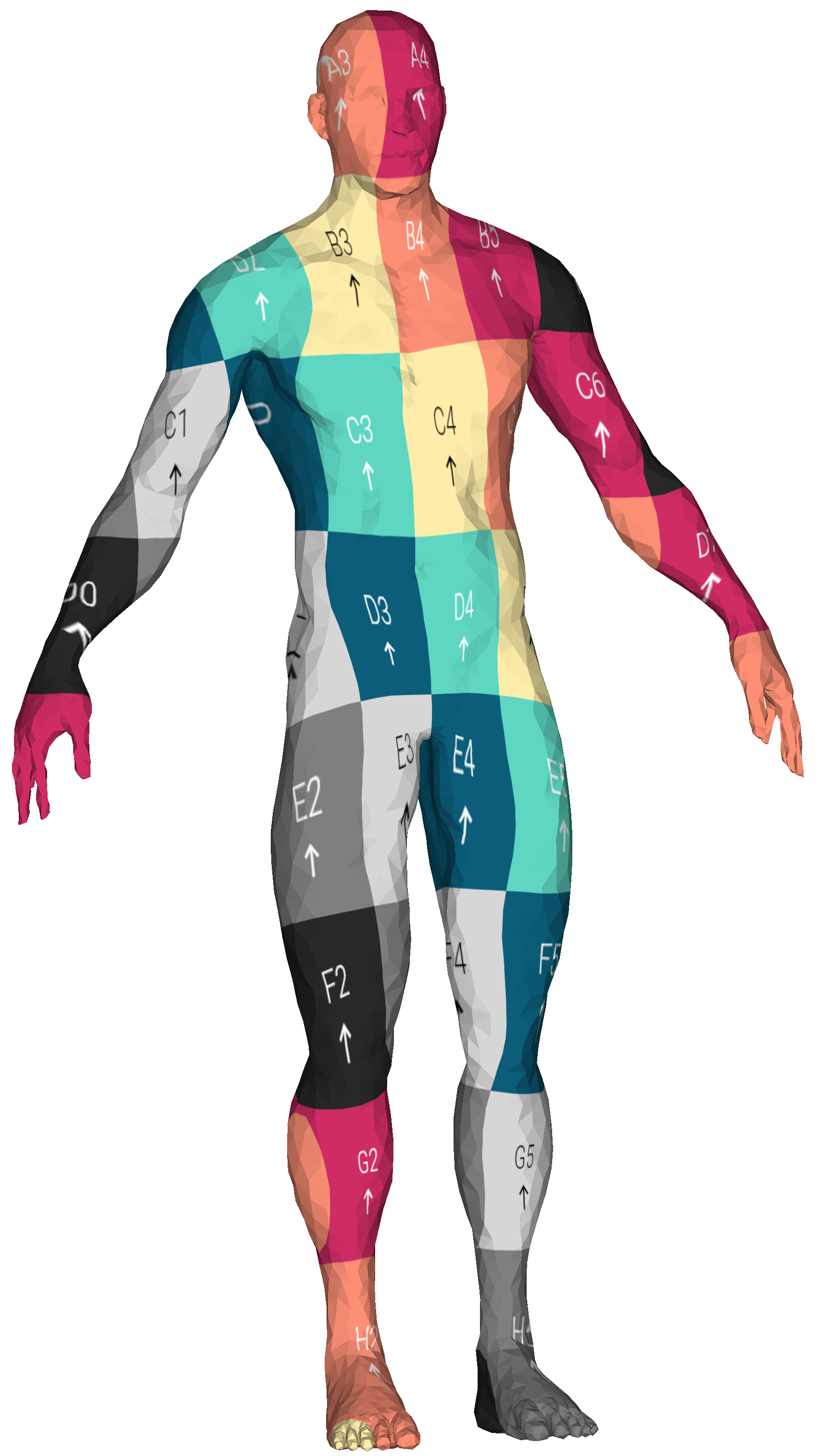} &
    \hspace{\hspaceCols}\includegraphics[height=\height]{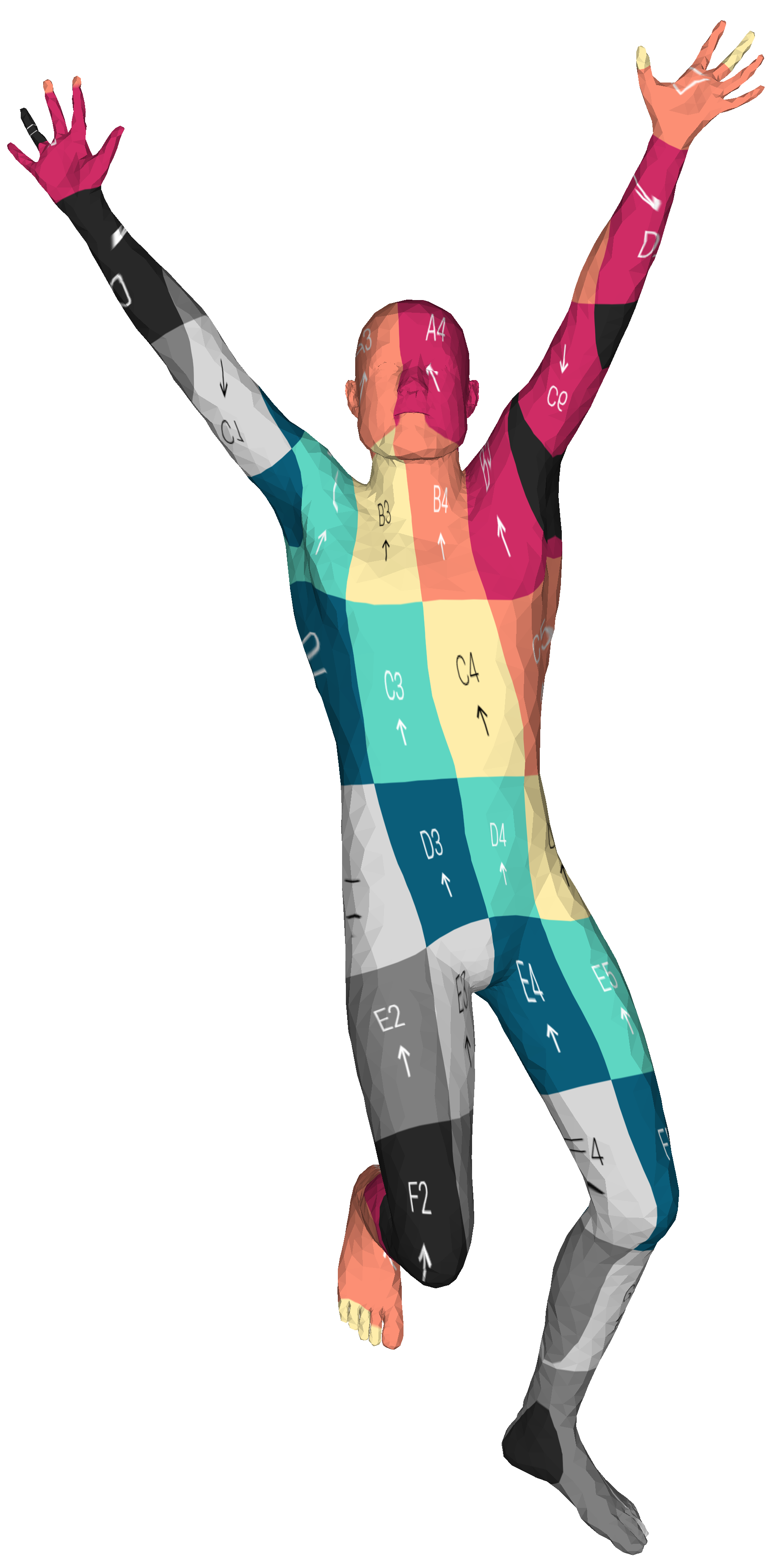} &
    \hspace{\hspaceCols}\includegraphics[height=\height]{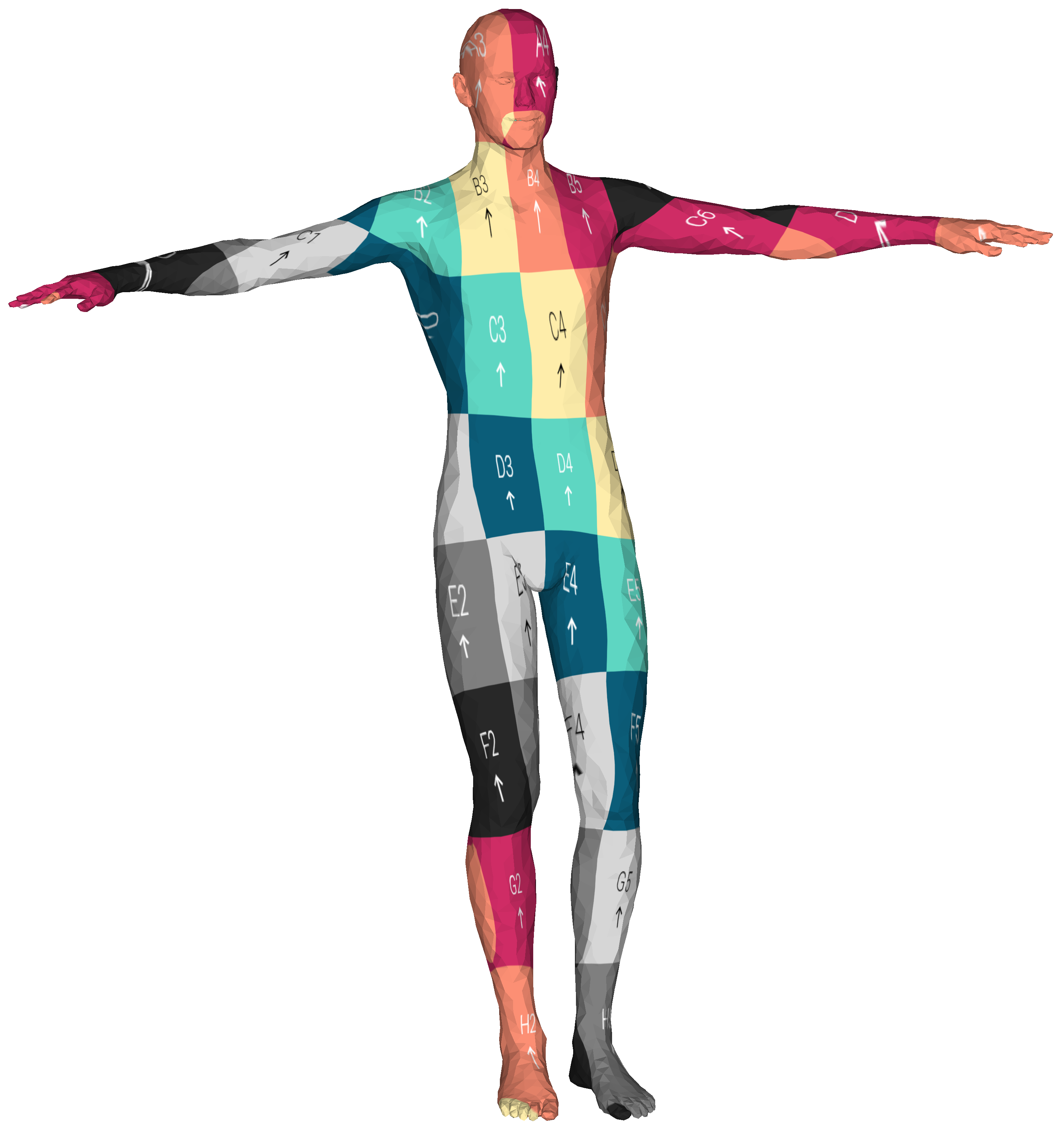}
\end{tabular}

\caption{Additional qualitative results of our method on SHREC'19.}
\label{fig:additional-shrec19}
\end{figure*}

\begin{figure*}[!htbp]
\centering
\def\imgOne{horse03}
\def\imgTwo{horse03-02}
\def\imgThree{horse03-hippo06}
\def\imgFour{horse03-cougar03}
\def\pathSmal{figs/ours/additional_smal/}
\def\hspaceCols{0.01cm}
\def\wspaceRows{0.15cm}
\def\height{2cm}
\begin{tabular}{cccc}
    \setlength{\tabcolsep}{0.5pt}
    {\small Source} & & & \\
    \vspace{\wspaceRows}
    \hspace{\hspaceCols}\includegraphics[height=\height]{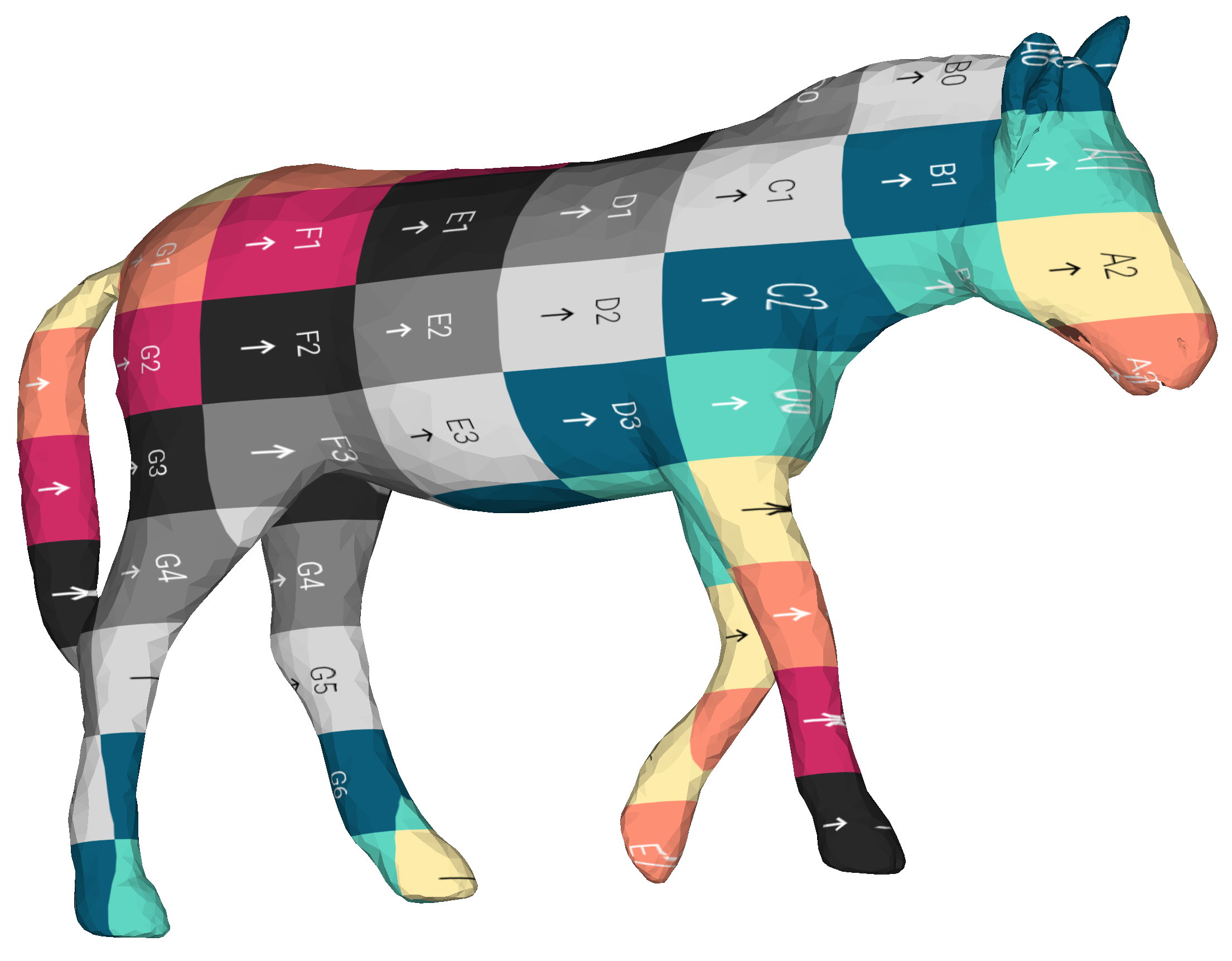} &
    \hspace{\hspaceCols}\includegraphics[height=\height]{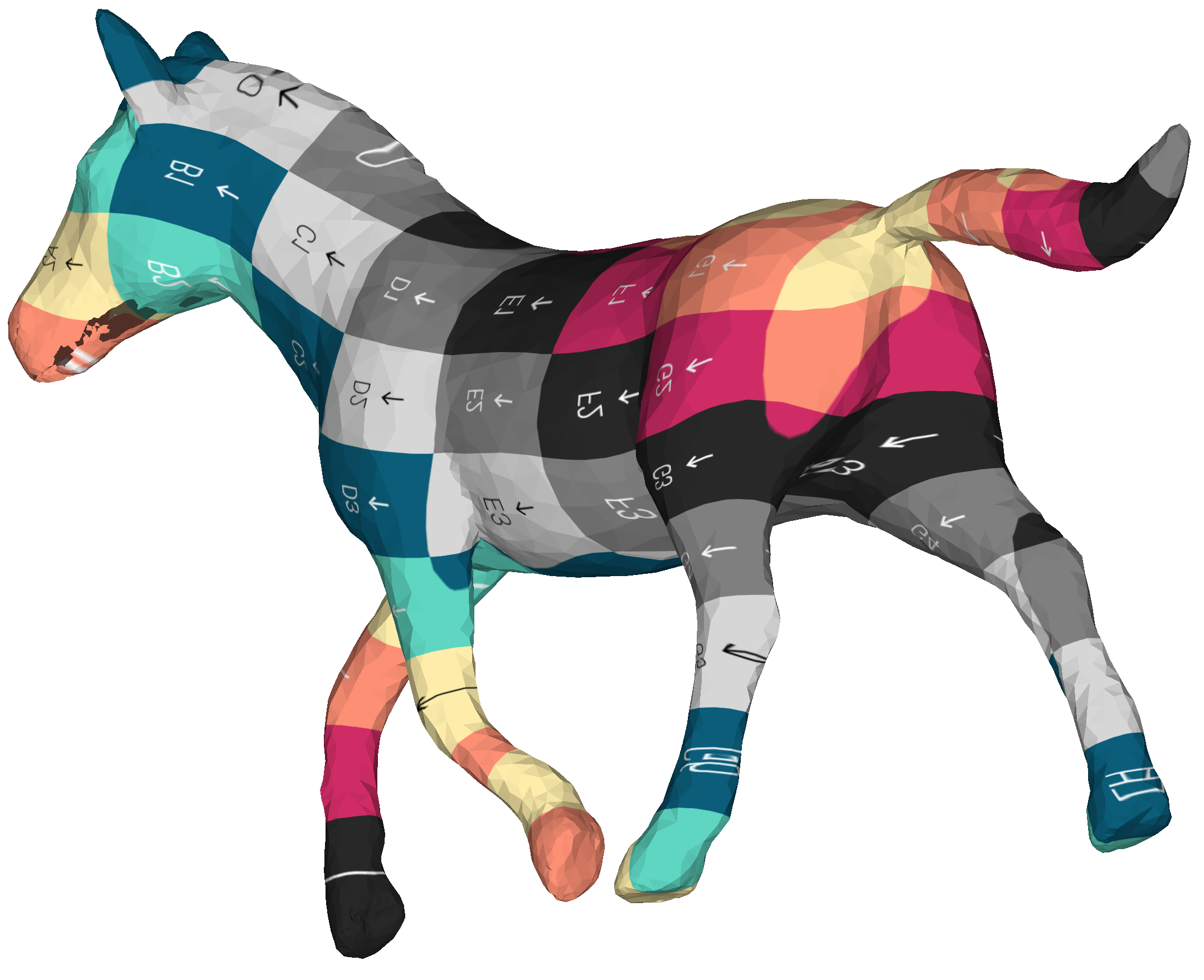} &
    \hspace{\hspaceCols}\includegraphics[height=\height]{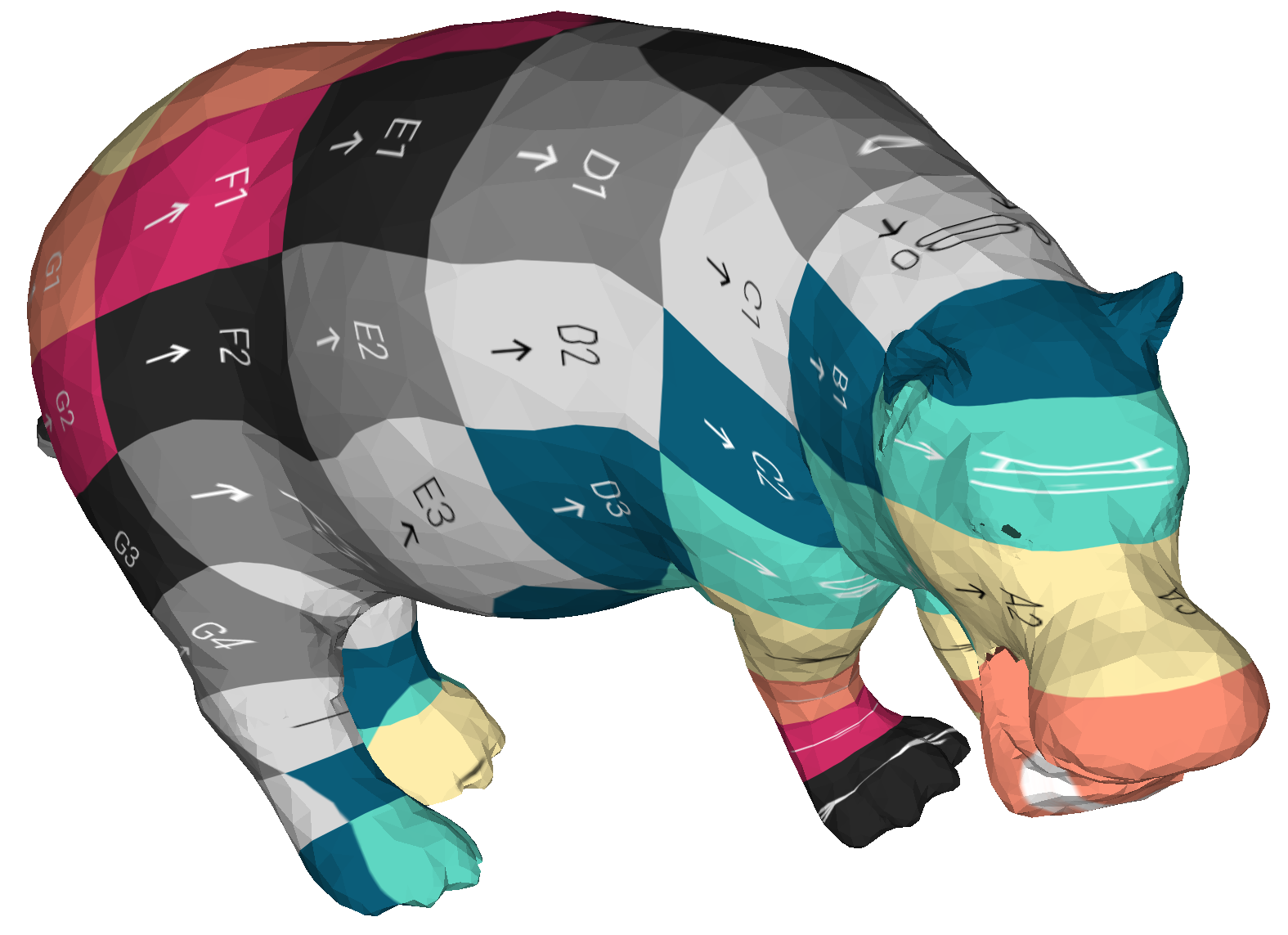} &
    \hspace{\hspaceCols}\includegraphics[height=\height]{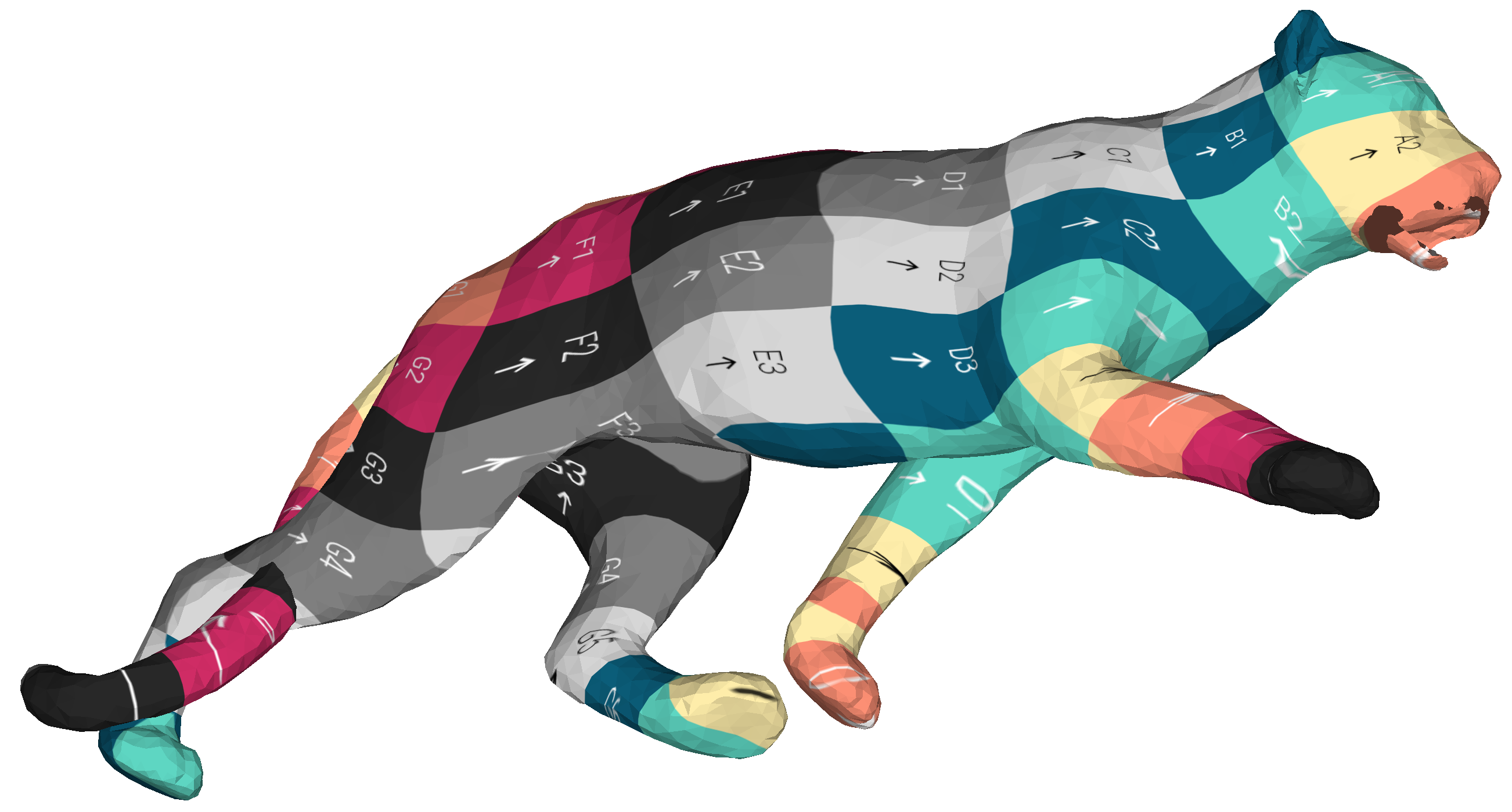}
\end{tabular}

\caption{Additional qualitative results of our method on SMAL.}
\label{fig:additional-smal}
\end{figure*}

\begin{figure*}[!htbp]
\centering
\def\imgOne{InvertedDoubleKickToKipUp194}
\def\imgTwo{InvertedDoubleKickToKipUp194-Standing2HMagicAttack0104300}
\def\imgThree{InvertedDoubleKickToKipUp194-Standing2HMagicAttack0108100}
\def\imgFour{InvertedDoubleKickToKipUp194-StandingReactLargeFromLeft00500}
\def\imgFive{InvertedDoubleKickToKipUp194-StandingReactLargeFromLeft01000}
\def\pathDTFourD{figs/ours/additional_dt4d/}
\def\hspaceCols{0.05cm}
\def\wspaceRows{0.15cm}
\def\height{3.0cm}
\begin{tabular}{ccccc}
    \setlength{\tabcolsep}{1pt}
    {\small Source} & & & & \\
    \vspace{\wspaceRows}
    \hspace{\hspaceCols}\includegraphics[height=\height]{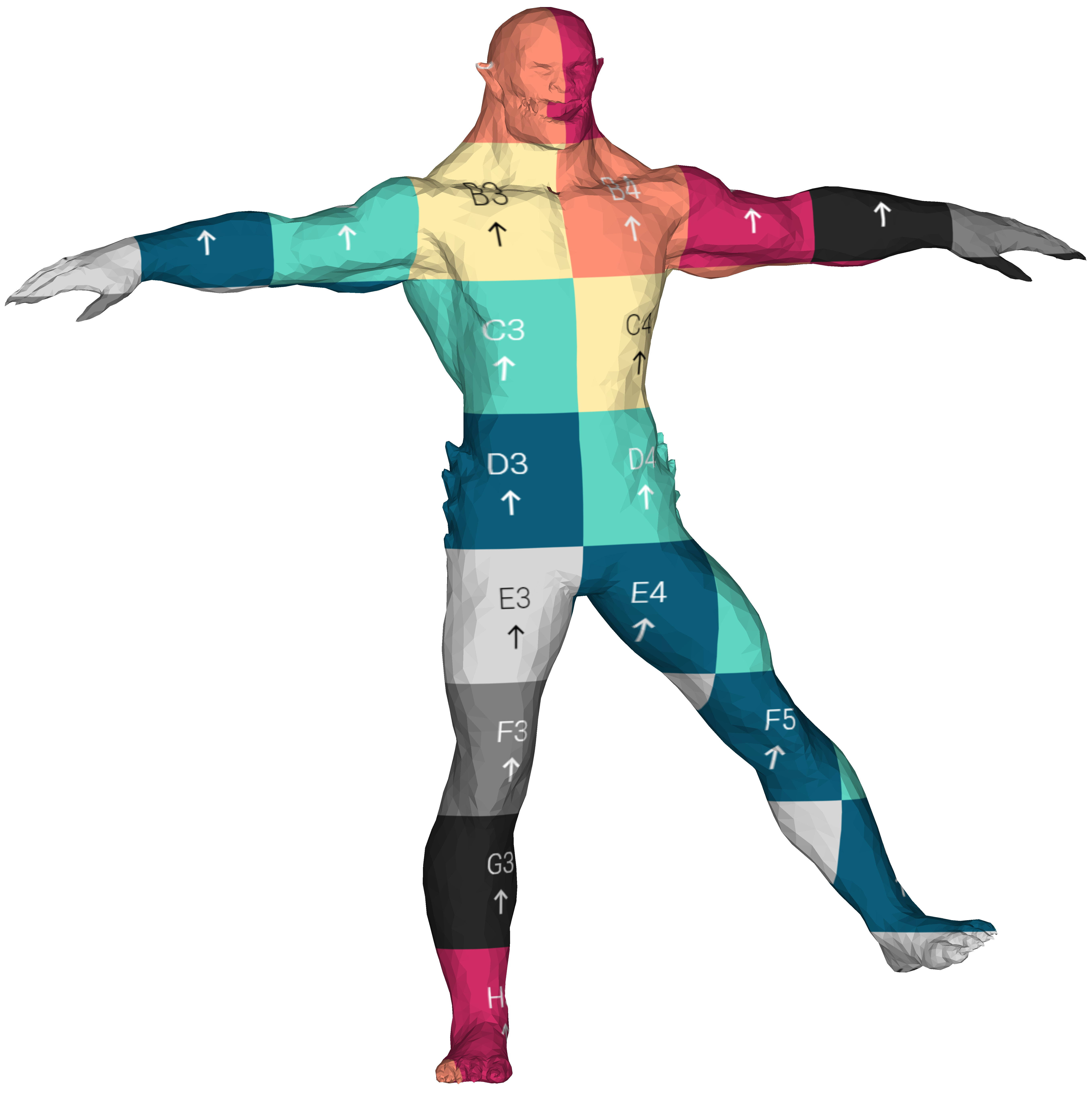} &
    \hspace{\hspaceCols}\includegraphics[height=\height]{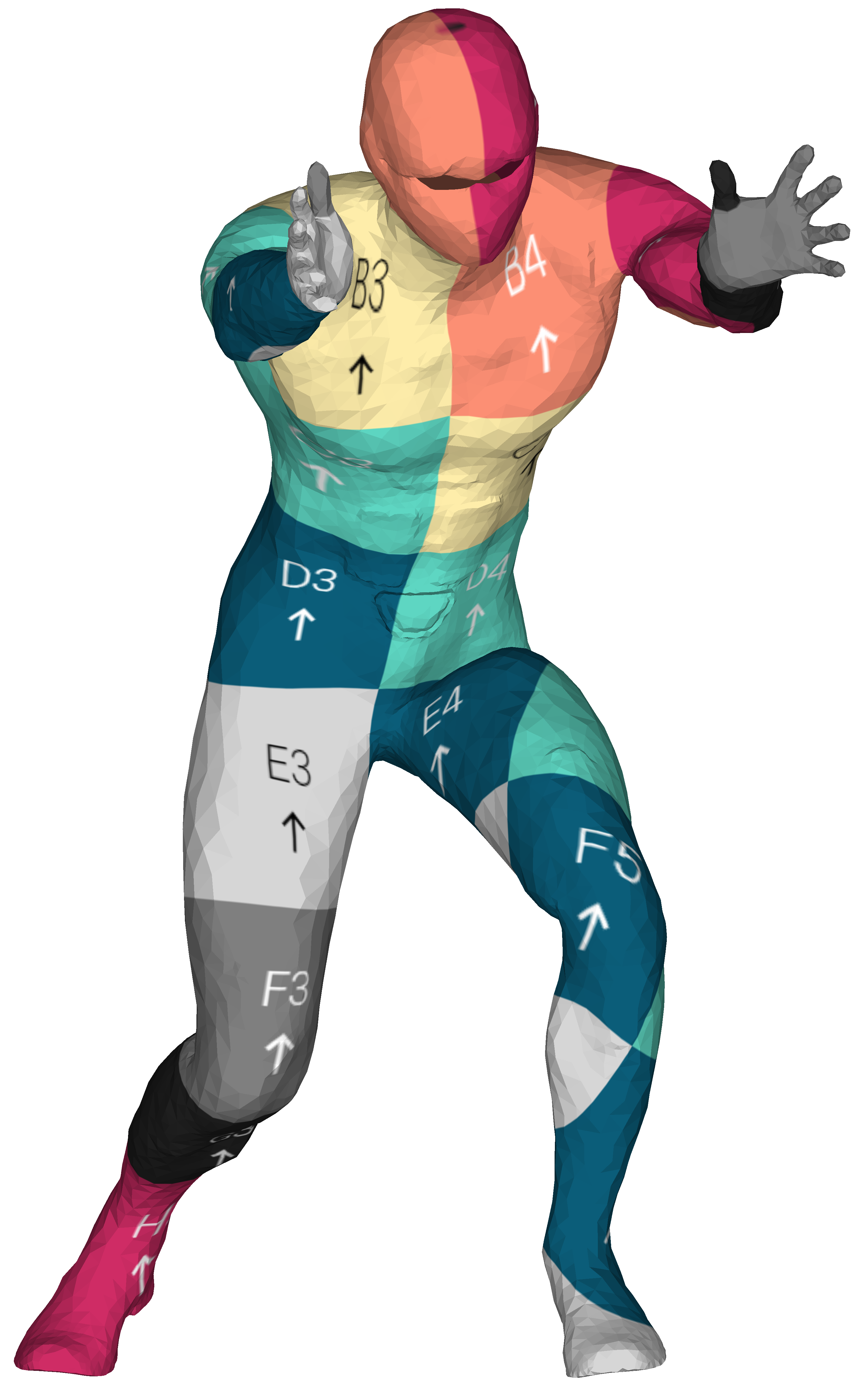} &
    \hspace{\hspaceCols}\includegraphics[height=\height]{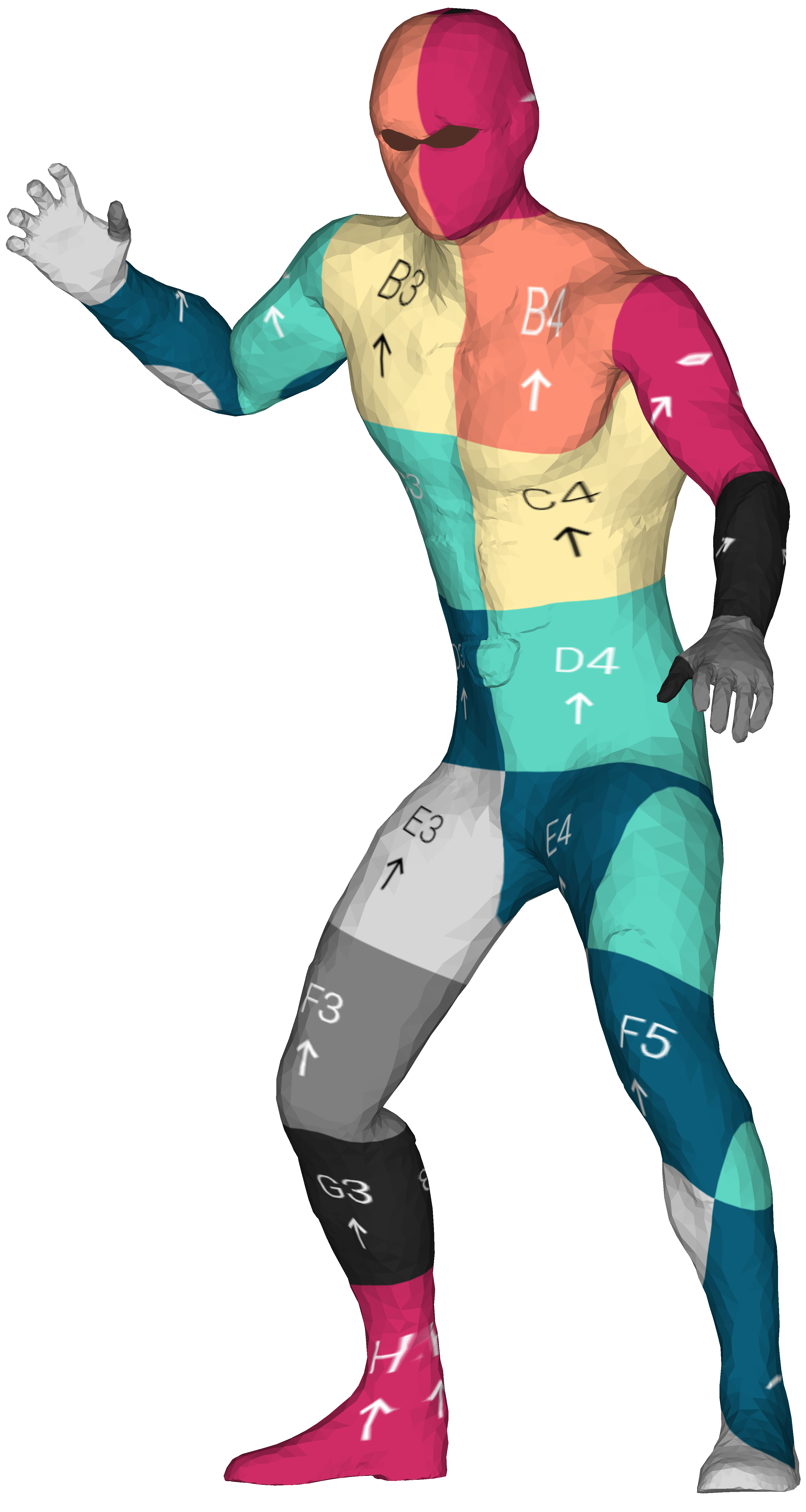} &
    \hspace{\hspaceCols}\includegraphics[height=\height]{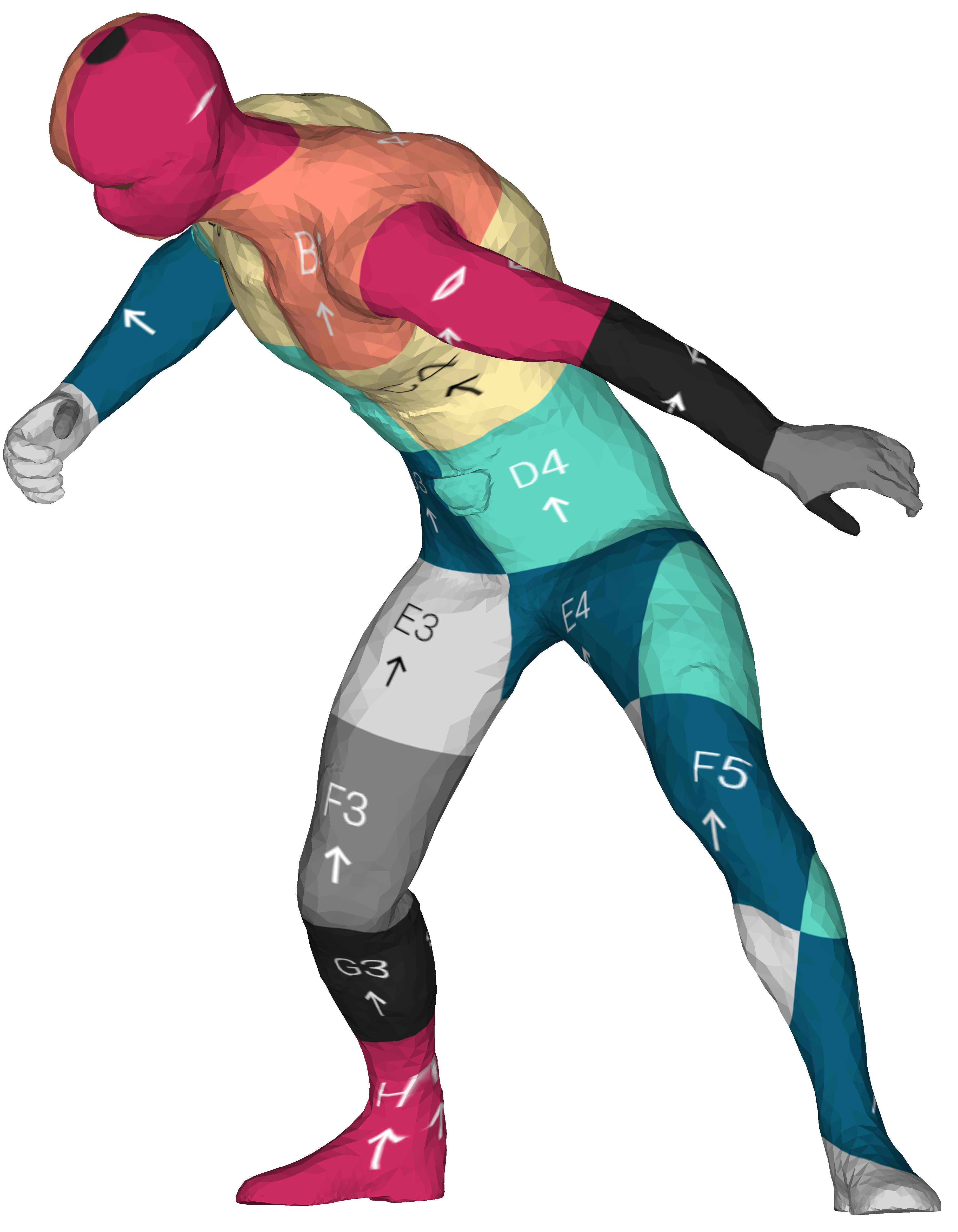} &
    \hspace{\hspaceCols}\includegraphics[height=\height]{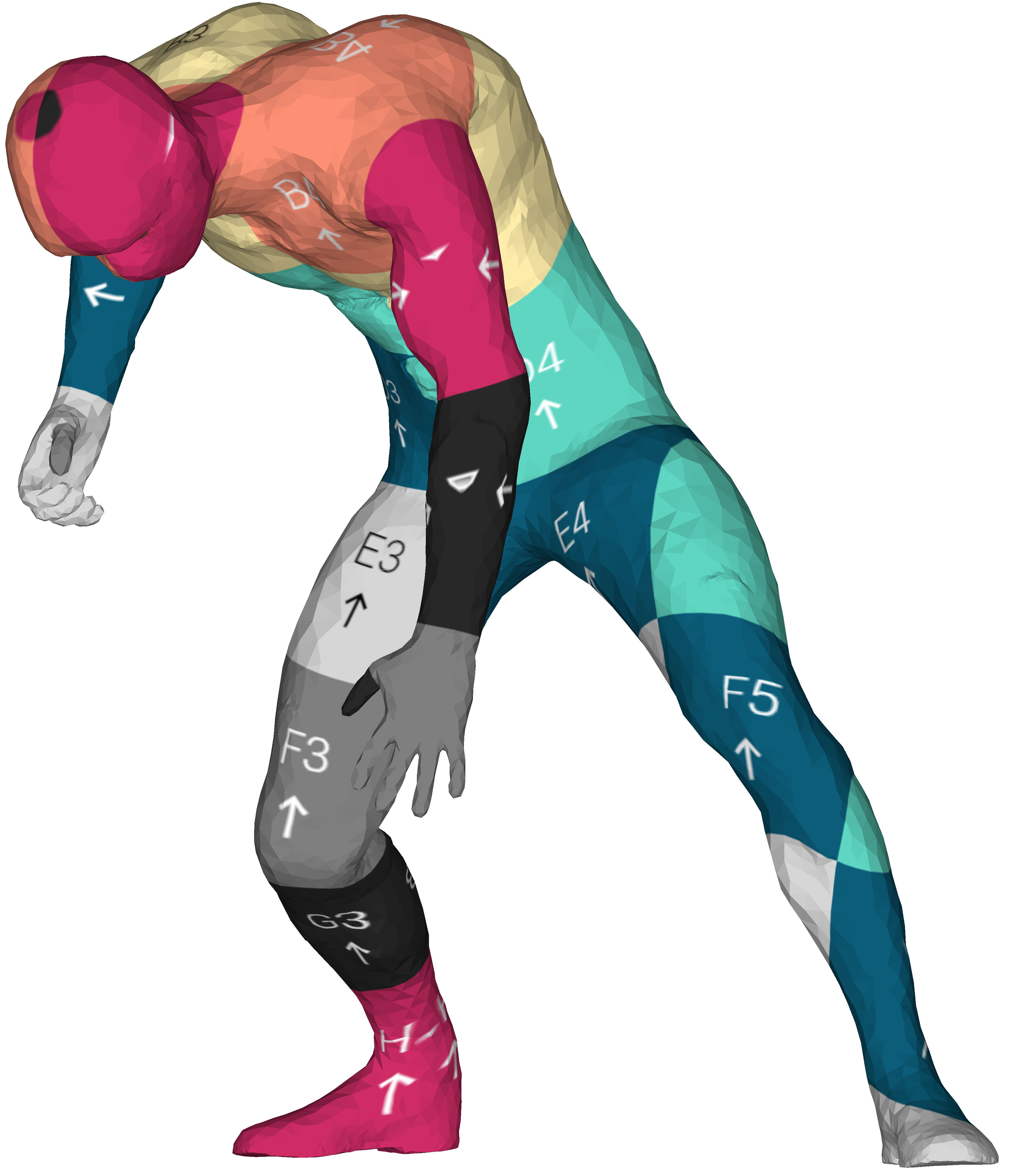}
\end{tabular}

\caption{Additional qualitative results of our method on DT4D-H inter class.}
\label{fig:additional-dt4d}
\end{figure*}

\begin{figure*}[!htbp]
\centering
\def\imgOne{kid00}
\def\imgTwo{kid00-kid01}
\def\imgThree{kid00-kid02}
\def\imgFour{kid00-kid04}
\def\imgFive{kid00-kid05}
\def\imgSix{kid00-kid08}
\def\pathTopkids{figs/ours/additional_topkids/}
\def\hspaceCols{0.1cm}
\def\wspaceRows{0.15cm}
\def\height{3.2cm}
\begin{tabular}{cccccc}
    \setlength{\tabcolsep}{1pt}
    {\small Source} & & & & & \\
    \vspace{\wspaceRows}
    \hspace{\hspaceCols}\includegraphics[height=\height]{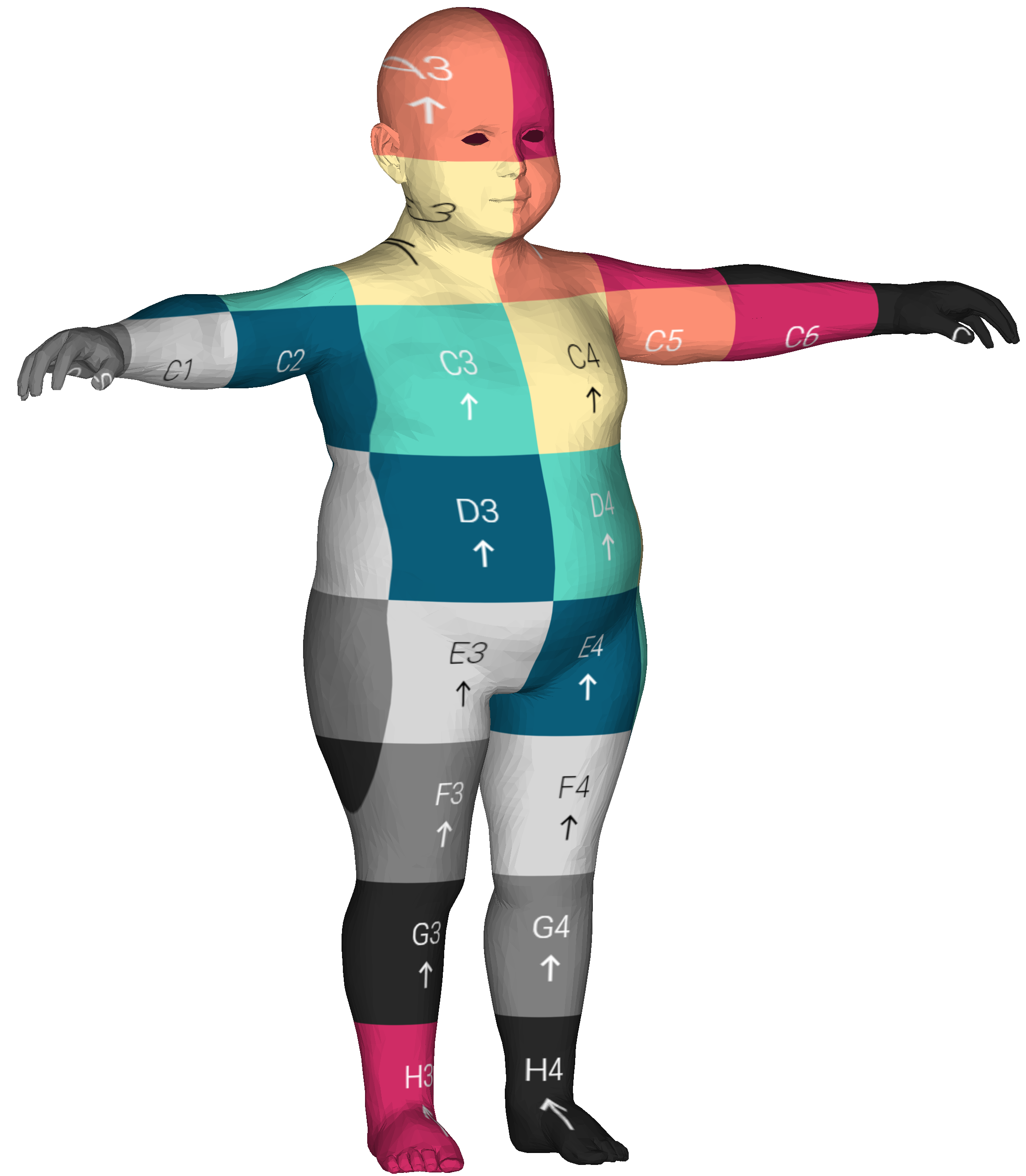} &
    \hspace{\hspaceCols}\includegraphics[height=\height]{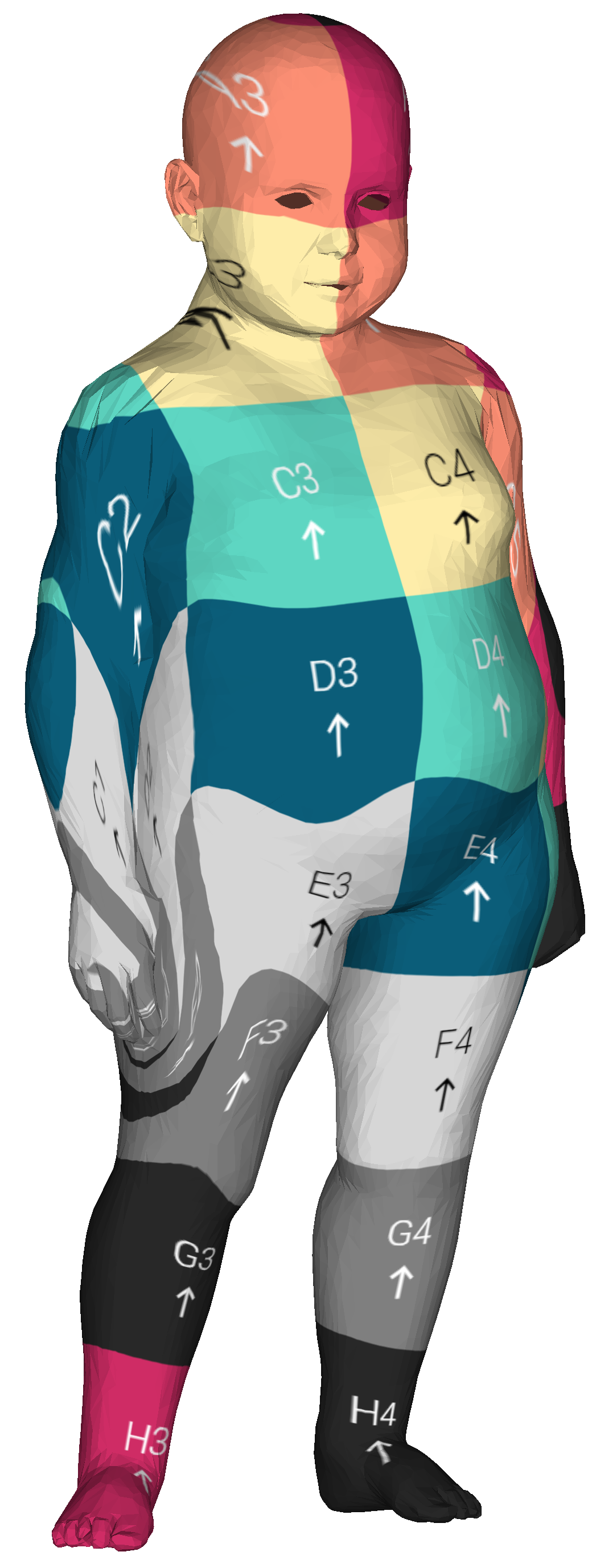} &
    \hspace{\hspaceCols}\includegraphics[height=\height]{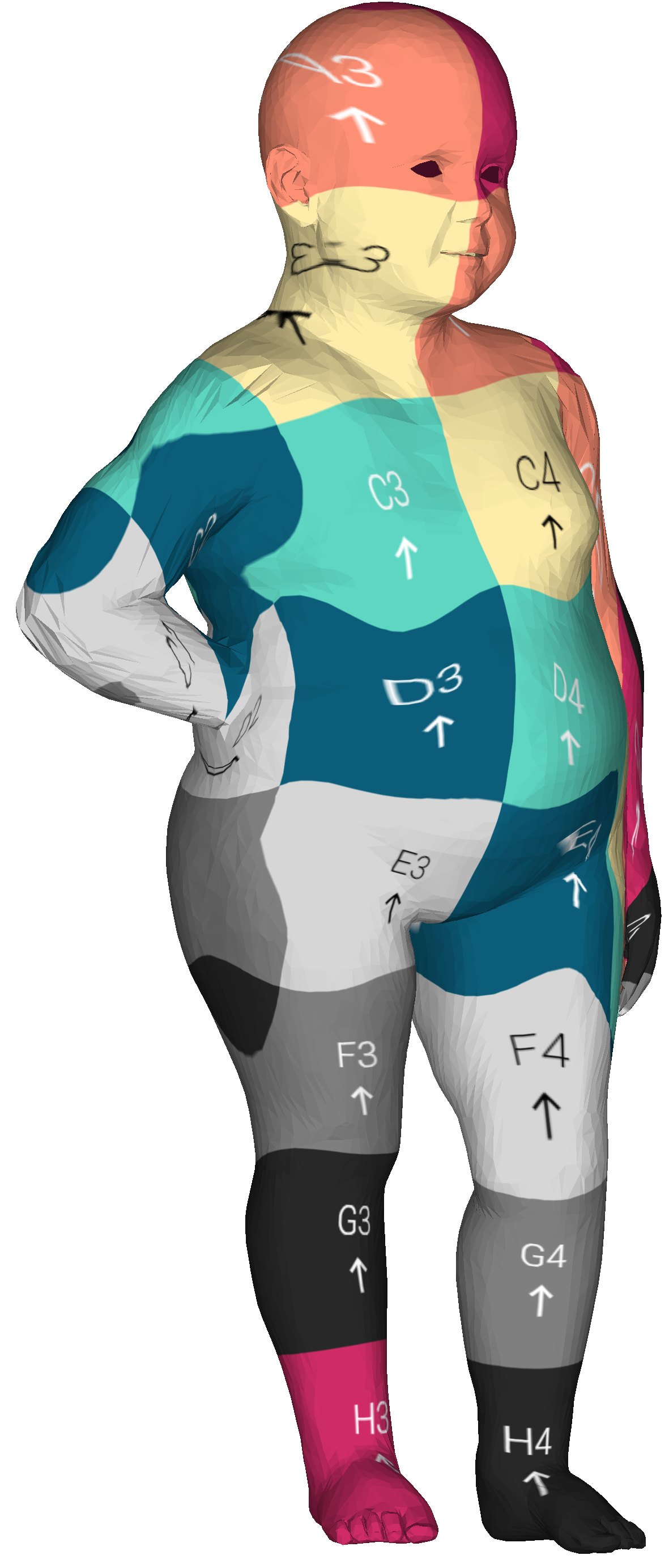} &
    \hspace{\hspaceCols}\includegraphics[height=\height]{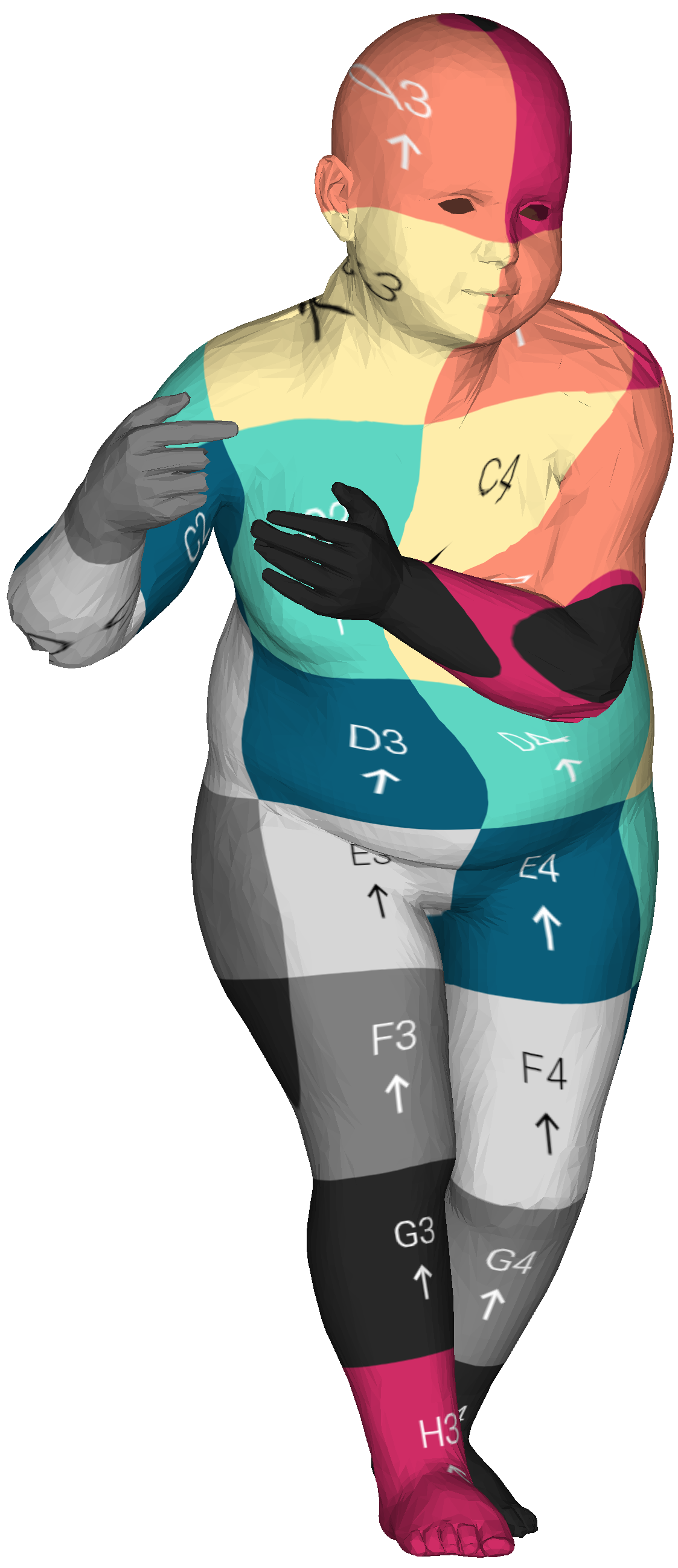} &
    \hspace{\hspaceCols}\includegraphics[height=\height]{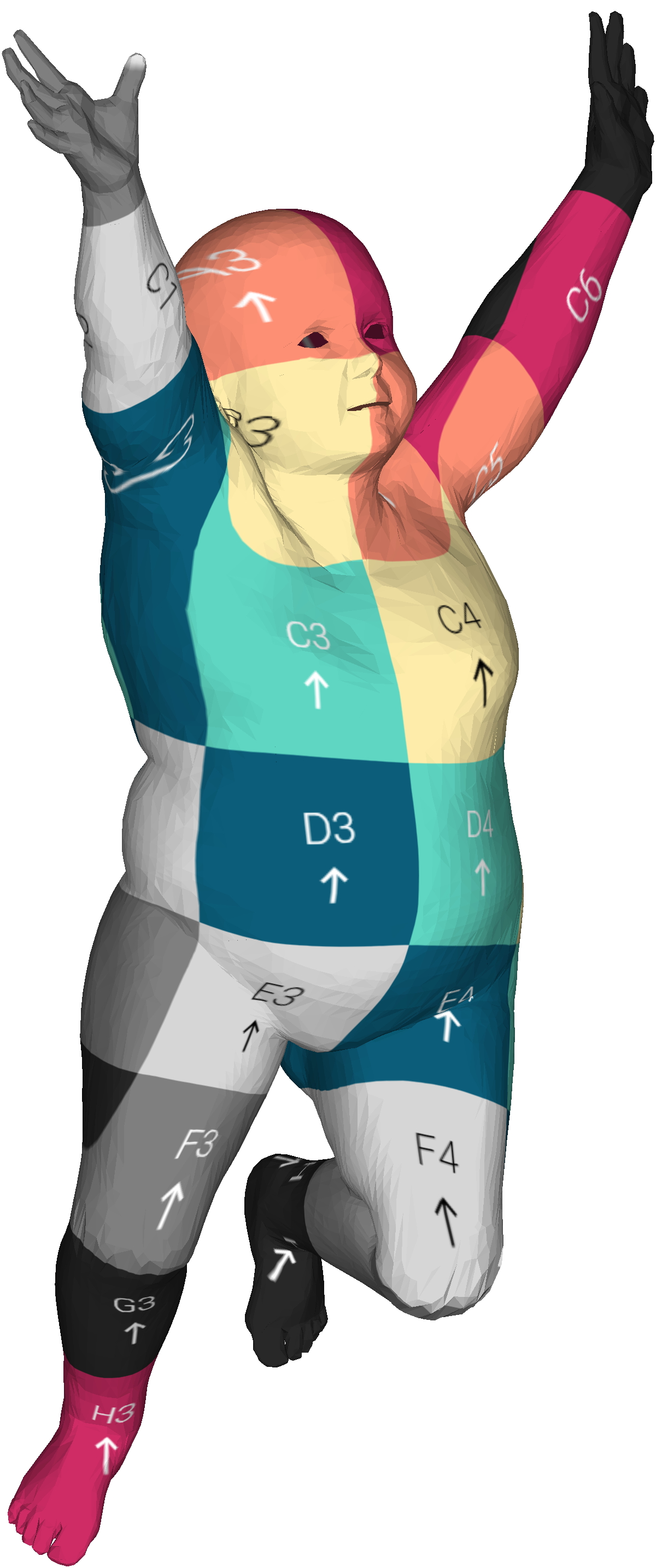} &
    \hspace{\hspaceCols}\includegraphics[height=\height]{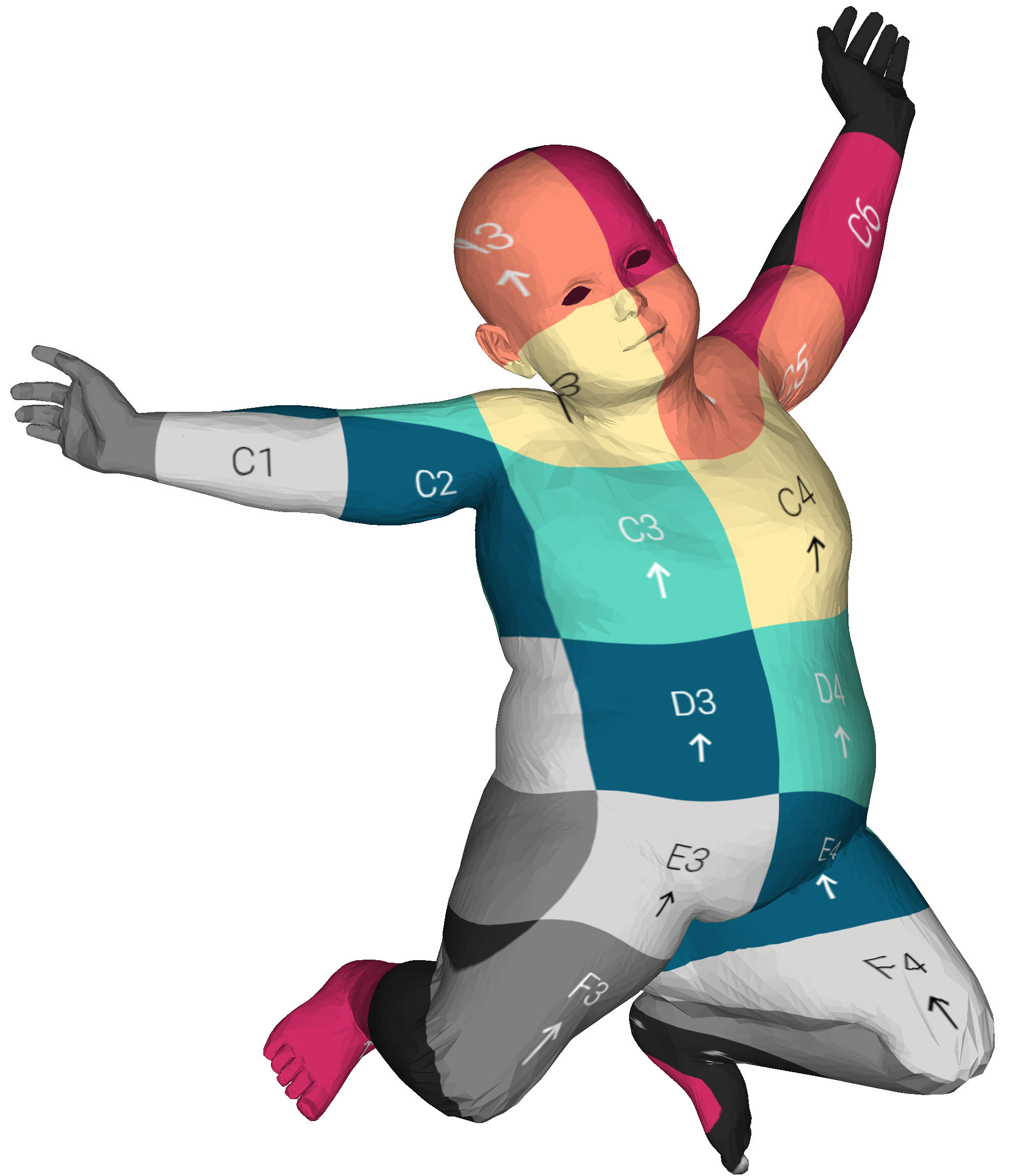}
\end{tabular}

\caption{Additional qualitative results of our method on TOPKIDS.}
\label{fig:additional-topkids}
\end{figure*}

\end{document}